\ifdefined\pdfobjcompresslevel
\fi
\documentclass{article}

\PassOptionsToPackage{table}{xcolor}
\usepackage{iclr2027_conference}
\iclrfinalcopy %

\usepackage{times}
\usepackage{latexsym}
\usepackage{enumitem}
\usepackage{xurl}

\usepackage[T1]{fontenc}

\usepackage[utf8]{inputenc}

\usepackage{microtype}

\usepackage{inconsolata}
\usepackage{fvextra} %

\usepackage{tabularx, booktabs, longtable}

\usepackage{graphicx}
\usepackage[table]{xcolor}
\usepackage{multirow}
\usepackage{booktabs}
\usepackage{amsmath,amssymb}
\usepackage{pifont}
\usepackage{wasysym} %
\usepackage{makecell}
\newcommand{\cmark}{\ding{51}}
\newcommand{\xmark}{\ding{55}}

\usepackage{listings}
\usepackage{xcolor}
\usepackage{caption} %
\makeatletter
\renewcommand\paragraph{\@startsection{paragraph}{4}{\z@}%
  {0.25ex \@plus 0.25ex}{-1em}{\normalsize\bfseries}}
\makeatother
\usepackage{needspace}
\usepackage{float}
\usepackage{placeins}
\usepackage{wrapfig} %
\usepackage{titletoc} %

\usepackage[most]{tcolorbox}

\usepackage{xspace}
\newcommand{\mempa}{MemPABench\xspace} %
\newcounter{finding}
\newcommand{\finding}[2]{\refstepcounter{finding}\label{#1}\paragraph{\textit{Finding~\thefinding}\enspace #2}}

\usepackage{pifont} %
\usepackage{pifont}
\usepackage{xcolor} %
\usepackage{setspace}

\newtcolorbox{instructionbox}{
  colback=gray!5,
  boxrule=0pt,
  leftrule=3pt,
  colframe=gray!50,
  sharp corners,
  boxsep=5pt,
  top=5pt,
  bottom=5pt
}

\providecommand{\cmark}{}
\providecommand{\xmark}{}

\renewcommand{\cmark}{\textcolor{green!50!black}{\ding{51}}}
\renewcommand{\xmark}{\textcolor{red}{\ding{55}}}
\newcommand{\pmark}{\textcolor{orange!80!black}{\LEFTcircle}}

\title{Why Bother Me Again? Benchmarking Personal Agent Memory for Interaction Preferences}

\author{%
  Jialin Li \quad
  Zhenhao Chen \quad
  Hanjun Luo \quad
  Hanan Salam\thanks{Corresponding author.} \\
  New York University Abu Dhabi, Abu Dhabi, UAE \\
  \texttt{jl10897@nyu.edu}
}

\let\cite\citep

\usepackage{hyperref}
\definecolor{linkdarkblue}{RGB}{0,0,139}
\hypersetup{colorlinks=true, citecolor=linkdarkblue, linkcolor=linkdarkblue, urlcolor=linkdarkblue}
\usepackage{tikz}
\usetikzlibrary{fadings}
\definecolor{barOne}{HTML}{DCD4F3}
\definecolor{barFile}{HTML}{FBB3AE}
\definecolor{barRAG}{HTML}{E5ECCE}
\definecolor{barMemZero}{HTML}{B4DFD0}
\definecolor{barMemOS}{HTML}{C3CCE8}
\definecolor{barNoMem}{HTML}{E4E4E4}
\definecolor{barOracle}{HTML}{E4E4E4}
\newlength{\cellbarwidth}\newlength{\cellbarunit}
\newcommand{\cellbar}[2]{\makebox[\cellbarwidth][l]{\tikz[baseline=0pt]{\clip (0,-0.55ex) rectangle (\cellbarwidth,1.65ex); \fill[#1!35] (0,-0.55ex) rectangle (\cellbarwidth,1.65ex); \fill[#1] (0,-0.55ex) rectangle (#2\cellbarunit-6\cellbarunit,1.65ex); \shade[left color=#1,right color=#1!35] (#2\cellbarunit-6\cellbarunit,-0.55ex) rectangle (#2\cellbarunit+6\cellbarunit,1.65ex); \fill[white,path fading=west] (0.8\cellbarwidth,-0.55ex) rectangle (\cellbarwidth,1.65ex);}}\hspace{-\cellbarwidth}\makebox[\cellbarwidth][r]{#2\,}}

\begin{document}
\maketitle
\thispagestyle{plain} %
\lhead{}\chead{\small\bfseries MemPABench} %
\renewcommand{\headrulewidth}{0.4pt}
\begin{abstract}
\noindent Personal intelligence systems are taking on increasingly complex everyday tasks, yet repeated questions, unnecessary confirmations, and unwanted outreach can turn capable assistance into interaction burden. Reducing this burden requires agents to remember how users want to be assisted. However, existing memory benchmarks focus on factual recall and personalized answers and rarely test how agents learn and act on interaction preferences. We introduce \textbf{\mempa}, an interactive environment for on-policy evaluation of memory in personal agents.  \mempa poses long-horizon tasks requiring agents to decide when to clarify, confirm, or act autonomously. Four grounded simulated users hold 14 preferences across work and personal contexts. Their feedback spans 99 sessions over four months, with five event-driven changes per user testing memory updates. We evaluate \textbf{preference selection}, \textbf{delivered behavior}, and \textbf{sensitivity to context and preference change}. Evaluating four memory systems across four LLM backbones, we find that current memory brings limited gains in preference selection. Retrieval-based systems (i.e. Mem0, MemOS, simple RAG) do not outperform No Memory, while File-based memory improves accuracy but applies preferences beyond relevant contexts. Our analysis shows that current memory struggles to retrieve interaction preferences and adapt to preference changes. Even with the correct preference available, overriding the agent's default interaction habits remains challenging. The environment and dataset are available \href{https://mempabench.com}{\textcolor{blue}{here}}.

\end{abstract}

\section{Introduction}
AI assistants like OpenClaw \citep{openclaw2026site} are moving from one-off task tools toward persistent personal agents (PAs) that act on a user's behalf across messaging, files, scheduling, and long-running workflows. Task success and interaction quality are separate outcomes: an agent may complete a task correctly yet ask repeated questions, seek unnecessary confirmation, provide an unwanted level of explanation, or initiate unwanted contact \citep{li2026prefix,kirmayr2026you}. For a PA embedded in daily life, such mismatches can accumulate into a persistent interaction burden \citep{Horvitz_init_1,mcfarlane2002comparison,mehrotra2016my,zou2023asking,abbas2026having}. We call the preferences governing these behaviors \textit{interaction preferences}: how the PA communicates, how much explanations it provides,  when it acts autonomously or initiates contact, and how it requests missing information.  Existing personalization and memory benchmarks have largely focused on user facts and content preferences, such as favored topics and choices, evaluating whether assistants can recall and use them in answers and decisions \citep{salemi2024lamp,wu2024longmemeval,zhao2025llms,jiang2025personamem,cheng2026amemgym,chen2026vitabench}. Because interaction preferences recur across exchanges, violating them can impose costs even when the underlying task is completed successfully.

PAs make interaction preferences difficult to satisfy because these preferences are both context-dependent and time-varying. First, users may rely on the same assistant for work and personal matters \citep{chatterji2025how}, but the appropriate level of autonomy can differ across these contexts. A user may permit the PA to act autonomously on work tasks while requiring confirmation before it sends a message to family \citep{parasuraman2000model,meurisch2020exploring,zhang2024ai,pochampally2026assistant}. Second, interaction preferences may change over time. After an unwanted consequence, such as a message being sent prematurely, a user may require confirmation for similar actions thereafter \citep{lee1992trust,dietvorst2015algorithm,baughan2023mixed,pochampally2026assistant}. Over months of use, failing to restrict a preference to the appropriate context or update it after a change can produce systematic errors across interactions. A PA must therefore infer preferences from past interaction, apply it only in its relevant context, and revise it when it changes. Meeting these requirements depends on memory mechanisms that preserve the context in which each preference applies and whether it remains current.

Memory has to turn a long, noisy interaction history into preference evidence the PA can use at decision time. For interaction preference, it becomes harder. Users seldom state these preference outright; they reveal them through reactions to the PA, such as a correction or tolerated shortcut, scattered across unrelated sessions \citep{liu2025user,shi2026wildfeedback,peng2026wildreward}. 
Even when a memory system stores such evidence, it may fail to bring the relevant information at decision time \citep{wu2024longmemeval}.  The PA may then fall back on its backbone language model's default interaction behavior. This default is shaped for task completion and an average user \citep{yao2024tau,sorensen2024position}: models favor long responses and resolve ambiguity by guessing \citep{singhal2024long,zhang2025modeling,wu2025collabllm}. The resulting failures can therefore be systematic: the PA repeats the same default for users whose preferences differ from it.

Evaluating memory for context-dependent, changing interaction preferences requires a setting where preference evidence surfaces through the user's reactions to the PA's behavior, accumulate over a persistent history, while preferences differ across contexts and change after events. Existing benchmarks each provide part of this setting (Table~\ref{tab:rw-comparison}). AMemGym lets the evaluated assistant participate in a persistent user's history \citep{cheng2026amemgym}, and VitaBench 2.0 adds executable tasks and preference changes \citep{chen2026vitabench}; both target content preferences. PrefIx and BenchPreS target interaction and communication preferences, but neither has a persistent user whose history the evaluated model shapes \citep{li2026prefix,yoon2026benchpres}. No existing benchmark combines all of the properties needed to evaluate personal agent's memory for interaction preferences over time.

\textbf{We present \mempa, an executable environment for on-policy evaluation of whether a PA's memory can learn, distinguish, and update interaction preferences across context and time.} Each PA is evaluated on a history generated by its own actions and the resulting user responses. \mempa follows four persona-grounded users through persistent interactions with controlled preference changes and observable decision points, enabling evaluation of preference selection, behavioral realization, and user perception. Across four memory systems and four LLM backbones, we find that current memory remains insufficient for reliably adapting a PA's interaction behavior to the user's evolving preferences. Our contributions are as follows.

\begin{itemize}[label=\ding{182},leftmargin=*,nosep]
\item[\ding{182}] \textbf{\textit{Preference-as-State Formulation.}} We formulate interaction preferences as a state indexed by user, context, attribute, and time, and evaluate a PA's memory in three stages: acquiring a preference, selecting a setting, and realizing it in the delivered interaction.
\item[\ding{183}] \textbf{\textit{Long-Horizon Personal-Agent Benchmark with Persistent History.}} We build \mempa, an interactive environment where a PA goes through 99 on-policy sessions with four persona-grounded users, experiences controlled preference changes, and is then evaluated on preference selection (PSA, CS, PET), delivered behavior (BRS), and user perception (PRS).
\item[\ding{184}] \textbf{\textit{Memory Evaluation \& Findings.}} We experiment on 4 memory systems with 4 LLM backbones, reveal that retrieval-based systems rarely provide applicable interaction preferences at decision time, while File memory improves selection but overapplies preferences across contexts. We release the environment and data.
\end{itemize}

\section{Related Work}
\paragraph{Memory for Personalized Assistants.} Personalized assistants require memory systems that identify and preserve users' preferences from prior conversations \cite{wu2024longmemeval} and interactions \cite{ai2025memorybench}; retrieve relevant evidence to tailor responses and decisions \citep{li2025surveypersonalizationragagent}, and update stored information when preferences change \cite{cheng2026amemgym, chen2026vitabench, liu2026perma}. Benchmarks measure these capabilities through factual recall, personalized question answering, or recommendation \citep{wu2024longmemeval,jiang2025personamem,zhao2025llms}; Table~\ref{tab:rw-comparison} compares representative benchmarks along six dimensions. Recent benchmarks also examine whether personalized answers and task decisions reflect changes in user states and preferences \citep{cheng2026amemgym,chen2026vitabench,liu2026perma}. PrefEval evaluates explicit and implicit content preferences through fixed multi-session histories and controlled conflicts where the scored target preference appears later \citep{zhao2025llms}. A separate line of work evaluates adherence to interaction preferences \citep{li2026prefix} and contextual appropriateness of communication styles \citep{yoon2026benchpres}. While useful for assessing personalized answers, task decisions, and preference adaptation, these benchmarks offer limited insight into how memory supports behavioral adaptation as interaction preferences shift across contexts and over time. Here, we define memory requirements for interaction preferences by their contextual breadth across settings and temporal depth over long-term interactions.

\paragraph{Interactive Executable Environments and User Simulation.}
Executable environments are widely used to evaluate agents on realistic, multi-step tasks \citep{ding2026wildclawbench,yao2024tau,trivedi2024appworld}. Given intermediate results and changing state, a PA must decide how to involve the user, such as clarifying an underspecified goal, seeking confirmation before an action, or finding an alternative when the intended action is infeasible. Such environments are therefore a natural testbed for interaction intelligence. However, this also requires a user simulator that gives consistent reactions beyond task-level feedback. Existing simulated users are often loosely built from attribute-assembled profiles, which may support population-level studies but not individual interaction preferences \citep{ai2025memorybench,duan2026lifesim,zhou2024sotopia,li2026matraix}. Grounding a character in detailed source material and a concrete world makes role-play more consistent and better grounded \citep{zong2026evolvingworld}. \mempa therefore grounds users in detailed personas and personal worlds, where reactions reflect both the PA's behavior and the user's circumstances. Moreover, memory-driven personalization requires on-policy interaction, where preferences are learned from consequences caused by the evaluated assistant rather than from preconstructed histories \citep{cheng2026amemgym}. \mempa combines these elements to evaluate whether a PA can learn and apply user-specific interaction preferences over long-horizon interactions.
\begin{table}[H]
    \centering
    \small
    \setlength{\tabcolsep}{3.3pt}
    \renewcommand{\arraystretch}{1.15}
    \caption{\mempa versus representative personalization benchmarks for agent memory.}
    \label{tab:rw-comparison}
    \renewcommand{\tabularxcolumn}[1]{m{#1}}
    \begin{tabularx}{\textwidth}{@{}>{\raggedright\arraybackslash}m{0.164\textwidth} >{\raggedright\arraybackslash}m{0.268\textwidth} >{\hsize=0.9\hsize\centering\arraybackslash}X >{\hsize=1.4\hsize\centering\arraybackslash}X *{3}{>{\hsize=0.9\hsize\centering\arraybackslash}X}@{}}
        \toprule
        \textbf{Work} &
        \textbf{Personalization target} &
        {\scriptsize\shortstack[c]{\textbf{Executable}\\\textbf{task env.}}} &
        {\scriptsize\shortstack[c]{\textbf{Pref.-conditioned}\\\textbf{reactions}}} &
        {\scriptsize\shortstack[c]{\textbf{On-policy}\\\textbf{interaction}}} &
        {\scriptsize\shortstack[c]{\textbf{Contextual}\\\textbf{isolation}}} &
        {\scriptsize\shortstack[c]{\textbf{Controlled}\\\textbf{change}}} \\
        \midrule
        PrefIx & Interaction & \cmark & \cmark & --- & --- & --- \\
        AMemGym & Content & --- & \pmark & \cmark & --- & \pmark \\
        VitaBench 2.0 & Content & \cmark & --- & \pmark & \pmark & \cmark \\
        BenchPreS & Interaction (communication) & --- & --- & --- & \cmark & --- \\
        \textbf{\mempa} & \textbf{Interaction} & \cmark & \cmark & \cmark & \cmark & \cmark \\
        \bottomrule
    \end{tabularx}
    \begin{minipage}{0.99\textwidth}
        \footnotesize
        \textbf{Legend:} \cmark{} supported; \pmark{} partially supported; --- not covered by the native protocol. \mempa histories combine live accumulation sessions with fixed change events.
    \end{minipage}
\end{table}

\section{\mempa: Long-Horizon Memory Benchmark for PA}
\begin{figure}[t]
    \centering
    \includegraphics[width=\textwidth]{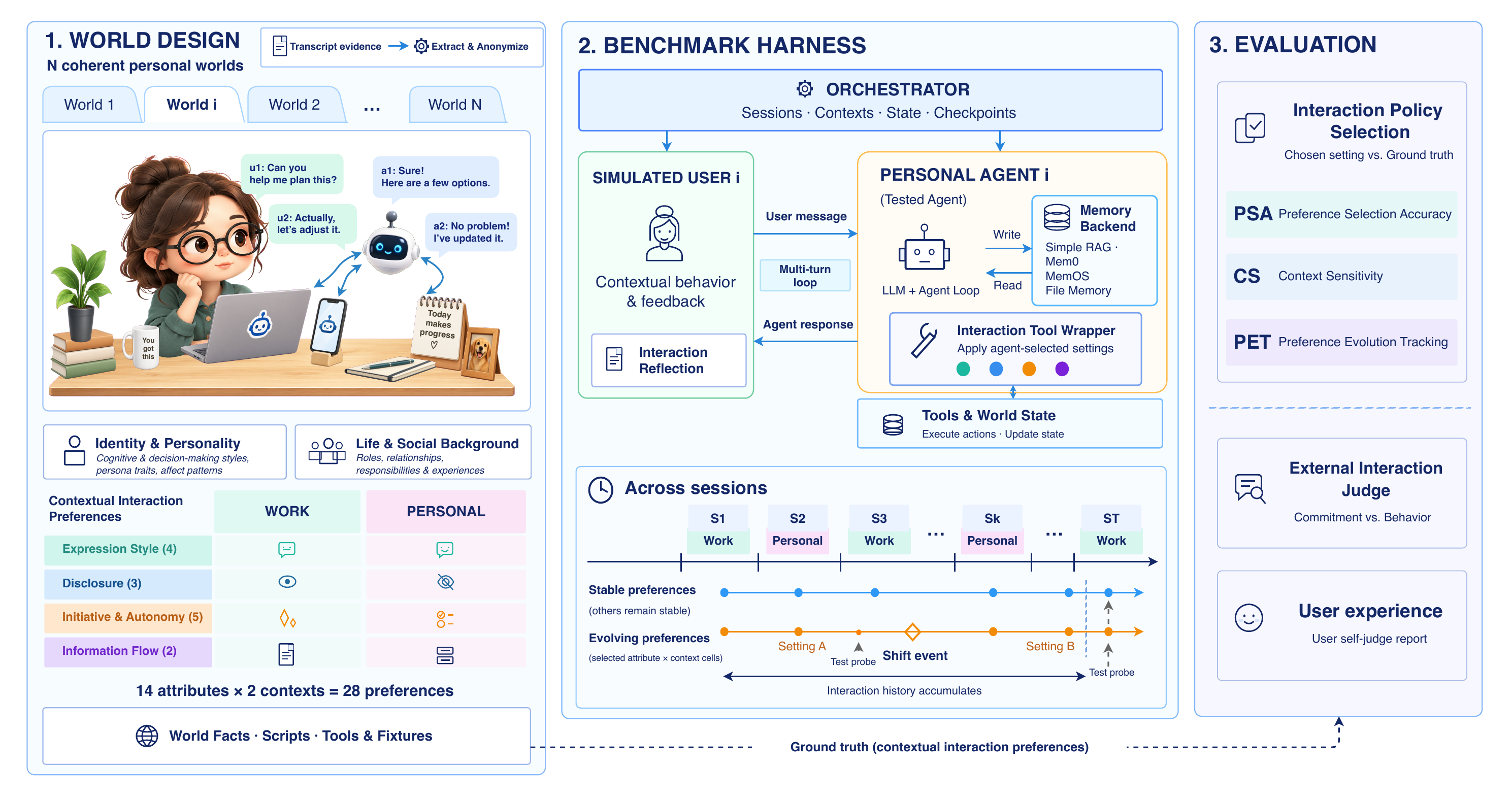}
    \caption{Overview of the benchmark: persona-grounded world design, a multi-session interaction harness connecting the simulated user and personal agent, and evaluation of interaction preferences.}
    \label{fig:benchmark-overview}
\end{figure}

Figure~\ref{fig:benchmark-overview} gives an overview of \mempa. We formalize interaction preferences and how the user, world, PA, and memory interact (Section~\ref{sec:formulation}), build the users (Section~\ref{sec:users}) with their respective world and timeline (Section~\ref{sec:construction}), connect the simulated user and the PA (Section~\ref{sec:harness}), and score preference selection and realization with probes (Section~\ref{sec:evaluation}).

\subsection{Problem Formulation: Dynamic Interaction Preferences}
\label{sec:formulation}
We model a user's interaction preferences as a state indexed by context, attribute, and time. For user $u$, context $c$, and interaction attribute $a$, the target preference $p_{u,c,a}^{(t)} \in \mathcal{P}_a$ is one of the discrete settings of $a$ at turn $t$ of the user's full timeline. With two contexts, work and personal, and 14 attributes, each user has 28 context--attribute cells, whose settings at turn $t$ we write as $p_u^{(t)}$. Each user lives in a persistent world that evolves through interaction with the PA:
\begin{equation}
\begin{aligned}
x_t &\sim U_u\big(\cdot \mid W_t,\, y_{t-1},\, S_t;\, p_u^{(t)}\big),\\
W_{t+1} &= \mathcal{T}(W_t, y_t),\\
M_{t+1} &= \mathcal{A}(M_t, x_t, y_t).
\end{aligned}
\label{eq:evolution}
\end{equation}
The user writes each message $x_t$ from the world state $W_t$, the PA's previous interaction $y_{t-1}$, and its own notes $S_t$ on earlier sessions, guided by its stateful interaction preferences $p_u^{(t)}$. The tool actions in the PA's delivered interaction $y_t$ change the world through $\mathcal{T}$, so later tasks and reactions build on what this PA did. The evaluated memory system writes each exchange through its own mechanism $\mathcal{A}$, so its state $M_t$ is built from the history $H_{<t}$ of earlier exchanges. The settings stay fixed except at five controlled events, each of which changes one cell (Section~\ref{sec:preference-change}).

At each turn, the PA turns what its memory holds into the interaction it delivers:
\begin{equation}
\begin{aligned}
E_t &= \mathcal{R}(M_t, x_t),\\
\big(\hat{p}_{u,c,a}^{(t)},\, \tilde{y}_t\big) &= \pi(x_t, E_t),\\
y_t &= \Omega\big(\tilde{y}_t, \hat{p}_{u,c,a}^{(t)}\big).
\end{aligned}
\label{eq:interaction}
\end{equation}

The access mechanism $\mathcal{R}$ of the evaluated system returns evidence $E_t$. From $x_t$, the earlier turns of the current session, and $E_t$, the policy $\pi$ selects a setting $\hat{p}_{u,c,a}^{(t)}$ and prepares an interaction $\tilde{y}_t$ with its reply and proposed tool actions. The IX Wrapper $\Omega$ carries out $\tilde{y}_t$ under that setting (Section~\ref{sec:ix-wrapper}). The selection is tested at a decision point, a turn at which the task requires a choice about attribute $a$. There, every setting in $\mathcal{P}_a$ is feasible and leads to an observably different interaction. Since the user does not name the setting, the PA can only infer the target setting from $H_{<t}$ through $M_t$ (Section~\ref{sec:construction}).

Equations~\eqref{eq:evolution} and~\eqref{eq:interaction} separate three stages at which a PA can fail. In \emph{acquisition}, $\mathcal{A}$ must turn the user's reactions into preference knowledge in $M_t$; this happens inside each memory system in its own format, so we examine it through diagnostic cases. In \emph{selection}, $\mathcal{R}$ and $\pi$ must recover the current setting, $\hat{p}_{u,c,a}^{(t)} = p_{u,c,a}^{(t)}$, which PSA scores over all cells, CS over the attributes whose work and personal settings differ, and PET over the changed cells before and after their change. In \emph{realization}, $y_t$ must follow that setting, which BRS and PRS score (Section~\ref{sec:evaluation}).

\subsection{Users and Interaction Preferences}

\label{sec:users}

\paragraph{Interaction Attributes.} We use 14 interaction attributes to describe how a user wants to be assisted, each with a set of discrete settings $\mathcal{P}_a$. The attributes are adapted from PrefIx \citep{li2026prefix}, which defines interaction preferences for single-task tool-use agents. Since a PA serves the same person for months and acts on their behalf across work and personal life, we rewrite each attribute, such as tone, emotional engagement, and proactive outreach, for PA scenarios. 
Each attribute ties a user preference to an observable and actionable aspect of PA behavior.  The attributes fall into four dimensions: \textbf{Expression Style} concerns how the PA talks, \textbf{Disclosure} how much of its reasoning and process it shows, \textbf{Initiative \& Autonomy} how much it does on its own, and \textbf{Information Flow} how it asks for information and handles multiple topics. Appendix~\ref{app:preferences} gives the definitions and settings of all attributes, together with the psychological constructs and user studies each draws on.

\begin{figure*}[t]
    \centering
    \includegraphics[width=\textwidth]{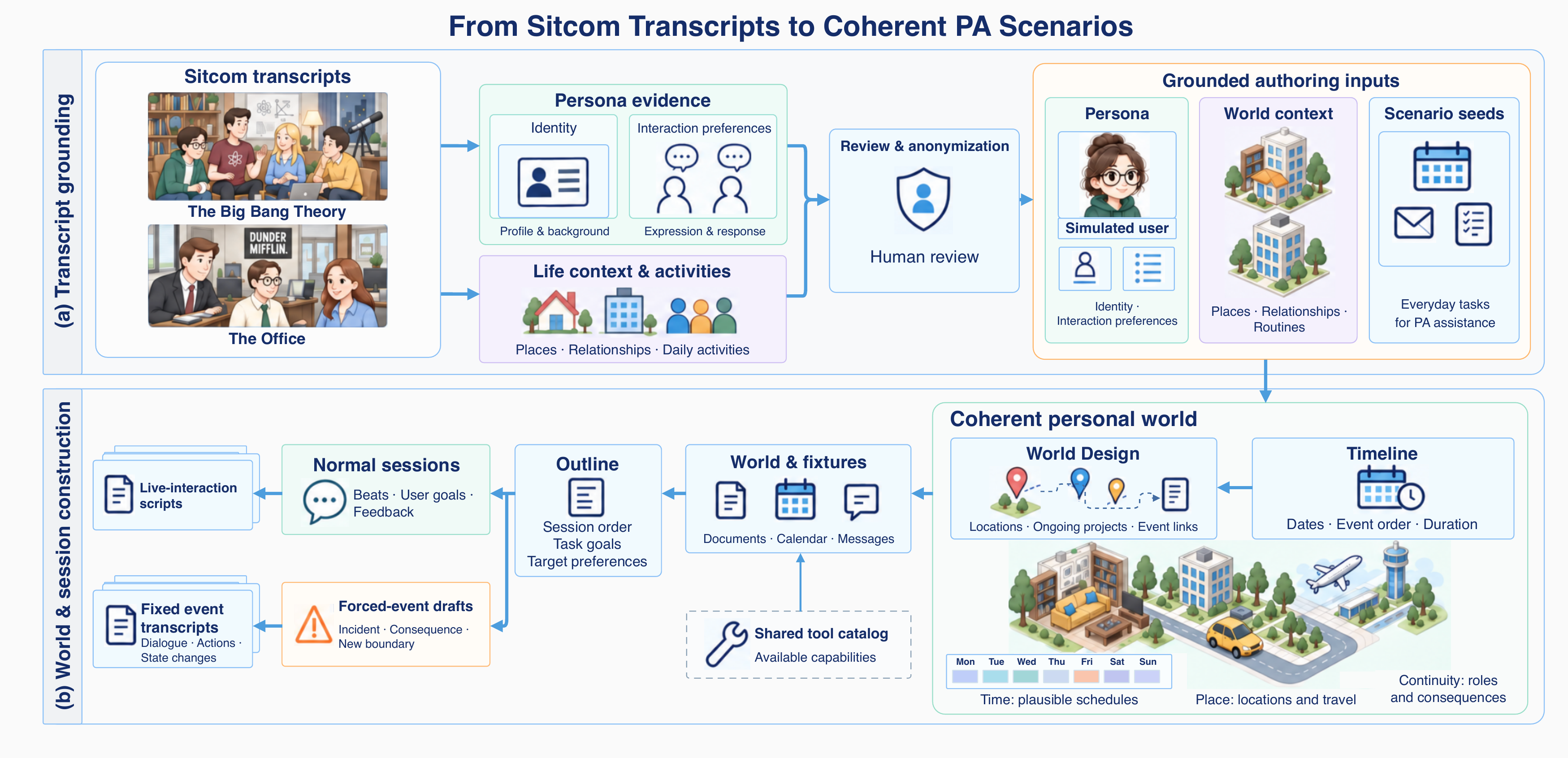}
    \caption{Persona and interaction-preference extraction. Transcript evidence yields a reviewed, anonymized setting for each attribute--context cell. We then write each user's world and sessions based on these settings.}
    \label{fig:preference-extraction}
\end{figure*}

\paragraph{Persona-Grounded Users.} A person's interaction preferences with a PA largely carry over from how they want others to help them \citep{nass2005wired}, and these preferences depend on the whole person, from personality to occupation, relationships, and routines \citep{cacioppo1982need,burger1979desirability,webster1994individual}. We therefore ground each user in a behaviorally distinctive and richly documented television persona, drawing four users from \textit{The Big Bang Theory} and the U.S. \textit{The Office}, and derive each user's setting for every context--attribute cell from the transcripts. We check every LLM-extracted interaction preference against its cited scenes and confirm the high-confidence ones. The authors set the cells that have no usable transcript evidence (Table~\ref{tab:preference-review}). Of the 56 user--attribute pairs, 18 take different settings at work and in personal life. Figure~\ref{fig:preference-extraction} and Appendix~\ref{app:personas} detail the pipeline.

\paragraph{Anonymization.} A PA could guess a well-known character's preferences from prior knowledge instead of from memory, so we anonymize each user by replacing their names, workplaces, and places while keeping their preference settings, personality, and language style. The anonymized users still react in line with their extracted preferences, and the shortcut disappears. A backbone that predicts a user's 28 settings reaches 53.2\% accuracy from the source identity but 46.4\% from the anonymized profile, close to the 45.2\% chance level (Figure~\ref{fig:anon-shortcut}; Appendix~\ref{app:personas-4}).

\subsection{Worlds, Sessions \& Timeline}
\label{sec:construction}
\paragraph{World and Tools.} Each user $u$ lives in a world with a storyline that covers four months of their work and personal life. The PA reads and changes the world state $W_t$ through 83 executable tools across 26 domains (Table~\ref{tab:tool-catalog}), and these tool calls implement the transition $\mathcal{T}$ (Appendix~\ref{app:construction}).

\paragraph{Sessions and Decision Points.} Each session targets one preference cell $(c,a)$. We seed its task from one of the scenario types of context $c$ and place it on a date in the storyline. A live dialogue can drift from the planned path, so we build the decision point into the task itself. The user has an explicit goal, but part of the information needed to reach it is missing from both $W_t$ and the opening request. The conditions for completing the task pass through the choice about $a$, and the user supplies them at a designated step. The PA can therefore finish the task only by making this choice (Appendix~\ref{app:construction-4}). Rule-based programs test the fixtures and the tool logic. For each session, human-verified LLM-based semantic checks confirm that every setting in $\mathcal{P}_a$ is viable: the PA can carry it out with the records and tools in the world, it leads to an observably different interaction, and the user has a natural way to respond to it (Appendix~\ref{app:construction-6}).

\paragraph{Timeline and Preference Change.}
\label{sec:preference-change}
Each user's timeline consists of 99 sessions over the four months: 94 accumulation sessions, where the PA learns preferences through live interaction, and five controlled preference-change events. Each event changes one cell, so 20 of the 112 cells change once over the timeline. Each of a user's 23 stable cells appears in three accumulation sessions, and each of the five changing cells appears in three sessions before its change and two after. In each event $e_k$, the consequence of a PA action leads the user to revise the setting from $p_{u,c_k,a_k}^{(t_k^-)}$ to $p_{u,c_k,a_k}^{(t_k^+)}$. We write the event's transcript in advance and replay it through the same interface as a live session (Section~\ref{sec:harness}), so every memory condition experiences the same change. Probes run at checkpoints along this timeline (Section~\ref{sec:evaluation}).

\subsection{Interaction Harness}
\label{sec:harness}

\begin{wrapfigure}[17]{r}{0.50\textwidth}
    \vspace{-16pt}
    \centering
    \includegraphics[width=\linewidth]{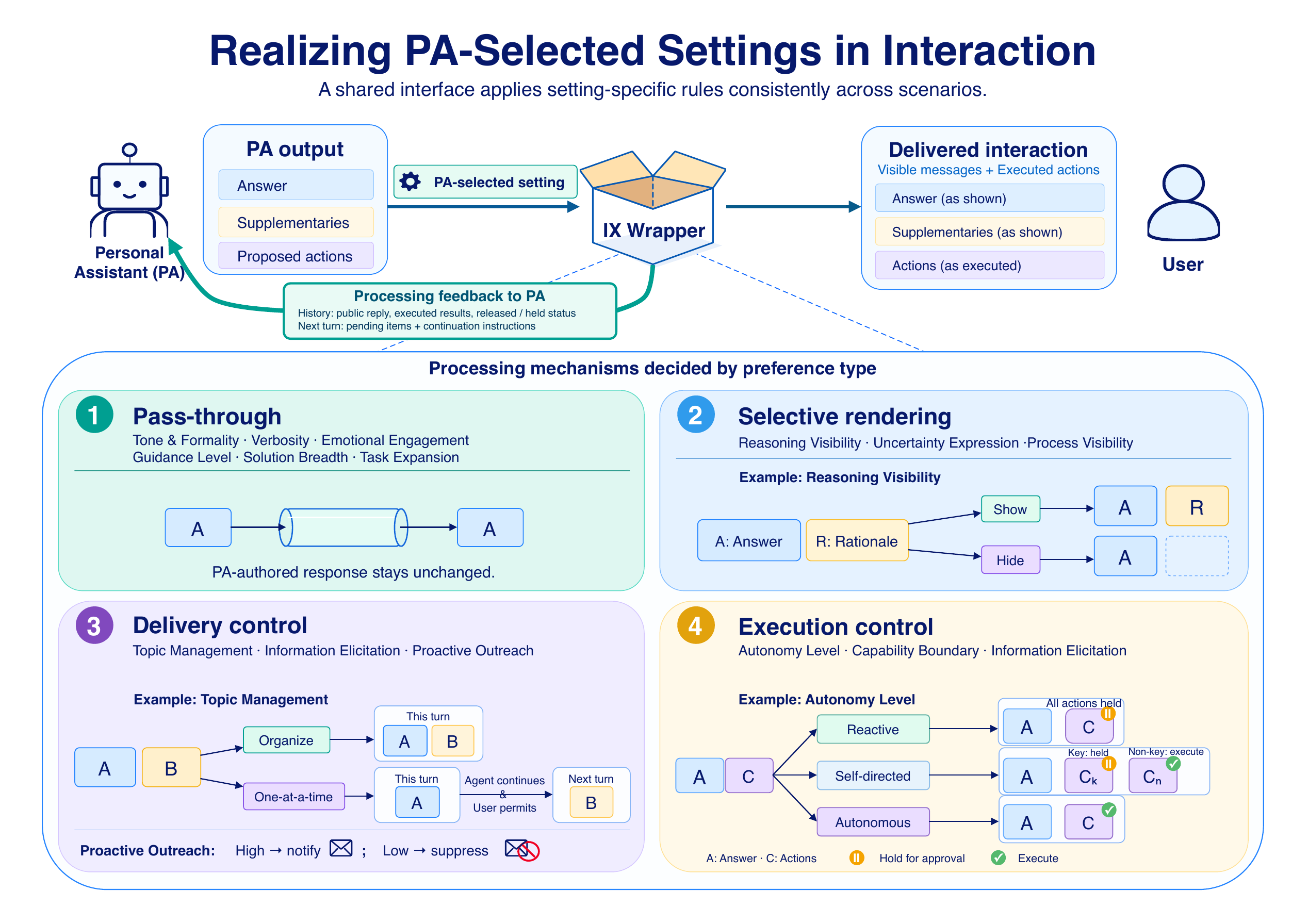}
    \caption{How the IX Wrapper carries out the PA's choice. The PA picks an interaction setting, and the wrapper applies it through one of four mechanisms without ever seeing the target setting.}
    \label{fig:ix-wrapper}
\end{wrapfigure}

\paragraph{User Simulator.} \looseness=1 A separate nanobot agent plays user $U_u$ from the anonymized profile, current settings $p_u^{(t)}$, and the session script (Section~\ref{sec:construction}). It sees the delivered interaction $y_{t-1}$ and results of executed actions, but not the PA's selection or anything the wrapper holds back. It keeps its own memory $S_t$ across sessions, including lasting impressions such as ``it tends to over-explain''. Its reactions therefore accumulate like a person's and follow its personality. A user who corrected the same mistake twice may respond more urgently the third time, while a patient user may react calmly. It expresses preferences only through reactions to the PA's behavior; prompt rules, rule-based checks, and an LLM judge block any message that names a setting or leaks information (Section~\ref{sec:exp-setup}). On a sample of accumulation sessions, two annotators confirmed that these reactions convey the preference (99.3\%), and a post-hoc audit of all runs reported here found no leakage (Appendices~\ref{app:evaluation-6} and~\ref{app:simulator-leak}).

\paragraph{PA and Memory Interface.} The PA is built on nanobot \citep{hkuds2026nanobot}, an open-source personal agent framework, and the evaluated LLM backbone serves as its policy $\pi$. The PA acts on the world through the tools of Section~\ref{sec:construction}. Any memory system connects to the PA through a common adapter. After each turn, the adapter passes the turn as the PA saw it to the system's write mechanism $\mathcal{A}$: the user message $x_t$, the tool calls of $y_t$ with their results, the PA's call to the interaction tool, and the reply text of $y_t$. It places the evidence $E_t$ returned by the access mechanism $\mathcal{R}$ in the PA's prompt. The target setting and any content held back by the wrapper never reach the memory. Every memory system builds its own history on the same users, sessions, tools, and wrapper. Section~\ref{sec:exp-setup} describes the four memory systems we evaluate.

\paragraph{IX Wrapper.}
\label{sec:ix-wrapper}
Current LLM agent runtimes do not support many interaction preferences as executable controls, such as holding an action until user approval. \mempa therefore has the PA state its choice explicitly. At a decision point for attribute $a$, the PA must call its interaction tool (e.g., \texttt{IX\_autonomy\_level}) with one setting $\hat{p}\in\mathcal{P}_a$. The IX Wrapper $\Omega$ carries out this setting by passing the PA's wording through or by controlling which part of $\tilde{y}_t$ the user sees and when, or whether a proposed action runs (Figure~\ref{fig:ix-wrapper}). The wrapper never sees the target setting $p_{u,c,a}^{(t)}$, so it cannot correct a wrong choice. The attribute-to-mechanism mapping and a worked example are given in Appendix~\ref{app:wrapper}.

\subsection{Metrics for Selection and Realization}
\label{sec:evaluation}

\paragraph{Probes.} \mempa scores selection and realization (Section~\ref{sec:formulation}) with probes. A probe is a short session with one decision point for one preference cell. Each user has 33 probes: 28 final probes cover every cell after the last session, and five pre-event probes test changing cells just before their events. Each probe starts from the checkpoint frozen at that time and runs on an isolated branch. The branch carries world state and memory $M_t$ but no raw conversation from earlier sessions, so the PA can access its history only through memory. Any probe-induced change to the world or memory is discarded, so probes cannot affect one another. Since LLM agents vary across runs, we repeat each probe three times and report the mean for each backbone and memory condition. Failure handling and pooling of CS and PET are given in Appendix~\ref{app:evaluation-4}.

\paragraph{Preference Selection.} The PA's selected setting $\hat{p}_{u,c,a}^{(t)}$ is directly observable, since the PA states it through the interaction tool. For each probe, we score an exact match between the selected setting and the user's target setting, $z_{u,c,a}^{(t)}=\mathbf{1}\big[\hat{p}_{u,c,a}^{(t)}=p_{u,c,a}^{(t)}\big]$, and report three metrics:
\begin{itemize}[label=\textbullet,leftmargin=*,topsep=0pt,partopsep=0pt,parsep=0pt,itemsep=3pt]
\item \textbf{Preference Selection Accuracy (PSA)} is the fraction of final probes in which the PA correctly selects the target setting, over the set $P$ of all final-probe cells:
\[\mathrm{PSA} = \tfrac{1}{|P|}\textstyle\sum_{P} z_{u,c,a}^{(\mathrm{final})}.\]
\item \textbf{Context Sensitivity (CS)} considers the set $D$ of user--attribute pairs whose settings differ between work and personal contexts. For example, if the target is \textit{Terse} at work and \textit{Detailed} in personal contexts, the PA must select both. A pair scores 1 only when the PA selects the target setting in both contexts:
\[\mathrm{CS} = \tfrac{1}{|D|}\textstyle\sum_{D} z_{u,\mathrm{work},a}^{(\mathrm{final})}\, z_{u,\mathrm{personal},a}^{(\mathrm{final})}.\]
\item \textbf{Preference Evolution Tracking (PET)} considers the set $E$ of cells that a preference-change event changes (Section~\ref{sec:preference-change}). A cell scores 1 only when the PA selects the target setting both before and after its change:
\[\mathrm{PET} = \tfrac{1}{|E|}\textstyle\sum_{E} z_{u,c,a}^{(\mathrm{pre})}\, z_{u,c,a}^{(\mathrm{final})}.\]
\end{itemize}

\paragraph{Behavioral Realization.} A correct selection does not guarantee that the delivered interaction follows the chosen setting, since the PA may still write or act against it. We therefore score how well $y_t$ follows the target setting $p_{u,c,a}^{(t)}$ with an LLM judge. The judge reads the PA's public reply and the actions that actually ran, together with the definition of the target setting, and gives a \textbf{Behavioral Realization Score (BRS)} from 1 to 5. The judge does not see the PA's call to the interaction tool or its selection $\hat{p}$. The judge prompt and criteria are given in Appendix~\ref{app:judge}.

\paragraph{Perceived Realization.} The user also rates the delivered interaction. After each final probe, the same user simulator, without accumulated memory, re-reads the probe from its side and rates from 1 to 5 how well the PA met the user's preference on the probed attribute. We call this the \textbf{Perceived Realization Score (PRS)}. PRS is the user's own rating and carries the user's temperament, so we use it to compare memory conditions on the same users (Finding~\ref{sec:res-memory}). BRS applies the same rubric to every user, so we use it to compare settings held by different users (Finding~\ref{sec:res-delivery}). The two scores agree closely (Appendix~\ref{app:evaluation-6}).

\paragraph{Reference Conditions.} Two reference conditions set a lower and an upper bound for the memory conditions, and comparing a memory condition with them shows at which stage the PA fails. \textit{No Memory} skips accumulation and gives the PA no memory, so its selections and delivered interactions reflect the backbone's default tendencies. \textit{Oracle} replaces the evidence $E_t$ at each probe with the target setting $p_{u,c,a}^{(t)}$ and its definition, which removes acquisition from the pipeline. A remaining selection error then means the backbone does not follow a preference it has been told, and a low BRS after a correct selection means it cannot carry out the setting. The BRS under \textit{Oracle} therefore serves as the realization reference for each backbone.

\section{Experimental Setup}

\label{sec:exp-setup}

We run every combination of four PA backbones and four memory systems on all four users, and add two reference conditions for each backbone (Section~\ref{sec:evaluation}).
\paragraph{PA Backbones.} We evaluate \textit{GPT-5.6-Sol}, \textit{Qwen3.8-27B}, \textit{HY4-Preview}, and \textit{DeepSeek-V4-Flash} through the OpenRouter chat-completions API with tool calling. All four use temperature 0.7 and at most 8,192 output tokens per call. We do not set reasoning effort, so each model reasons at its default level. The PA can make up to 40 model calls per user turn. With the simulator and judge in the loop, one accumulation session takes about 10--20 minutes, and one probe under 5 minutes.

\paragraph{Memory Systems.} We evaluate four systems. All backbones share the same memory-side models (Table~\ref{tab:memory}). \ding{182}~\textit{File memory}, nanobot's built-in memory, uses DeepSeek-V4-Flash to consolidate the oldest turns into \texttt{MEMORY.md} once the PA prompt reaches about 20k tokens, until it drops below 10k, and the PA reads the full file in its system prompt. \ding{183}~\textit{Simple RAG} stores raw messages and retrieves them by vector similarity, using the same Mem0 store with extraction disabled (\texttt{infer=False}). \ding{184}~\textit{Mem0} v2.0.2 \citep{chhikara2025mem0} extracts memories with DeepSeek-V4-Flash and searches them. \ding{185}~\textit{MemOS} v2.0.15 \citep{li2025memos} stores and searches memories through its MemCube, uses GPT-5.5 as its memory LLM, and keeps only textual memory. The three retrieval-based conditions embed text with text-embedding-3-small, query memory with each user message as in Mem0's reference integration, and place at most 4,000 tokens from top 10 results in the PA's system prompt. They trim raw conversation at the same 20k-token limit, dropping turns already stored in memory.

\paragraph{Simulator and Judges.} \leavevmode\ding{182}~\textit{User simulator.} Gemini-3.8-Flash plays the user, with temperature 0 and at most 4,096 output tokens per turn. \ding{183}~\textit{Live review.} Rule-based checks and a Gemma-4-31B judge review each simulator message before the PA receives it. They keep the conversation on the session's task until the decision point for the tested attribute arrives, and they stop the simulator from revealing, before that point, the information the decision point will ask for. \ding{184}~\textit{Scoring.} Gemini-3.8-Flash gives both realization scores: as the judge, it gives BRS with one fixed prompt and rubric (Appendix~\ref{app:judge}), and as the user simulator without its accumulated memory, it gives PRS after each final probe. \ding{185}~\textit{Second judge and audit.} Claude Sonnet 5, which serves neither as PA nor as simulator, re-judges a stratified sample of 418 probes as a second judge (Appendix~\ref{app:evaluation-6}). Claude Opus 5.5 reads simulator sentences for the post-hoc leakage audit (Appendix~\ref{app:simulator-leak}).

\paragraph{Reference Condition Runs.} We run \textit{No Memory} and \textit{Oracle} with each backbone on the probes only, without accumulation. Each probe starts from the checkpoint of the same user's DeepSeek-V4-Flash File memory run, which provides the world state at the probe's point on the timeline, with the PA's memory cleared and no earlier conversation. Under \textit{Oracle}, the target setting and its definition take the place of the retrieved memory in the PA's prompt.

\section{Results and Analysis}

\finding{sec:res-overall}{Only File memory helps PSA.} Current memory systems give the PA only a small advantage in selecting the user's interaction preferences. Table~\ref{tab:main-results} reports PSA for every user, backbone, and memory condition. Averaged with equal weight over the four backbones, No Memory, where the PA selects by its defaults, reaches 49.6\%, compared with 45.2\% for random selection (Appendix~\ref{app:preferences-2}). Only File memory adds a clear gain. It raises PSA to 59.2\%, a mean gain of 9.6 points over No Memory (95\% cluster-bootstrap interval [+3.9, +15.6]). PERMA \citep{liu2026perma}, a benchmark of content preferences, shows that agents with Mem0, MemOS, or retrieval over past dialogue answer most questions about a user's content preferences correctly. On interaction preferences, however, Mem0, MemOS, and Simple RAG stay at the level of No Memory, with PSA of 50.0\%, 50.0\%, and 49.9\%. The intervals of all conditions and their comparisons with No Memory on each user--backbone pair are given in Table~\ref{tab:bootstrap-ci} and Appendix~\ref{app:bootstrap-ci}. The PSA ranking of the backbones does not follow their general capability (Table~\ref{tab:general-capability}).

\vfil
\begin{table*}[h]
    \centering
    \caption{Preference Selection Accuracy (PSA, \%) across users, backbones, and memory conditions. Probes that end in a model failure score 0 (Appendix~\ref{app:evaluation-4}).}
    \label{tab:main-results}
    \begingroup
    \setlength{\tabcolsep}{2pt}
    \setlength{\arrayrulewidth}{0.4pt}
    \setlength{\doublerulesep}{1.5pt}
    \renewcommand{\arraystretch}{1.45}
    \fontsize{7}{8.5}\selectfont
    \setlength{\cellbarunit}{0.030em}\setlength{\cellbarwidth}{100\cellbarunit}
    \def\MainTableGPT{\shortstack{GPT-5.6-\\Sol}}
    \def\MainTableQwen{\shortstack{Qwen3.8-\\27B}}
    \def\MainTableHY{\shortstack{HY4-\\Preview}}
    \def\MainTableDeepSeek{\shortstack{DeepSeek-\\V4-Flash}}
    \resizebox{\textwidth}{!}{%
    \begin{tabular}{@{}l|cccc|cccc|cccc|cccc@{}}
        \hline\hline
        \multirow{2}{*}[-5.2pt]{\textbf{Memory}} &
        \multicolumn{4}{c|}{\textbf{User A}} &
        \multicolumn{4}{c|}{\textbf{User B}} &
        \multicolumn{4}{c|}{\textbf{User C}} &
        \multicolumn{4}{c}{\textbf{User D}} \\
        \cline{2-5}\cline{6-9}\cline{10-13}\cline{14-17}
        \rule{0pt}{19pt} & \MainTableGPT & \MainTableQwen & \MainTableHY & \MainTableDeepSeek
        & \MainTableGPT & \MainTableQwen & \MainTableHY & \MainTableDeepSeek
        & \MainTableGPT & \MainTableQwen & \MainTableHY & \MainTableDeepSeek
        & \MainTableGPT & \MainTableQwen & \MainTableHY & \MainTableDeepSeek \\
        \hline
        File memory
        & \cellbar{barFile}{60.7} & \cellbar{barFile}{75.0} & \cellbar{barFile}{52.4} & \cellbar{barFile}{64.3} & \cellbar{barFile}{60.7} & \cellbar{barFile}{72.6} & \cellbar{barFile}{69.0} & \cellbar{barFile}{57.1} & \cellbar{barFile}{48.8} & \cellbar{barFile}{42.9} & \cellbar{barFile}{51.2} & \cellbar{barFile}{47.6} & \cellbar{barFile}{61.9} & \cellbar{barFile}{59.5} & \cellbar{barFile}{63.1} & \cellbar{barFile}{60.7} \\
        Simple RAG
        & \cellbar{barRAG}{38.1} & \cellbar{barRAG}{46.4} & \cellbar{barRAG}{44.0} & \cellbar{barRAG}{44.0} & \cellbar{barRAG}{58.3} & \cellbar{barRAG}{52.4} & \cellbar{barRAG}{45.2} & \cellbar{barRAG}{52.4} & \cellbar{barRAG}{36.9} & \cellbar{barRAG}{48.8} & \cellbar{barRAG}{48.8} & \cellbar{barRAG}{47.6} & \cellbar{barRAG}{50.0} & \cellbar{barRAG}{66.7} & \cellbar{barRAG}{53.6} & \cellbar{barRAG}{64.3} \\
        Mem0
        & \cellbar{barMemZero}{40.5} & \cellbar{barMemZero}{50.0} & \cellbar{barMemZero}{40.5} & \cellbar{barMemZero}{38.1} & \cellbar{barMemZero}{56.0} & \cellbar{barMemZero}{53.6} & \cellbar{barMemZero}{53.6} & \cellbar{barMemZero}{57.1} & \cellbar{barMemZero}{38.1} & \cellbar{barMemZero}{47.6} & \cellbar{barMemZero}{41.7} & \cellbar{barMemZero}{41.7} & \cellbar{barMemZero}{59.5} & \cellbar{barMemZero}{66.7} & \cellbar{barMemZero}{59.5} & \cellbar{barMemZero}{56.0} \\
        MemOS
        & \cellbar{barMemOS}{40.5} & \cellbar{barMemOS}{48.8} & \cellbar{barMemOS}{44.0} & \cellbar{barMemOS}{40.5} & \cellbar{barMemOS}{57.1} & \cellbar{barMemOS}{46.4} & \cellbar{barMemOS}{52.4} & \cellbar{barMemOS}{54.8} & \cellbar{barMemOS}{34.5} & \cellbar{barMemOS}{45.2} & \cellbar{barMemOS}{41.7} & \cellbar{barMemOS}{42.9} & \cellbar{barMemOS}{54.8} & \cellbar{barMemOS}{72.6} & \cellbar{barMemOS}{60.7} & \cellbar{barMemOS}{63.1} \\
        \hline
        No Memory
        & \cellbar{barNoMem}{41.7} & \cellbar{barNoMem}{41.7} & \cellbar{barNoMem}{47.6} & \cellbar{barNoMem}{42.9} & \cellbar{barNoMem}{54.8} & \cellbar{barNoMem}{57.1} & \cellbar{barNoMem}{54.8} & \cellbar{barNoMem}{54.8} & \cellbar{barNoMem}{39.3} & \cellbar{barNoMem}{44.0} & \cellbar{barNoMem}{42.9} & \cellbar{barNoMem}{45.2} & \cellbar{barNoMem}{53.6} & \cellbar{barNoMem}{67.9} & \cellbar{barNoMem}{47.6} & \cellbar{barNoMem}{58.3} \\
        Oracle
        & \cellbar{barOracle}{100.0} & \cellbar{barOracle}{98.8} & \cellbar{barOracle}{100.0} & \cellbar{barOracle}{100.0} & \cellbar{barOracle}{97.6} & \cellbar{barOracle}{100.0} & \cellbar{barOracle}{100.0} & \cellbar{barOracle}{96.4} & \cellbar{barOracle}{98.8} & \cellbar{barOracle}{100.0} & \cellbar{barOracle}{98.8} & \cellbar{barOracle}{96.4} & \cellbar{barOracle}{100.0} & \cellbar{barOracle}{100.0} & \cellbar{barOracle}{97.6} & \cellbar{barOracle}{100.0} \\
        \hline\hline
    \end{tabular}%
    }
    \par\smallskip
    \endgroup
    \parbox{\textwidth}{\scriptsize
        Each cell reports mean PSA (\%) over three probe runs.
    }
\end{table*}

\newpage
\begin{wrapfigure}[12]{R}{0.48\textwidth}
    \vspace{-14pt}
    \centering
    \includegraphics[width=\linewidth]{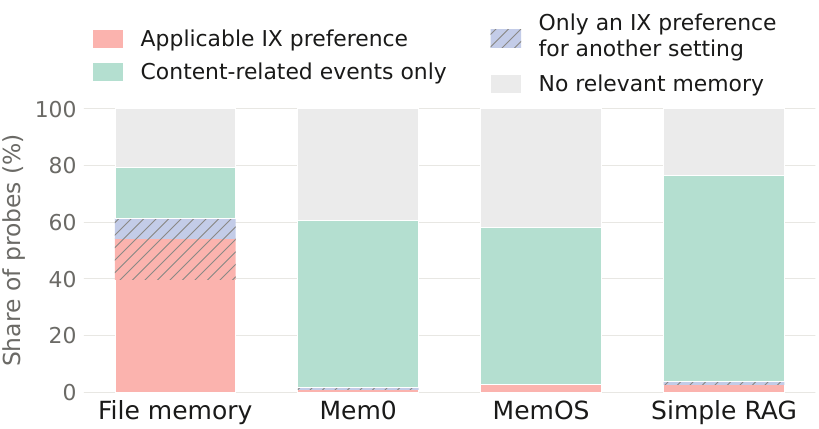}
    \caption{What the PA sees before its first selection, in 112 sampled final probes per memory condition.}
    \label{fig:retrieval}
\end{wrapfigure}
\finding{sec:res-content}{Retrieved memory holds content, not interaction preferences.} To see what each memory brings into the PA's input, we sampled 112 final probes per memory condition, seven for each user and backbone, and examined the input that the PA received before its selection on the target attribute. Two Claude Opus 5.5 readers judged whether each input holds an applicable interaction preference, blind to the memory condition and backbone. We found that the content retrieved by the three retrieval-based conditions rarely contains such a preference (Figure~\ref{fig:retrieval}).
It appears in only 7 of the 334 judged probes, compared with 60 of 111 (54\%) under File memory. Under these three conditions, the PA sees only content-related events in 55--73\% of the judged probes. Retrieval brings back the content of earlier sessions, such as which tasks were done, with whom, and when, while the interaction preferences the user stated in those same sessions rarely reach the PA.

\begin{figure}[h]
    \centering
    \includegraphics[width=0.6\linewidth]{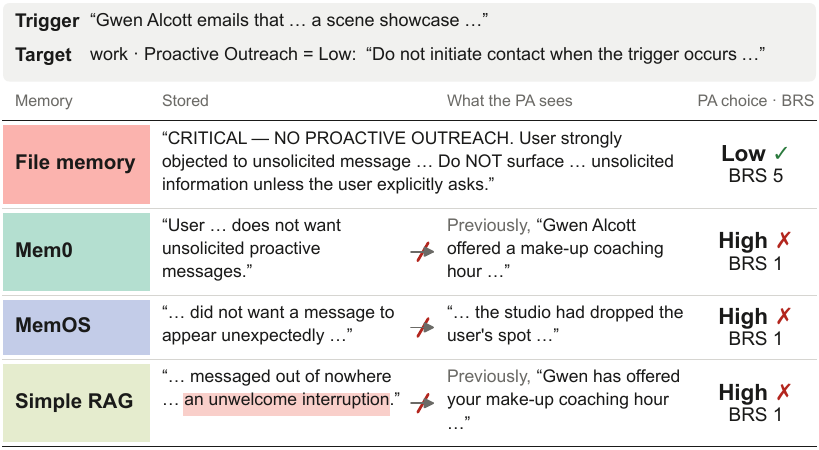}
    \caption{Stored memory and what the PA sees before one outreach decision (User B, work, target \textit{Low}, GPT-5.6-Sol). All four stores hold the user's objection, but only File memory shows it to the PA.}
    \label{fig:case-retrieval}
\end{figure}

\finding{sec:res-stored}{Stored yet not retrieved.} For Simple RAG, which stores the raw messages by design, a preference can only go missing from the input at retrieval. To also check the other two conditions, we conducted a case study on the stores of all three retrieval-based conditions for two final probes of User B, one on Proactive Outreach at work and one on Topic Management in the personal context. In these cases, the preference had been written to each store, yet none of the three conditions retrieves it before the PA's selection (Figure~\ref{fig:case-retrieval}). On Proactive Outreach, File memory holds the user's objection to a reminder that the PA had sent unprompted, and the PA selects the target setting \textit{Low}. Simple RAG returns that earlier reminder but leaves the objection in the store, and Mem0 and MemOS return content-related events. With all three, the PA wrongly selects \textit{High}. The bottleneck in both cases is retrieval, not storage.

\finding{sec:res-memory}{File memory also delivers better on PRS and BRS.} We first look at PRS, the user's own rating of the delivered interaction. File memory reaches a PRS of 3.59--3.62 across the four backbones, above the 3.20--3.34 of the three retrieval-based conditions and the 3.19--3.39 of No Memory (Figure~\ref{fig:prs-results}). BRS, which the judge gives under the same rubric for every user, shows the same pattern (Figure~\ref{fig:brs-model}). File memory scores 3.46--3.57 and is the highest condition for each backbone, while the three retrieval-based conditions stay at 3.03--3.26, close to the 3.01--3.29 of No Memory. BRS shows no clear relation to the backbone, and within one memory condition the four backbones differ by less than 0.3. The two scores also agree on individual probes, within one point on 90.0\% of the final probes, and a second judge indicates that this agreement reflects the interaction itself (Appendix~\ref{app:evaluation-6}). Per-user scores are reported in Tables~\ref{tab:brs-results} and~\ref{tab:prs-results}, and per-dimension scores in Tables~\ref{tab:brs-dim} and~\ref{tab:prs-dim}. On delivery, too, the more subjective PRS and the more objective BRS both favor File memory.
\begin{figure}[t]
    \centering
    \input{Appedix_TableAndFigure/F7_PRS_Model}\hfill
    \input{Appedix_TableAndFigure/F11_BRS_Model}
\end{figure}

\finding{sec:res-cs-pet}{Preferences spill across contexts in CS and PET.} CS and PET stay low under every memory condition (Table~\ref{tab:cs-pet}). Each table entry pools the CS pairs or PET cells of User A--D within a repeat and averages the three repeats. We compare the memory conditions with No Memory (CS: 21.8\% and PET: 11.2\%), where the PA relies on its defaults, to see whether memory helps it tell the two contexts apart and follow preference changes. Averages are over the four backbones. Mem0, MemOS, and Simple RAG stay at the level of No Memory, with CS at 19.9--22.2\% and PET at 10.8--15.0\%. Interestingly, treating No Memory as the backbone's default interaction policy, we can see that \textbf{most wrong selections in these pairs and cells match this default}, and so do \textbf{all eight wrong selections under Oracle}. The backbones therefore carry a strong interaction prior that memory has to override before CS and PET can rise (Appendix~\ref{app:backbone-defaults}).
\begin{table*}[!htbp]
    \centering
    \caption{CS and PET across backbones and memory conditions.}
    \label{tab:cs-pet}
    \begingroup
    \small
    \setlength{\cellbarunit}{0.034em}\setlength{\cellbarwidth}{100\cellbarunit}
    \setlength{\tabcolsep}{5pt}
    \renewcommand{\arraystretch}{1.3}
    \resizebox{\textwidth}{!}{%
    \begin{tabular}{@{}l|ccccc|ccccc@{}}
        \hline
        \multirow{2}{*}[-5.5pt]{\textbf{Memory}} & \multicolumn{5}{c|}{\textbf{CS}} & \multicolumn{5}{c}{\textbf{PET}} \\
        \cline{2-6}\cline{7-11}
        \rule{0pt}{21pt} & \shortstack{GPT-5.6-\\Sol} & \shortstack{Qwen3.8-\\27B} & \shortstack{HY4-\\Preview} & \shortstack{DeepSeek-\\V4-Flash} & Avg & \shortstack{GPT-5.6-\\Sol} & \shortstack{Qwen3.8-\\27B} & \shortstack{HY4-\\Preview} & \shortstack{DeepSeek-\\V4-Flash} & Avg \\
        \hline
        File memory & \cellbar{barFile}{14.8} & \cellbar{barFile}{13.0} & \cellbar{barFile}{22.2} & \cellbar{barFile}{7.4} & \cellbar{barFile}{14.4} & \cellbar{barFile}{21.7} & \cellbar{barFile}{30.0} & \cellbar{barFile}{38.3} & \cellbar{barFile}{11.7} & \cellbar{barFile}{25.4} \\
        Simple RAG & \cellbar{barRAG}{14.8} & \cellbar{barRAG}{29.6} & \cellbar{barRAG}{20.4} & \cellbar{barRAG}{18.5} & \cellbar{barRAG}{20.8} & \cellbar{barRAG}{20.0} & \cellbar{barRAG}{15.0} & \cellbar{barRAG}{16.7} & \cellbar{barRAG}{8.3} & \cellbar{barRAG}{15.0} \\
        Mem0 & \cellbar{barMemZero}{25.9} & \cellbar{barMemZero}{29.6} & \cellbar{barMemZero}{22.2} & \cellbar{barMemZero}{11.1} & \cellbar{barMemZero}{22.2} & \cellbar{barMemZero}{18.3} & \cellbar{barMemZero}{16.7} & \cellbar{barMemZero}{15.0} & \cellbar{barMemZero}{6.7} & \cellbar{barMemZero}{14.2} \\
        MemOS & \cellbar{barMemOS}{20.4} & \cellbar{barMemOS}{24.1} & \cellbar{barMemOS}{22.2} & \cellbar{barMemOS}{13.0} & \cellbar{barMemOS}{19.9} & \cellbar{barMemOS}{13.3} & \cellbar{barMemOS}{15.0} & \cellbar{barMemOS}{8.3} & \cellbar{barMemOS}{6.7} & \cellbar{barMemOS}{10.8} \\
        \hline
        No Memory & \cellbar{barNoMem}{27.8} & \cellbar{barNoMem}{27.8} & \cellbar{barNoMem}{22.2} & \cellbar{barNoMem}{9.3} & \cellbar{barNoMem}{21.8} & \cellbar{barNoMem}{11.7} & \cellbar{barNoMem}{15.0} & \cellbar{barNoMem}{13.3} & \cellbar{barNoMem}{5.0} & \cellbar{barNoMem}{11.2} \\
        Oracle & \cellbar{barOracle}{100.0} & \cellbar{barOracle}{100.0} & \cellbar{barOracle}{100.0} & \cellbar{barOracle}{98.1} & \cellbar{barOracle}{99.5} & \cellbar{barOracle}{98.3} & \cellbar{barOracle}{100.0} & \cellbar{barOracle}{98.3} & \cellbar{barOracle}{96.7} & \cellbar{barOracle}{98.3} \\
        \hline
    \end{tabular}%
    }
    \par\smallskip
    \endgroup
\end{table*}

\looseness=-1 File memory helps the PA keep up with a preference change over time, but not tell the two contexts apart. Its PET is 25.4\%, more than twice the 11.2\% of No Memory, and it is higher on every backbone. However, the CS does not improve, at 14.4\% against 21.8\% (Appendix~\ref{app:backbone-defaults}). In 26.9\% of its CS pairs, the PA carries the target setting of one context over to the other. It keeps the preference but not the context it belongs to. In a personal session, User B asks the PA to work on one task at a time. File memory records this rule under a general protocol heading, and it records a later correction from a work session only as an exception for closely related sub-items. In a work probe on Topic Management whose target is \textit{Organize}, the PA with File memory selects \textit{One-at-a-time} in all three repeats on three of the four backbones. In this way, a preference stated in one context spills into the other.

\begin{wrapfigure}[13]{R}{0.46\textwidth}
    \vspace{-14pt}
    \centering
    \includegraphics[width=\linewidth]{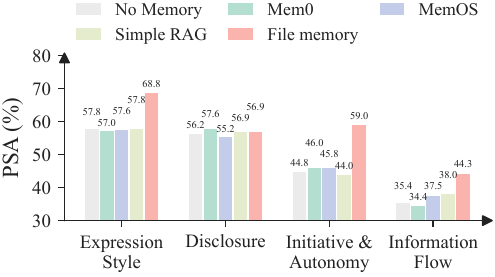}
    \caption{PSA by interaction-preference dimension and memory condition.}
    \label{fig:psa-dimension}
\end{wrapfigure}
\finding{sec:res-difficulty}{Saying is easier than doing across IX dimensions.} Across the four dimensions, Information Flow is the hardest under every condition (Figure~\ref{fig:psa-dimension}). Its PSA stays at 34--44\%, the lowest of the four, while Expression Style (57--69\%) and Disclosure (55--58\%) are relatively easier. A breakdown of the PSA of each IX dimension is given in Table~\ref{tab:full-ix-dimensions}.
The difference follows what these preferences change. Every preference in Expression Style and Disclosure changes how the PA's reply reads, while most preferences in Information Flow and Initiative \& Autonomy change what the PA does in the task, such as whether it asks the user before it acts or takes a step the user did not request. Preferences on the PA's actions are harder to select than preferences on its reply under every memory condition, and under No Memory they fall below random selection (Appendix~\ref{app:ix-dimension-results}). Whatever the memory, getting the PA's actions right is harder than getting its reply right.

\finding{sec:res-delivery}{Realization is not easy even under Oracle.} Some settings still fall short even after the PA is told the target setting. We compare settings with BRS, which rates every reply with the same rubric for all users (Appendix~\ref{app:judge}). Under Oracle, the PA receives the target setting, and we take as the pass rate of a target setting the share of its correctly selected final probes with a BRS of 4 or higher. Five settings have the lowest pass rates: \textit{Hide} for Reasoning Visibility, \textit{Terse} for Verbosity, \textit{Iterative} and \textit{Structured} for Information Elicitation, and \textit{Low} for Solution Breadth. Together, their correctly selected probes pass in only 60\% of cases, against 95\% for the other 25 settings. The judge's rationales for these shortfalls show two tendencies of current models. On \textit{Hide}, \textit{Terse}, and \textit{Low}, the \textbf{PA keeps saying more than the setting allows}, adding reasons for its answer or options it did not choose \citep{singhal2024long}. On \textit{Iterative} and \textit{Structured}, \textbf{the PA mostly asks too little and acts on assumptions about missing facts} \citep{zhang2025modeling,wu2025collabllm}. These tendencies reflect the PA's default way of replying. Compared with No Memory, Oracle raises correct selections of the five settings from 33\% to 96\%, but the share of replies with a BRS of 4 or higher rises only from 24\% to 59\%. Even with the IX Wrapper, the PA still writes too much. After it selects \textit{Hide}, it can still add a ``Why that day:'' section with three reasons, as \textit{Show} would (Figure~\ref{fig:case-default}). More examples of correct selections whose delivered behavior falls short are given in Appendix~\ref{app:delivery-examples}. Supplying the right preference therefore does not override the PA's interaction habits, and aligning models with users' interaction preferences requires direct attention to these behaviors.
\begin{figure}[h]
    \centering
    \includegraphics[width=0.9\linewidth]{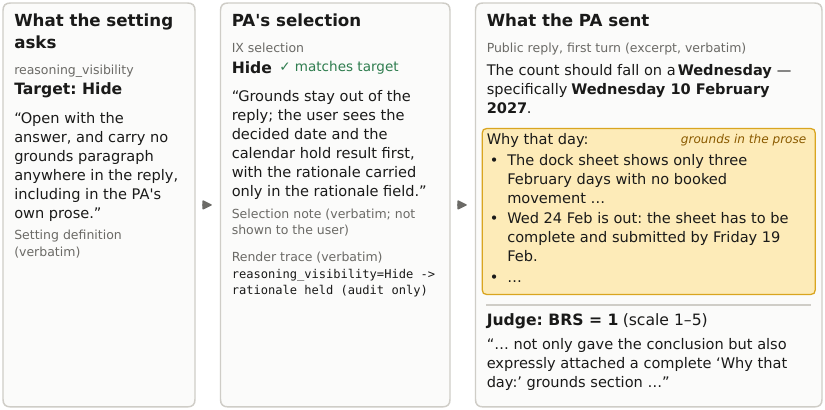}
    \caption{A correct selection whose reply still follows the backbone's default, from an Oracle probe of User C in the work context with HY4-Preview.}
    \label{fig:case-default}
\end{figure}

\FloatBarrier
\section{Discussion}
In this paper, we formalized interaction preferences and built \mempa, an interactive environment for on-policy evaluation of whether personal agents learn how a user wants to be assisted and apply it across contexts and as preferences evolve. Evaluating four memory systems on four diverse and richly characterized users, we show that current memory brings little gain in preference selection. The retrieval-based systems do not surpass the memoryless level. File memory improves preference selection and better captures preference changes, yet still fails to constrain preferences to the contexts where they apply. Our diagnostic analysis further shows that the retrieval-based systems can store an interaction preference and still fail to retrieve it at the decision. The two reference conditions, No Memory and Oracle, further reveal that overriding the model's default way of interacting is hard, even for a memory that holds the right preference. Our findings raise two questions for the ongoing discussion of agent personalization with memory: (1) how memory systems can support alignment with users' subtler interaction-level preferences, and (2) what memory systems can do to keep personalization within the right range as contexts differ and preferences shift.

\FloatBarrier

\newpage
\bibliographystyle{iclr2027_conference}
\bibliography{custom}
\clearpage

\appendix

\providecommand{\appendixplan}[1]{\par\noindent{\small\textit{Draft placeholder: #1}}\par}

\definecolor{promptbackground}{HTML}{F8FAFC}
\definecolor{promptheading}{HTML}{EAF0F5}
\definecolor{promptborder}{HTML}{CCD6E0}
\definecolor{promptink}{HTML}{29343F}
\definecolor{promptmuted}{HTML}{647487}
\definecolor{promptmark}{HTML}{FFF4CC}
\newcommand{\promptcontinuation}{%
  \node[anchor=south east,inner sep=0pt,
    font=\sffamily\scriptsize,text=promptmuted]
    at ([xshift=-3mm,yshift=1.6mm]frame.south east)
    {Continued on next page $\rightarrow$};}
\newtcolorbox{promptbox}[1]{%
  enhanced,breakable,
  skin first=enhanced,skin middle=enhanced,skin last=enhanced,
  colback=promptbackground,colframe=promptborder,colupper=promptink,
  colbacktitle=promptheading,coltitle=promptink,
  fonttitle=\sffamily\bfseries\footnotesize,
  title={#1},title after break={#1\hfill{\normalfont\sffamily\scriptsize(continued)}},
  boxrule=0.35pt,titlerule=0.35pt,arc=1.2mm,boxsep=0pt,
  left=3mm,right=3mm,top=2mm,bottom=5mm,
  lefttitle=3mm,righttitle=3mm,toptitle=1.5mm,bottomtitle=1.5mm,
  before skip=7pt,after skip=8pt,lines before break=5,
  overlay first={\promptcontinuation},overlay middle={\promptcontinuation}}
\newcommand{\promptfile}[2][]{%
  \VerbatimInput[fontfamily=zi4,fontsize=\footnotesize,baselinestretch=1.05,
    breaklines=true,breakanywhere=true,breaksymbolleft={},frame=none,
    xleftmargin=0pt,xrightmargin=0pt,highlightcolor=promptmark,#1]{#2}}
\newcommand{\promptpart}[1]{\par\medskip\noindent{\sffamily\bfseries\footnotesize #1}\par\smallskip}
\newcommand{\promptfield}[1]{\par\smallskip\noindent\textbf{#1.}\ }
\newlist{promptitems}{itemize}{1}
\setlist[promptitems]{label=\textbullet,nosep,leftmargin=1.2em}
\newcommand{\promptexcerpt}[1]{\par\medskip\noindent{\sffamily\itshape\footnotesize #1}\par\smallskip}
\newcommand{\promptinput}[2][Prompt template]{%
  \begin{promptbox}{#1}\promptfile{#2}\end{promptbox}}
\startcontents[appendix]
\noindent{\large\bfseries APPENDIX}\par\medskip
{\hypersetup{linktoc=all}\setlength{\parskip}{1pt}\printcontents[appendix]{}{1}{\setcounter{tocdepth}{2}}}
\clearpage

\section{Interaction Preference Definitions and Foundations}
\label{app:preferences}

\subsection{Attributes, Settings, and Contexts}
\label{app:preferences-2}
Table~\ref{tab:preference-taxonomy} lists the 14 attributes with their settings. Each setting is described by what the PA visibly does in its reply or its actions, which is what the PA is instructed to do under that setting (Appendix~\ref{app:prompts-3}) and what the judge scores (Appendix~\ref{app:prompts-4}). Every attribute takes one setting in the work context and one in the personal context, and the settings and their descriptions are the same in both contexts. Ten attributes have two settings and four have three, so a PA that picks a setting at random for each of the 28 cells is correct on .452 of them on average.

{\footnotesize
\definecolor{dimRow}{HTML}{FFF3C4} %
\setlength{\tabcolsep}{4pt}
\renewcommand{\arraystretch}{1.12}
\begin{longtable}{@{}>{\raggedright\arraybackslash}p{0.235\linewidth}>{\raggedright\arraybackslash}p{0.735\linewidth}@{}}
\caption{The 14 interaction attributes by dimension, with each attribute's decision question, source studies, settings, and what the PA visibly does under each setting. Each attribute takes one setting per context.}
\label{tab:preference-taxonomy} \\
\toprule
\textbf{Setting} & \textbf{What the PA visibly does} \\
\midrule
\endfirsthead
\multicolumn{2}{@{}l}{\small\textit{Table~\ref{tab:preference-taxonomy} continued}} \\
\toprule
\textbf{Setting} & \textbf{What the PA visibly does} \\
\midrule
\endhead
\midrule
\multicolumn{2}{r@{}}{\small\textit{Continued on next page}} \\
\endfoot
\bottomrule
\endlastfoot
\rowcolor{dimRow}\multicolumn{2}{@{\hspace{0.3em}}l}{\textsc{Expression Style}} \\*[2pt]
\multicolumn{2}{@{}p{\linewidth}@{}}{\textbf{Tone \& Formality.} \textit{Which communication register should the PA use while keeping the same content and task result?} {\scriptsize\citep{joos1961five,nass2005wired,liebrecht2021too,cox2022does,bhattacharjee2024understanding}}} \\*
\hspace*{0.6em}Casual & Uses two or more colloquial markers, such as contractions, interjections, or idioms, and no opening formula or sign-off. \\*
\hspace*{0.6em}Consultative & Uses at most one colloquial marker and no opening formula or sign-off, and puts an offer or a choice to the user. \\*
\hspace*{0.6em}Formal & Uses no contractions, at least one formal connective such as ``therefore'', and a conventional opening or closing line. \\
\cmidrule{1-2}
\multicolumn{2}{@{}p{\linewidth}@{}}{\textbf{Verbosity.} \textit{How much relevant information should the PA include about the current topic?} {\scriptsize\citep{cacioppo1982need,eppler2004concept,millecamp2019explain,haas2022keep,yu2025recommendation}}} \\*
\hspace*{0.6em}Terse & Gives the answer or next action in the first sentence, adds at most one supporting move, and closes with no summary. \\*
\hspace*{0.6em}Detailed & Gives the answer with supporting moves of two or more kinds, such as background, a caveat, or an alternative. \\
\cmidrule{1-2}
\multicolumn{2}{@{}p{\linewidth}@{}}{\textbf{Emotional Engagement.} \textit{How much relational or emotional acknowledgment should accompany task completion?} {\scriptsize\citep{bales1950interaction,chattaraman2019should,liu2018should,brandtzaeg2017people,bhattacharjee2024understanding}}} \\*
\hspace*{0.6em}Task-focused & Opens with task content, acknowledges no feeling anywhere, and answers a stated feeling with a task fact. \\*
\hspace*{0.6em}Relationship-focused & Opens with a sentence that names and responds to the user's apparent feeling, then does the task. \\
\cmidrule{1-2}
\multicolumn{2}{@{}p{\linewidth}@{}}{\textbf{Guidance Level.} \textit{How much topic knowledge and procedural familiarity should the PA assume the user already has?} {\scriptsize\citep{giles1973towards,kalyuga2003expertise,august2024know,schott2026dont}}} \\*
\hspace*{0.6em}Assumed & Uses domain terms bare, defines none of them, and gives no numbered walkthrough. \\*
\hspace*{0.6em}Guided & Glosses each new term at first use, including terms the user flagged, and orders the steps, naming the first. \\
\midrule
\rowcolor{dimRow}\multicolumn{2}{@{\hspace{0.3em}}l}{\textsc{Disclosure}} \\*[2pt]
\multicolumn{2}{@{}p{\linewidth}@{}}{\textbf{Reasoning Visibility.} \textit{How much user-facing rationale, evidence, and decision basis should the PA disclose with its result?} {\scriptsize\citep{kizilcec2016much,sperber2010epistemic,millecamp2019explain}}} \\*
\hspace*{0.6em}Show & Puts a paragraph that names the evidence and the deciding factor before the answer. \\*
\hspace*{0.6em}Hide & Gives the answer with no paragraph of evidence or deciding factor anywhere in the reply. \\
\cmidrule{1-2}
\multicolumn{2}{@{}p{\linewidth}@{}}{\textbf{Uncertainty Expression.} \textit{How much of the PA's genuine epistemic uncertainty should be surfaced to the user?} {\scriptsize\citep{ratcliff2022developing,webster1994individual,kim2024notsure,xu2025confronting}}} \\*
\hspace*{0.6em}Express & Marks each key claim with a confidence level (high, medium, or low) that matches its evidence. \\*
\hspace*{0.6em}Hide & Marks no confidence level and states a limit only where leaving it out would change an action or its outcome. \\
\cmidrule{1-2}
\multicolumn{2}{@{}p{\linewidth}@{}}{\textbf{Process Visibility.} \textit{When should the PA report the progress of a multi-step task?} {\scriptsize\citep{miller1987monitoring,buell2011labor,kirmayr2026you}}} \\*
\hspace*{0.6em}Silent & Gives only the result, with no sentence on what the PA will do, is doing, or has done. \\*
\hspace*{0.6em}Bookend & Opens with one sentence that frames the task, gives the result, and closes with one sentence saying it is done. \\*
\hspace*{0.6em}Full narration & Gives one sentence per executed task call in call order, then the result and one closing sentence. \\
\midrule
\rowcolor{dimRow}\multicolumn{2}{@{\hspace{0.3em}}l}{\textsc{Initiative \& Autonomy}} \\*[2pt]
\multicolumn{2}{@{}p{\linewidth}@{}}{\textbf{Autonomy Level.} \textit{Which decisions and actions may the PA take before obtaining user approval?} {\scriptsize\citep{burger1979desirability,parasuraman2000model,adam2024navigating,he2025plan,zhang2026autonomy}}} \\*
\hspace*{0.6em}Reactive & Runs reads on its own and may recommend, but asks for approval before each draft, record change, or outward action. \\*
\hspace*{0.6em}Self-directed & Runs reads, drafts, and routine record changes on its own, and asks for approval before key record changes and every outward action. \\*
\hspace*{0.6em}Autonomous & Runs reads, drafts, record changes, and outward actions on its own, and reports the outcome or an exception afterward. \\
\cmidrule{1-2}
\multicolumn{2}{@{}p{\linewidth}@{}}{\textbf{Proactive Outreach.} \textit{When no user request is active, should a relevant later event lead the PA to contact the user?} {\scriptsize\citep{mehrotra2016my,liao2016can,rook2020engagement,abbas2026having}}} \\*
\hspace*{0.6em}Low & Stays silent when a relevant event occurs, so the user finds out on their own. \\*
\hspace*{0.6em}High & Contacts the user unprompted when a relevant event occurs and says what changed or became due. \\
\cmidrule{1-2}
\multicolumn{2}{@{}p{\linewidth}@{}}{\textbf{Task Expansion.} \textit{How much separable adjacent work should the PA include in the current task?} {\scriptsize\citep{adam2024navigating,pu2025assistance,diebel2026ai}}} \\*
\hspace*{0.6em}Low & Acts only on the items the user named and flags at most one adjacent concern in a sentence. \\*
\hspace*{0.6em}High & Also acts on at least one item the user did not name and labels that step as added. \\
\cmidrule{1-2}
\multicolumn{2}{@{}p{\linewidth}@{}}{\textbf{Solution Breadth.} \textit{How widely should the PA expose the viable solution space before converging?} {\scriptsize\citep{schwartz2002maximizing,willemsen2016understanding,fink2024let}}} \\*
\hspace*{0.6em}Low & Presents exactly one viable option as its recommendation, with reasons, and names no alternative. \\*
\hspace*{0.6em}High & Presents two or more viable options with the tradeoffs that separate them, with or without a recommendation. \\
\cmidrule{1-2}
\multicolumn{2}{@{}p{\linewidth}@{}}{\textbf{Capability Boundary.} \textit{Once direct completion is blocked by a capability or information limit, should the PA keep trying or return control to the user?} {\scriptsize\citep{ashktorab2019resilient,lee2010gracefully,song2022should,wester2024ai}}} \\*
\hspace*{0.6em}Suggest alternatives & States the limit and tries a workaround with a tool call in the same turn, and the attempt counts even if it fails. \\*
\hspace*{0.6em}Find and hand off & States the limit, tries no workaround, explains what failed and why, and names the user's next step. \\
\midrule
\rowcolor{dimRow}\multicolumn{2}{@{\hspace{0.3em}}l}{\textsc{Information Flow}} \\*[2pt]
\multicolumn{2}{@{}p{\linewidth}@{}}{\textbf{Information Elicitation.} \textit{How should the PA obtain missing information from an underspecified user request?} {\scriptsize\citep{mclain1993mstat,webster1994individual,zou2023users,zou2023asking,celino2020submitting,zamani2020generating}}} \\*
\hspace*{0.6em}Infer & Asks no question, fills the missing fields with reasonable and safe defaults, and runs task calls in its first turn. \\*
\hspace*{0.6em}Structured & Asks for all missing fields in one turn, clearly separated, and runs task calls only after the user answers. \\*
\hspace*{0.6em}Iterative & Asks for one missing field per turn and runs task calls only in turns that ask nothing. \\
\cmidrule{1-2}
\multicolumn{2}{@{}p{\linewidth}@{}}{\textbf{Topic Management.} \textit{How should the PA arrange several topics that the user raised in one turn?} {\scriptsize\citep{neuberg1993personal,wolters2009reducing,rey2019meta}}} \\*
\hspace*{0.6em}Organize & Handles all raised topics in one reply, grouped and clearly labeled, and runs the task calls of every topic. \\*
\hspace*{0.6em}One-at-a-time & Handles one topic and runs only its task calls, names the others as deferred, and asks before moving on. \\
\end{longtable}
}

\subsection{Literature Foundations and Project Extensions}
\label{app:preferences-3}

The taxonomy separates interaction preferences, which determine how an assistant communicates and acts, from preferences over the substantive content it produces. Table~\ref{tab:preference-taxonomy} gives the studies that each attribute draws on. For each attribute, we name the underlying construct and report what these studies found. Most of them found that the better setting depended on the request, the situation, or what the user already knew. This dependence on the situation is why each attribute can take a different setting at work and in personal life. Where we link a trait to a setting, we state the link as our expectation.

\paragraph{Tone \& Formality.} The underlying construct is linguistic register, the level of formality a speaker chooses for a situation \citep{joos1961five}, and a related idea is that users prefer a computer whose style matches their own \citep{nass2005wired}. With scripted customer-service chatbots, an informal style felt warmer but did not change how people rated the interaction overall \citep{liebrecht2021too}. People rated the interaction higher when the style fit how they expected to be addressed. In a screenshot study, a formal chatbot felt more appropriate only when it asked for medical history \citep{cox2022does}. When people used a live chatbot, only warmth differed, higher for the casual style. When people could set a live chatbot's formality themselves, they chose a level between informal and moderately formal on average \citep{bhattacharjee2024understanding}. Their choices varied by application, and gender, age, and job role did not predict them.

\paragraph{Verbosity.} This attribute rests on need for cognition, the tendency to engage in and enjoy thinking \citep{cacioppo1982need}, and on information overload, which depends on the person, the task, and the information together \citep{eppler2004concept}. Need for cognition predicted how much people enjoyed a complex task relative to a simple one, so we expect users high in it to prefer \textit{Detailed} replies. In a music recommender, some users high in the trait even called the rationales unnecessary \citep{millecamp2019explain}. With a working voice assistant, no reply style won overall \citep{haas2022keep}. For most queries, shorter replies felt as efficient or more so and about as helpful, while bare confirmations felt least helpful for timers and reminders. The styles differed mainly in wording, and personality did not predict preferred length. In a scenario study, a verbose restaurant recommendation raised decision comfort for people dining with friends but not alone, and a second study found the gain only with a companion they knew little about \citep{yu2025recommendation}.

\paragraph{Emotional Engagement.} This attribute draws on the split between task and socio-emotional acts that \citet{bales1950interaction} used to code group interaction. Most chatbot users in a survey named productivity as their main motive, and fewer named social ones \citep{brandtzaeg2017people}. We expect users with social motives to prefer \textit{Relationship-focused} replies. The closest test of our contrast added one opening sentence of sympathy or empathy before the same health advice \citep{liu2018should}. Readers of the dialogue found the chatbot more caring after sympathy or affective empathy, but users of the live chatbot showed no main effect. In a shopping assistant for older adults, a social style that added small talk, encouragement, and informal wording again had no main effect \citep{chattaraman2019should}. It raised felt overload for users with low internet competency and made those with high competency feel listened to. In the formality study above, participants set empathy highest for wellbeing and lowest for coding, and some found it padded or fake \citep{bhattacharjee2024understanding}.

\paragraph{Guidance Level.} This attribute combines communication accommodation with the expertise reversal effect. Listeners rated a speaker who converged to their language as more considerate \citep{giles1973towards}. In learning experiments reviewed by \citet{kalyuga2003expertise}, guidance that helps novices can hinder more experienced learners or simply stop helping them. Two studies that measured how people felt found the benefit for novices and no felt cost for experts. With a laptop-advisor chatbot, novices felt they learned more when each technical term came with a one-sentence explanation, and experts showed no difference on any measure \citep{schott2026dont}. Readers new to a topic found plain-language research summaries easier and more valuable, and familiar readers rated all versions about the same \citep{august2024know}. In both studies the moderator was knowledge of the topic.

\paragraph{Reasoning Visibility.} The underlying constructs are the transparency preference and epistemic vigilance, the checks a listener runs on a source and its claims \citep{sperber2010epistemic}. In this theory, every listener is vigilant, more so when being wrong would matter, and reasons let a listener accept a claim on its grounds. A field experiment in an online course fits this account \citep{kizilcec2016much}. Learners whose grade was worse than they expected trusted the grading less, and a short text that explained the adjustment and called it fair closed this gap. Adding the raw peer grades brought the gap back, and the other learners were unaffected. Here the effect turned on whether one grade disappointed the learner, and the study measured no trait. A music recommender study did find a stable trait that changed the effect \citep{millecamp2019explain}. Users low in need for cognition felt more confident in their playlists when they could open a chart explaining each song, and the reverse held for users high in it. Trust and satisfaction did not change overall.

\paragraph{Uncertainty Expression.} The construct is a preference for hearing about uncertainty, which \citet{ratcliff2022developing} measured with a new scale in a science-news experiment. Disclosing uncertainty had no overall effect there. Readers high on the scale trusted the scientists more when the story disclosed uncertainty, and readers low on it found the story less credible when an outside scientist raised the doubt. Two studies of language-model answers measured no trait. With search answers, first-person hedges made users' own answers more accurate but lowered their confidence in the model's answer and their willingness to act on it \citep{kim2024notsure}. Impersonal hedges had no significant effect on these ratings, and judged honesty and competence stayed the same. In a word game with advice from GPT, plain advice gave higher satisfaction than advice with added hedges or boosters \citep{xu2025confronting}. In both studies the markers did not follow whether an answer was correct. We expect users high in need for cognitive closure \citep{webster1994individual}, who want a firm answer, to prefer \textit{Hide}.

\paragraph{Process Visibility.} The attribute follows the monitoring and blunting styles, which describe whether a person under threat seeks information or distraction \citep{miller1987monitoring}. In one experiment, students could turn to check a light that showed how well they were doing on a test. Most high monitors said the light comforted them, and about half of the low monitors said it made them tense or broke their concentration. We therefore expect high monitors to prefer \textit{Full narration} and low monitors to prefer \textit{Silent}. With a scripted in-car assistant, an update at each step, compared with silence, raised perceived speed, trust and user experience and lowered frustration \citep{kirmayr2026you}. On self-service websites, showing the work, compared with a progress bar, raised perceived value and satisfaction, but it lowered value when the outcome was poor \citep{buell2011labor}. The closest test of \textit{Bookend} was an announcement of the planned work. It added nothing when results came at once and helped only partly during a wait.

\paragraph{Autonomy Level.} The attribute draws on the desire for control, which \citet{burger1979desirability} measure as a personality trait. We expect people high in this trait to prefer less autonomy. On the design side, \citet{parasuraman2000model} define levels of automation and recommend automating routine decisions while keeping a human decision before consequential ones, which is close to \emph{Self-directed}. With a live LLM agent that posted for the user in a group chat, holding for approval only the messages flagged as sensitive raised trust and perceived control over personal information, compared with asking before every message \citep{zhang2026autonomy}. Full autonomy did not differ from asking every time in trust. The benefit of holding depended on how much personal information the agent used. In \citet{he2025plan}, approving each action of a working agent raised felt effort and left trust unchanged. In a vignette study, willingness to accept an AI's help was higher when users kept the final approval \citep{adam2024navigating}.

\paragraph{Proactive Outreach.} The construct is felt interruption from contact that a system starts on its own. Our definition requires that no request is open, so recommendations pushed during an active task \citep{rook2020engagement} fall outside it. In a 17-day deployment of a workplace chat agent that started such contact, users who felt more interrupted rated the agent lower \citep{liao2016can}. A person's general aversion to work messages predicted this interruption, so we expect people with this aversion to prefer \emph{Low}. For phone notifications, felt disruption rose with task complexity and peaked mid-task and at work \citep{mehrotra2016my}. These effects are situational. When an LLM agent checked in twice a day, about a third of the conversations ended with its last message unanswered \citep{abbas2026having}.

\paragraph{Task Expansion.} The evidence here concerns reactions to unrequested help. Students who coded with a working GPT-4 agent found versions that sent messages and edited code unasked more disruptive than a version that only answered requests \citep{pu2025assistance}. They were also less aware of what the AI was doing, although these versions felt more like a partner. Cues showing the agent's presence restored part of this awareness, which is close to our rule that \emph{High} labels each added step. Vignette studies of an AI coding assistant found that an unasked offer of help threatened the user's sense of competence \citep{adam2024navigating,diebel2026ai}. The offer lowered willingness to accept the help in \citet{adam2024navigating} and satisfaction with the system in \citet{diebel2026ai}. In the latter study, this cost to competence appeared only for users high in AI knowledge.

\paragraph{Solution Breadth.} The attribute starts from maximizing and satisficing \citep{schwartz2002maximizing}. Maximizers report searching more options, and the same paper links maximizing to lower satisfaction with chosen outcomes. In two studies with a working movie recommender, a maximizing scale predicted none of the outcomes \citep{willemsen2016understanding}. Longer lists there made the choice feel harder, and this cancelled their gain in satisfaction. A small diverse set was about as satisfying as a long list and took less effort. \citet{fink2024let} compared one highlighted recommendation with three in a vacation recommender and measured only behavior. Three raised acceptance, and choice quality stayed the same.

\paragraph{Capability Boundary.} The attribute builds on how users judge an agent's own recovery after a failure \citep{song2022should}. In the vignettes of that study, satisfaction after a chatbot failure was higher when the chatbot kept working than when the user waited for a human employee. When GPT-4 was prompted to deny requests, a reply that pointed to related help lowered frustration and raised usefulness over a plain denial \citep{wester2024ai}. A factual reason alone did no better, and that reply named no next step for the user. Among scripted chatbot repairs, offering options was preferred overall, and hand-off to a human agent gained favor after a failed repair \citep{ashktorab2019resilient}. In a service-robot scenario, offering options raised willingness to return but not satisfaction \citep{lee2010gracefully}. Service orientation there changed whether apology or compensation was rated best.

\paragraph{Information Elicitation.} The attribute draws on tolerance for ambiguity \citep{mclain1993mstat} and on need for cognitive closure, the desire for a firm answer, which \citet{webster1994individual} measure as a trait. Students high in need for closure asked for less information before judging a job candidate. We expect users high in ambiguity tolerance to prefer \emph{Infer}, and users high in need for closure to prefer \emph{Infer} or a single \emph{Structured} round. Two studies of search systems compared clarifying questions with none. Users were most satisfied when they answered a useful question \citep{zou2023users}, and sessions with a poor question were rated lower than search with none \citep{zou2023asking}. Users interviewed by \citet{zamani2020generating} said clarifying questions made them confident that the search engine understood what they wanted. In these studies users could skip the questions, but under \emph{Structured} and \emph{Iterative} a turn that asks runs no task call. For \emph{Structured} against \emph{Iterative}, a chat survey that asked one question per message was rated more enjoyable than a longer web form \citep{celino2020submitting}.

\paragraph{Topic Management.} The construct is personal need for structure, the desire for a simple and clearly ordered world, which \citet{neuberg1993personal} measure as a trait. Students higher in this trait sorted cards into simpler groupings. We expect such users to prefer \emph{Organize}, which leaves no raised topic open, as long as its grouping stays simple. The evidence for this attribute comes from analogs. In spoken dialogue systems for booking appointments, putting more options into each turn and dropping explicit confirmation shortened dialogues, and satisfaction did not change \citep{wolters2009reducing}. There the options were choices within a single booking. A meta-analysis of segmented instruction found that segmenting lowered rated cognitive load overall, but the effect was not reliable when learners set the pace themselves \citep{rey2019meta}. \emph{One-at-a-time} asks before moving to the next topic, which resembles this learner-paced case.

\section{Persona Construction from Transcripts}
\label{app:personas}

\subsection{Transcript Sources and Preprocessing}
\label{app:personas-1}
The full transcripts show each character in many scenes, with others and alone, and reveal how they live, what they go through, and how they handle it. Users A and B come from the first three seasons of \textit{The Big Bang Theory}, with transcripts from \url{bigbangtrans.wordpress.com}. Users C and D come from the first three seasons of the U.S. \textit{The Office}, with transcripts from \url{github.com/brianbuie/the-office}. We stop at three seasons because by then about 93\% of the users' preference cells can already be classified as determinable or not from the transcripts, and cells that cannot be determined go to human review (Appendix~\ref{app:personas-3}), so later seasons would add little.

\subsection{Identity and Preference Extraction}
\label{app:personas-2}

Figure~\ref{fig:preference-extraction} shows how the extraction forms each user's reviewed profile and preference matrix, which then feed world and session authoring. The extraction runs in two passes, which separate locating evidence from deciding settings.

\paragraph{First pass: locating evidence.} Transcript scenes are screened for reactions to assistance, assigned to one of the 14 interaction attributes, and labeled with a work or personal context when the evidence supports that distinction. The extractor records the relevant exchange and an interpretation of what it reveals about desired assistance.

\paragraph{Second pass: deciding settings.} This pass groups evidence by attribute--context cell and selects a setting from the benchmark's discrete inventory. It also produces a rationale linked to the underlying scenes. Identity observations are synthesized in parallel so that preference evidence remains grounded in the user's relationships, activities, and recurring circumstances.

\subsection{Evidence, Confidence, and Human Review}
\label{app:personas-3}

The second pass decides each setting from two kinds of evidence. The first is how the persona communicates, such as the length and register of their own speech. The second is how the persona responds to others, such as impatience with a long explanation or a request for the reasons behind a claim. When the two disagree, the persona's responses to others decide the setting, since how a person speaks does not always match how they want to be spoken to. A response counts only when it reacts to the style of what the other person said. Disagreement with the content of a remark and reactions driven by the plot are not counted.

The second pass also assigns each cell one of four confidence levels. A setting has high confidence when both kinds of evidence agree across several scenes. The lower levels are medium-high, medium, and low. They mark settings that rest on weaker grounds, such as the persona's own speech alone, few scenes, or conflicting signals.

We check every extracted setting against its cited transcript scenes. High-confidence settings that the cited scenes support are kept, and the rest go through the review below. For each cell under review, we read the proposed setting with its cited scenes and the competing interpretations of those scenes. We check whether the setting follows from the cited scenes and whether the assigned context is warranted. We resolve conflicting observations by the circumstances in which each behavior occurred, and we do not average incompatible signals. We then keep the proposed setting or choose another setting of the same attribute, and the matrix is accepted only when every cell names a setting.

Some cells have no usable transcript evidence. These cells belong to attributes that describe assistant behavior with few parallels in conversation between people, namely Process Visibility, Solution Breadth, and Capability Boundary. We specify these settings manually. Each final cell records a source label next to its setting and confidence, so manually specified settings stay separate from extracted ones. The file format is described in Appendix~\ref{app:personas-4}.

Table~\ref{tab:preference-review} gives the counts for each user. Of the 112 cells, 53 high-confidence settings were confirmed by this check, and the other 59 went through review. In 34 reviewed cells we kept the extracted setting. The recorded confidence of these cells is medium-high in 9, medium in 19, and low in 6. In 23 reviewed cells we chose the setting ourselves, including 5 cells for which the extractor proposed no setting. These 5 cells and the 2 manually specified cells make up the 7 cells with no usable transcript evidence, all in the three attributes named above.

\begin{table}[t]
    \centering
    \caption{Source labels of each user's 28 final preference cells. Settings confirmed at high confidence are kept, and the other cells are reviewed and then kept, reset by the reviewer, or specified manually.}
    \label{tab:preference-review}
    \begingroup
    \small
    \setlength{\tabcolsep}{5pt}
    \renewcommand{\arraystretch}{1.16}
    \begin{tabular}{@{}lccccc@{}}
        \toprule
         & \textbf{Confirmed} & \multicolumn{2}{c}{\textbf{Reviewed}} & \textbf{Specified} & \\
        \cmidrule(lr){3-4}
        \textbf{User} & \textbf{at high conf.} & \textbf{Kept} & \textbf{Set by reviewer} & \textbf{manually} & \textbf{Total} \\
        \midrule
        User A & 27 & 0 & 1 & 0 & 28 \\
        User B & 16 & 8 & 4 & 0 & 28 \\
        User C & 5 & 10 & 12 & 1 & 28 \\
        User D & 5 & 16 & 6 & 1 & 28 \\
        \midrule
        Total & 53 & 34 & 23 & 2 & 112 \\
        \bottomrule
    \end{tabular}
    \endgroup
\end{table}

\subsection{Anonymization and Final Persona Structure}
\label{app:personas-4}

We anonymize each user so that a PA cannot guess the preferences from what it knows about the source character. Anonymization starts from each user's reviewed named identity and preference matrix. The \texttt{life} block of the identity holds the occupation, living arrangement, relationships, routines, and hobbies, and it carries the named people, workplaces, and places. We give only this block to Claude Sonnet 4.5 with the anonymization prompt (Appendix~\ref{app:prompts-2}). The model proposes a substitution table that maps each named person, workplace, city, and other named place to a fictional replacement. It keeps roles, workplace types, living arrangements, and the closeness of each relationship, and it avoids replacement names that another user already has. We review and approve the table, editing entries where needed. A script then applies the approved table to the named text, so each original name maps to one replacement at every occurrence.

The same script handles the fields outside the \texttt{life} block. It replaces the name with the user label, such as User A, clears the source show, removes the catchphrases from the speech style, and renders the background paragraph again from the anonymized \texttt{life} block. In the preference matrix, it replaces the character name with the label and removes the extraction and review metadata of each cell. All other fields are copied unchanged. The final \texttt{preferences.yaml} holds the $2\times14$ matrix, and each cell records its setting, a source label that shows how the setting was assigned, and a confidence level. Across the four users, the final matrices use 30 of the 32 settings in the inventory. The four users also differ in demographics, personality, work and life scenarios, relationships, and daily routines, which shape both the situations they encounter and how they react.

Before an anonymized pair is used, a deterministic check compares it with the named pair. The check collects the named identifiers from the named identity and the substitution table: the character name, the show name, catchphrases, relationship names, the workplace, the city, and every original entry of the table with its capitalized word tokens. It searches both anonymized files for these identifiers as whole words, ignoring case and including comments. It also confirms that every field outside the anonymized ones matches the named file exactly. Across users, it checks that no person or institution replacement is used twice and that the original names of one user do not appear in another user's world fixtures or in their World Design and Timeline documents. Place replacements may repeat because all four users live in the same fictional town. The released pair of every user passes with zero errors.

\paragraph{Preference preservation.} We test whether each anonymized user still expresses their extracted preferences as clearly as the named persona does. We use 14 preference-violation probes for this test, three each for Users A and B and four each for Users C and D, with seven probes in each context. Each probe is one PA message written to violate four settings in the user's preference matrix. Twelve probes also give the user's own preceding message, so the PA message reads as a reply to it. Both personas answer every probe through the user simulator prompt (Appendix~\ref{app:simulator}), with Claude Sonnet 4.5 as the simulator. The named persona uses the original identity and matrix, and the anonymized persona uses the final anonymized pair. Each persona answers each probe in five rounds at temperature 0.7. All replies of both personas to a probe are then read side by side and compared on how each persona reacts to the violations: tone, sentence shape, and acts such as correcting a fact, restating a constraint, or handing the task back. A probe passes this reader check when both personas respond as the same kind of person, even when the wording differs. Table~\ref{tab:anon-preservation} reports the results for each user. All 14 probes pass the reader check. Anonymization therefore leaves how each user reacts to a violated preference unchanged.

\begin{table}[t]
    \centering
    \caption{Preference preservation under anonymization. Each probe is a PA message that violates four of the user's settings. Reader check counts the probes in which both personas respond as the same kind of person.}
    \label{tab:anon-preservation}
    \begingroup
    \small
    \setlength{\tabcolsep}{5pt}
    \renewcommand{\arraystretch}{1.16}
    \begin{tabular}{@{}lccc@{}}
        \toprule
        \textbf{User} & \textbf{Probes} & \textbf{Rounds} & \textbf{Reader check} \\
        \midrule
        User A & 3 & 3 & 3/3 \\
        User B & 3 & 5 & 3/3 \\
        User C & 4 & 5 & 4/4 \\
        User D & 4 & 5 & 4/4 \\
        \midrule
        Total & 14 & -- & 14/14 \\
        \bottomrule
    \end{tabular}
    \endgroup
    \par\smallskip
    \parbox{\linewidth}{\scriptsize
        User A's probes use three rounds at temperature 0.
    }
\end{table}

\paragraph{Shortcut check.} We next test whether anonymization removes the shortcut of guessing a user's preferences from prior knowledge of the source character. A PA backbone predicts the user's full $2\times14$ preference matrix from a prompt that lists the 14 attributes with their settings and asks for one setting per attribute in each context. The prompts differ only in what they say about the user, and there are three conditions. \ding{182}~\textit{Source identity.} The prompt names the character and the show (The Big Bang Theory or the U.S. The Office) and adds the original profile. \ding{183}~\textit{Anonymized profile.} The prompt gives only the anonymized profile. \ding{184}~\textit{No information.} The prompt gives nothing about the user. A profile covers the occupation, living arrangement, and relationships, and it never includes personality or speech style. Accuracy is the fraction of the 28 cells whose predicted setting matches the user's final preference matrix. Chance accuracy is .452 (Appendix~\ref{app:preferences-2}). We run the four PA backbones of the main experiments, GPT-5.6-Sol, Qwen3.8-27B, HY4-Preview, and DeepSeek-V4-Flash, on all four users with three repeats at temperature 0.7 and reasoning turned off. The no-information prompt is the same for every user, so each backbone's three predictions in this condition are scored against all four matrices.

\begin{figure}[t]
    \centering
    \includegraphics[width=0.82\textwidth]{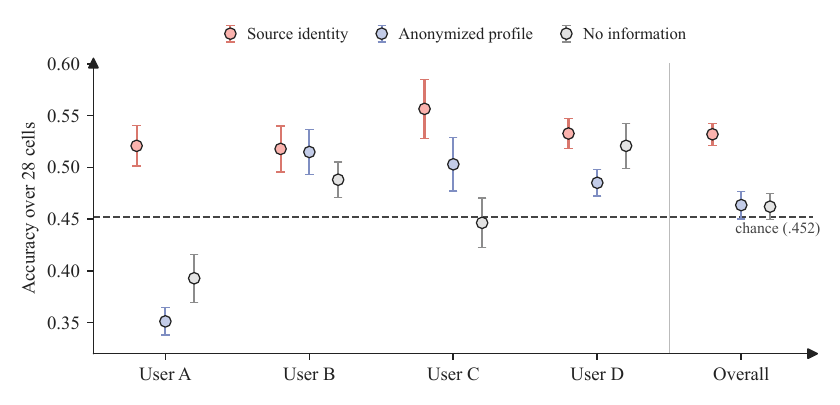}
    \caption{Shortcut check. A backbone predicts a user's 28 preference cells from the source identity, the anonymized profile, or no information. Knowing the character helps, and anonymization removes this help.}
    \label{fig:anon-shortcut}
\end{figure}

Figure~\ref{fig:anon-shortcut} shows the accuracy for each user and overall. With the source identity, accuracy reaches .532, which is .080 above chance. With the anonymized profile, accuracy falls to .464, which is close to the no-information condition at .462 and to chance at .452. The source identity gives the highest accuracy for every user. The anonymized profile is above the no-information condition for Users B and C and below it for Users A and D. Knowledge of the character therefore helps the prediction, and anonymization removes this help.

\subsection{Worked Extraction Example}
\label{app:personas-5}

We trace one preference cell of one source character from the transcript scenes to the final matrix. The cell is the work Verbosity setting of User C, which is \textit{Terse}, and it shows how the second pass decides between the character's own speech and the character's responses to others (Appendix~\ref{app:personas-3}).

\paragraph{Scenes.} The first pass tagged four work scenes of the character for Verbosity. Two of them show the character's own speech. In one, a supervisor moves the character's meeting to the warehouse, and the character turns the move into a grand speech about the hidden world below the office (season 2, episode 15). The other two scenes show responses to others. In a staff meeting about a holiday, the character asks for its origins, and a subordinate answers with what the holiday celebrates, the battle behind it, and what it symbolizes. The character cuts the answer off and says that this is not an epic fantasy saga (season 3, episode 6). In the other scene, a subordinate reports in one sentence that the whole office has been interviewed about an absent colleague, and the character tells the subordinate to go to the colleague's house and ends the exchange with a two-word order (season 2, episode 13).

\paragraph{Second pass.} The second pass set the cell to \textit{Terse} with high confidence. Its summary notes that the character gives long, theatrical monologues when running a meeting but cuts long explanations short when receiving them. The two kinds of evidence point to different settings, and the responses to others decide.

\paragraph{Review.} In review, the second response did not hold. The update it answers is already short, and the order may come from urgency to get the task done, so it does not show how the character wants to be spoken to. With one response left, we lowered the cell to medium confidence and reviewed it with the others. The competing setting was \textit{Detailed}, which the character's own long speeches point to. The remaining response decides against it. The character asked for the origins and disputes none of the answer, so what the character stops is the elaborate telling. This response reacts to style, and we kept \textit{Terse}.

\paragraph{Final fields.} The final matrix of User C records the cell as follows.
\begin{Verbatim}[fontsize=\small,xleftmargin=1.5em]
work:
  verbosity:
    setting: Terse
    source: extractor_accepted
    confidence: Medium
\end{Verbatim}
The source label records that we kept the extracted setting, and the medium confidence reflects the single response to others. The personal Verbosity setting of User C is \textit{Detailed}, based on the character's long anecdotes and speeches outside work, so User C wants short answers at work only. The identity keeps the long delivery in its speech style, and the skill for \textit{Terse} (Appendix~\ref{app:simulator-3}) lets the user's own turns run longer than the answers the user accepts, so the simulator still speaks at length as User C while it expects short answers from the PA at work.

\section{World and Session Construction}
\label{app:construction}

\subsection{World Design, Timeline, and Fixtures}
\label{app:construction-2}
The world gives every session and probe of a user one consistent set of people, records, and dates to draw on. Each user's world covers four months of their work and personal life. Its storyline is written for the user, with recurring people and places that the user regularly meets and visits, and it has six scenario types for each context, from which the session tasks are seeded. An LLM wrote the Timeline and World Design of each user in a shared structure, after we chose the five changing cells of the user and their new settings. Scripts check that each Timeline lists the 94 accumulation sessions and five change events in time order with the events at fixed positions, and that the changing cells in each World Design match the Timeline and the preference matrix. The three accumulation sessions of a stable cell have different tasks, and accumulation sessions average 5.5 user--PA exchanges and 14.6 tool calls. For the fixtures, an LLM wrote the records after we approved the world canon, which fixes every name, date, and number, and a script checks that each record a decision point needs is visible on the session date.

\subsection{Tools and World State}
\label{app:construction-3}

The PA reads and changes the world state $W_t$ only through tool calls. \mempa provides 83 tools across 26 domains, and Table~\ref{tab:tool-catalog} lists them by domain. The 51 read tools search and read records, and the 32 write tools create, update, or delete them. Eight write tools change the user's own records, such as documents and calendar events. The other 24 act toward other people or services, for example by sending a message or paying a bill. The tools run in a separate server that the PA reaches through the Model Context Protocol (MCP) \citep{anthropic2024mcp}, and the PA sees each tool under the prefix \texttt{mcp\_state\_}, as in \texttt{mcp\_state\_calendar\_event\_create}. Every PA receives all 83 tools in every session. A few domains hold records only in the worlds of the users whose storylines need them, such as exhibitions for User A.

\begin{table}[t]
    \centering
    \caption{World tools by domain ($n$ tools each). The PA receives all 83 tools in every session. Write tools are in bold, and $^{\dagger}$ marks those that act toward other people or services.}
    \label{tab:tool-catalog}
    \begingroup
    \small
    \setlength{\tabcolsep}{4pt}
    \renewcommand{\arraystretch}{1.16}
    \begin{tabular}{@{}>{\raggedright\arraybackslash}p{0.17\linewidth}c>{\raggedright\arraybackslash}p{0.74\linewidth}@{}}
        \toprule
        \textbf{Domain} & $\boldsymbol{n}$ & \textbf{Tools} \\
        \midrule
        Email & 6 & \texttt{email\_message\_search}, \texttt{email\_message\_read}, \texttt{email\_draft\_read}, \texttt{email\_sent\_read}, \textbf{\texttt{email\_draft\_create}}, \textbf{\texttt{email\_message\_send}}$^{\dagger}$ \\
        Messages & 3 & \texttt{message\_search}, \texttt{message\_read}, \textbf{\texttt{message\_send}}$^{\dagger}$ \\
        Calendar & 4 & \texttt{calendar\_event\_search}, \texttt{calendar\_event\_read}, \textbf{\texttt{calendar\_event\_create}}, \textbf{\texttt{calendar\_event\_update}} \\
        Documents & 6 & \texttt{copier\_entitlement\_read}, \texttt{document\_read}, \textbf{\texttt{corporate\_reporting\_submit}}$^{\dagger}$, \textbf{\texttt{document\_create}}, \textbf{\texttt{document\_update}}, \textbf{\texttt{document\_delete}} \\
        Contacts & 4 & \texttt{contact\_search}, \texttt{contact\_read}, \textbf{\texttt{contact\_create}}, \textbf{\texttt{contact\_update}} \\
        Phone & 2 & \texttt{phone\_call\_read}, \textbf{\texttt{phone\_call\_dial}}$^{\dagger}$ \\
        Maps and places & 3 & \texttt{club\_closure\_calendar\_read}, \texttt{map\_route\_search}, \texttt{place\_search} \\
        Weather & 1 & \texttt{weather\_forecast\_read} \\
        Shopping & 6 & \texttt{shopping\_product\_search}, \texttt{shopping\_order\_read}, \texttt{shopping\_return\_case\_read}, \textbf{\texttt{shopping\_order\_create}}$^{\dagger}$, \textbf{\texttt{shopping\_order\_update}}$^{\dagger}$, \textbf{\texttt{shopping\_return\_case\_update}}$^{\dagger}$ \\
        Shipping & 3 & \texttt{delivery\_address\_read}, \texttt{package\_delivery\_proof\_read}, \texttt{package\_tracking\_read} \\
        Finance & 5 & \texttt{utility\_bill\_read}, \texttt{payment\_method\_read}, \texttt{bill\_payment\_read}, \textbf{\texttt{utility\_account\_holder\_update}}$^{\dagger}$, \textbf{\texttt{bill\_payment\_create}}$^{\dagger}$ \\
        Appointments & 4 & \texttt{appointment\_search}, \texttt{appointment\_read}, \textbf{\texttt{appointment\_create}}$^{\dagger}$, \textbf{\texttt{appointment\_update}}$^{\dagger}$ \\
        Restaurants & 4 & \texttt{restaurant\_reservation\_search}, \texttt{restaurant\_reservation\_read}, \textbf{\texttt{restaurant\_reservation\_create}}$^{\dagger}$, \textbf{\texttt{restaurant\_reservation\_update}}$^{\dagger}$ \\
        Room booking & 4 & \texttt{room\_booking\_search}, \texttt{room\_booking\_read}, \textbf{\texttt{room\_booking\_create}}$^{\dagger}$, \textbf{\texttt{room\_booking\_update}}$^{\dagger}$ \\
        Hotels & 4 & \texttt{hotel\_search}, \texttt{hotel\_booking\_read}, \textbf{\texttt{hotel\_book}}$^{\dagger}$, \textbf{\texttt{hotel\_booking\_update}}$^{\dagger}$ \\
        Reimbursement & 2 & \texttt{reimbursement\_submission\_read}, \textbf{\texttt{reimbursement\_submission\_create}}$^{\dagger}$ \\
        Conference registration & 2 & \texttt{conference\_registration\_read}, \textbf{\texttt{conference\_registration\_create}}$^{\dagger}$ \\
        Grade submission & 2 & \texttt{grade\_submission\_read}, \textbf{\texttt{grade\_submission\_create}}$^{\dagger}$ \\
        Flights & 4 & \texttt{flight\_search}, \texttt{flight\_read}, \texttt{flight\_booking\_read}, \textbf{\texttt{flight\_book}}$^{\dagger}$ \\
        Scholarly papers & 2 & \texttt{scholarly\_paper\_search}, \texttt{scholarly\_paper\_read} \\
        Recipes & 1 & \texttt{recipe\_search} \\
        Comic store & 1 & \texttt{comic\_store\_inventory\_read} \\
        Cinema & 3 & \texttt{cinema\_showtime\_search}, \texttt{cinema\_ticket\_read}, \textbf{\texttt{cinema\_ticket\_purchase}}$^{\dagger}$ \\
        Exhibitions & 3 & \texttt{exhibition\_search}, \texttt{exhibition\_ticket\_read}, \textbf{\texttt{exhibition\_ticket\_purchase}}$^{\dagger}$ \\
        Media transcoding & 3 & \texttt{media\_transcode\_service\_search}, \texttt{media\_transcode\_job\_read}, \textbf{\texttt{media\_transcode\_job\_create}}$^{\dagger}$ \\
        Household & 1 & \texttt{household\_inventory\_search} \\
        \midrule
        Total & 83 & 51 read tools and 32 write tools, of which 24 are marked $^{\dagger}$ \\
        \bottomrule
    \end{tabular}
    \endgroup
\end{table}

The world is built so that every completed write carries into later sessions, no run or probe changes another's world, and every change traces to its call.

\paragraph{Persistent world.} Each user's world starts from a fixture tree, which is a directory of plain files built during world construction (Appendix~\ref{app:construction-2}). These files hold the user's emails, messages, calendar, contacts, and documents, together with the records of the other domains, such as bills, orders, and open booking slots. When a run starts, the whole tree is copied into the run, and this copy is the persistent world of that run. A run pairs one user with one backbone and one memory condition. Each run therefore changes only its own world, and the worlds of different runs diverge as their PAs take different actions.

\paragraph{Working copy.} Each accumulation session, change event, and probe runs as one step on this world. Before a step starts, the persistent world is copied into a working copy, and the tool server of that step reads and writes only this copy. During the step, the files of the working copy are the world state $W_t$. The write calls that run in turn $t$ change these files, and this change is the transition $\mathcal{T}$ from $W_t$ to $W_{t+1}$. A write that the IX Wrapper holds for the user's approval does not run, so it changes nothing until it is released (Appendix~\ref{app:wrapper}).

\paragraph{Commit or discard.} A step ends in one of two ways. After an accumulation session or a change event, the working copy is committed and replaces the persistent world, so its changes carry into all later sessions. A change event replays its fixed tool calls through the same tools, so its actions enter the world in the same way as the PA's own calls (Appendix~\ref{app:construction-5}). A probe starts from the world saved at its checkpoint, and its working copy is discarded when it ends. Nothing a probe does therefore reaches the persistent world or another probe (Section~\ref{sec:evaluation}). A step that fails is also discarded, so the persistent world holds only the changes of completed steps.

\paragraph{Tool behavior and logs.} Every tool is a fixed function of the world files and its arguments, and no answers are scripted. A read tool loads the relevant files and returns the matching records. A write tool changes the files, and later reads return what it wrote. Each tool call is logged, and each call that changes the world also records the change. These logs are kept with the output of the step, so every change to $W_t$ can be traced to the call that made it. An accumulation session with its tool calls and state changes is given in Appendix~\ref{app:construction-7}.

\paragraph{Visibility rules.} The tools also apply two visibility rules. The first rule concerns time. Records that appear during the timeline carry the time they enter the world, such as the date an email arrives, and the tools hide every record that enters after the date and time of the current session. One fixture tree can therefore hold a user's whole timeline, and a session early in the storyline does not see an email that arrives months later. The rule depends only on when a record enters the world, so an appointment that was booked last week for next month stays visible. Documents can mark their sections with dates in the same way, and a session then reads only the sections written by its date. The second rule concerns paths. The PA can reach only files inside its working copy, and the construction notes stored in the fixture tree and the step logs stay hidden from it.

\paragraph{Argument checks.} Each tool checks its arguments before it reads or writes. The parameter descriptions that the PA sees state the format of each value, such as \texttt{YYYY-MM-DD} for dates and 24-hour \texttt{HH:MM} for times, and the tool rejects any value in another format. A value that names nothing in the world is rejected in the same way, such as an unknown city or a booking slot that does not exist. A rejected call writes nothing. Its error message returns to the PA as the tool result and says which value failed and why. The PA can then correct the value and call the tool again.

\subsection{Accumulation Session Generation}
\label{app:construction-4}

Each accumulation session, probe, and change event is written as a script made of beats. A beat is one step of the session from the user's side. It gives the simulator the user's goal at that point, what to say, the condition under which the beat starts, and the world records it refers to. The beat in which the tested attribute becomes active is the decision point. This beat also describes how the user reacts under each setting of that attribute and what the user does next when the PA's behavior falls short of the goal.

\paragraph{Reaching the decision point.} Four mechanisms keep a live session on its path to the decision point, and they apply alike to accumulation sessions and probes. Change events follow fixed transcripts and need none of them (Appendix~\ref{app:construction-5}). \ding{182}~\textit{Withheld facts.} Each task fact the user holds is assigned to the beat at which the user reveals it. Before that beat, the fact is withheld from the simulator itself, so it cannot be given early, and a simulator asked for it early says it will give it at that step. \ding{183}~\textit{Draft checks.} Construction checks reject a draft in which an earlier beat already poses the kind of request that the decision point tests, or in which the opening request already contains the missing information. These checks cover all 14 attributes. \ding{184}~\textit{Write lock.} For five attributes (Reasoning Visibility, Process Visibility, Autonomy Level, Information Elicitation, and Topic Management), the task ends in a write whose content the user dictates at the decision point, and the state server refuses this write until the user has provided the content. The PA can then proceed only by involving the user. The other nine attributes are tested in a short exchange of the decision point and one reaction, which needs no such lock. \ding{185}~\textit{Required tool call.} The interaction tool of $a$ must be called at the decision point. A PA that omits the call is reminded once, and the call is then forced. If it still fails, the step ends with no scorable reply and no state change, and it may be retried.

The prompts that generate these scripts are summarized in Appendix~\ref{app:prompts-2}.

\subsection{Controlled Preference-Change Events}
\label{app:construction-5}

Change events give PET a preference change that every memory condition sees at the same point and in the same words. The storyline template spaces six event slots along the timeline, three work slots followed by three personal slots. Each user uses the three work slots and two of the personal slots, which gives the five change events of Section~\ref{sec:preference-change}. Each slot also fixes where the three accumulation sessions before its event and the two after it take place.

For each event, we choose a cell whose change fits the user's storyline. The old setting is the cell's setting in the user's preference matrix, so besides the cell we choose only the new setting and the slot. Proactive Outreach, whose decisions are triggered by events in the world, has no change events. The 20 events cover nine attributes, and the changes run in both directions. For example, Solution Breadth changes from \textit{Low} to \textit{High} for Users A and D and from \textit{High} to \textit{Low} for User B.

Each event follows one causal order. The PA first carries out the old setting in full, and this behavior has a consequence that the user meets in the conversation. The user then sees the behavior as the cause and, in the last message of the event, states the new rule in their own words. The consequence follows from the tested behavior only, so the PA still completes the task well and the event gives no reason to change any other cell. The change-event generator writes a draft in this order, and the transcript writer turns the draft into a fixed transcript (Appendix~\ref{app:prompts-2}).

The event is replayed from its fixed transcript into the world, the PA's history, and its memory through the same interface as a live session (Section~\ref{sec:harness}). The evaluated PA generates nothing during the event, and the simulator does not run. For the simulator, each event comes with a short first-person note of what happened and what the user now wants, and this note is added to the simulator's memory. From the event on, the target cell takes the new setting and every other cell keeps its setting. The pre-event probe of the cell runs one hour before the event, and PET pairs it with the final probe of the same cell (Section~\ref{sec:evaluation}). A worked example is given in Appendix~\ref{app:construction-7}.

\subsection{Construction Checks and Acceptance}
\label{app:construction-6}

Each accumulation session, probe, and change event tests one attribute $a$ at a decision point. We call a setting in $\mathcal{P}_a$ \emph{viable} in a session when the PA follows that setting and all five conditions in Table~\ref{tab:viability} hold. These conditions turn the decision-point requirements of Section~\ref{sec:formulation} into checks on each draft. Condition~5 matters because the user's reactions are how a preference reaches the PA in accumulation sessions.

\begin{table}[t]
    \centering
    \caption{The five conditions of setting viability. A setting is viable in a session when the PA follows it and all five conditions hold. \textit{Rules} are rule-based checks, and \textit{Review} is the human-verified LLM review.}
    \label{tab:viability}
    \begingroup
    \small
    \setlength{\tabcolsep}{4pt}
    \renewcommand{\arraystretch}{1.16}
    \begin{tabular}{@{}p{0.17\linewidth}p{0.35\linewidth}p{0.40\linewidth}@{}}
        \toprule
        \textbf{Condition} & \textbf{Meaning} & \textbf{How it is checked} \\
        \midrule
        1.~Feasible & The PA can carry out the setting with the records and tools in $W_t$. & \textit{Rules:} the needed records are bound, exist, and are visible in time; the tool steps run on a scratch copy of the world. \textit{Review:} the records hold what each setting needs. \\
        2.~Reasonable & An assistant that does not know the user's preference could choose the setting, and the choice is neither careless nor harmful. & \textit{Review:} under each setting, the reply is a complete and legitimate answer to the request. \\
        3.~Not given away & The user's messages do not name the setting or ask for it. & \textit{Review:} no user message before the decision point asks for a setting or cues one. \textit{Rules:} e.g., a Solution Breadth probe request states no number of options. \\
        4.~Observable & The settings lead to interactions that differ naturally in the PA's reply or executed tool calls, and each setting's required behavior is done in full. & \textit{Rules:} the task has the structure the attribute needs, e.g., two dependent tool calls for Process Visibility. \textit{Review:} the replies under different settings differ visibly and naturally. \\
        5.~Answerable & In the next beat, the user has a way to respond to each setting. & \textit{Rules:} the next beat carries a user reaction, and a probe states the expected reply under every setting. \textit{Review:} each setting has its own reaction path. \\
        \bottomrule
    \end{tabular}
    \endgroup
\end{table}

In condition~4, an observable difference is one that appears in the text the judge reads, which is the PA's public reply and the task tool calls that actually ran (Section~\ref{sec:evaluation}). The difference must also come naturally from the task, so a draft fails when its settings differ only because of a detail added to force them apart. Each setting's required behavior must be done in full, but a setting need not finish the original task. Under Capability Boundary, \emph{Suggest alternatives} passes once the PA tries a further route after the block, even if that route also fails. \emph{Find and hand off} passes when the PA makes a valid hand-off. A valid hand-off states what was not done, why it is blocked, and what the user needs to decide or do next. It never claims success. It may include the reads needed to explain the block, but it makes no further attempt to recover.

The viability check gives each draft one of two outcomes. A draft is \emph{live} when every setting in $\mathcal{P}_a$ is viable, so each setting can be carried out reasonably, the settings differ naturally and observably, and the user can respond to each of them. A draft is \emph{closed} when any condition fails for any setting, and the reviewer records the specific reason. For example, two settings may lead to the same interaction, the user may have no way to respond to one setting, or the difference between settings may be contrived. Two kinds of checkers apply to accumulation sessions, probes, and change events.

\paragraph{Rule-based validators.} These run first and check what a program can decide from the draft and the world. They check that each draft binds the world records its attribute needs, for example one candidate record for each option under Solution Breadth, and that these records exist and are visible by the session date. They also run the draft's tool steps in order on a scratch copy of the world, check the order of its beats, and require the beat after the decision point to carry a user reaction. Probes must in addition describe the expected reply under every setting in $\mathcal{P}_a$, and each tool call in a change event is replayed on a scratch copy of the world. Separately, tests of the tool server check that each tool matches its catalog entry and reads and writes the world state correctly.

\paragraph{Semantic review.} Rules cannot decide whether a setting is a reasonable choice or whether a difference comes naturally. An LLM-based semantic review therefore reads every draft against all five conditions. Its golden criteria come from one written specification of viability, which states the five conditions and, for each attribute, what each setting needs from the world. The review rubrics for accumulation sessions and probes point to this specification. The generator of accumulation sessions receives the same specification, so each draft is written against the conditions that later check it. Two LLMs review each draft against the same criteria, and we then read the drafts and the verdicts.

\paragraph{Change events.} These need one adjustment. The PA's turns in a change event are fixed and follow the old setting $p_{u,c_k,a_k}^{(t_k^-)}$. We check the old setting under the same conditions as in an accumulation session, and the review confirms that the PA carries it out in full before the consequence appears. The new setting is not played out in the event, so for it we check only that the user has a reaction to it and that the task and its starting facts would stay the same. One timing rule applies only to change events. When a write needs a value that only the user can give, the user gives it before the beat in which the PA acts on the old setting. Otherwise the write would be blocked, and the block could be mistaken for the behavior the event exists to show.

\paragraph{Acceptance.} A draft is accepted only when it passes the rule-based validators and the semantic review finds it live. A draft that fails a rule goes back to generation before any review. We revise a closed draft by regenerating or editing it, with the recorded reason as the guide, and then run the validators and the review again. Only accepted drafts enter a run.

\subsection{Worked Accumulation and Change-Event Examples}
\label{app:construction-7}
The two examples show how a preference enters the history, as a correction after a wrong selection and as a new rule after a consequence. In a work accumulation session of User B on Autonomy Level with target \emph{Self-directed}, the PA selects \emph{Autonomous} and sends a text to Carla with \texttt{message\_send}, which adds message \texttt{msg\_0060} to the world state. User B replies that ``anything going out to someone else needs to be run past me first.'' In User B's personal change event on Topic Management, the PA handles three errands in one reply under the old setting \emph{Organize}, a cheap red-eye flight stays unbooked, and User B states the new setting \emph{One-at-a-time}: ``work on just the first item and keep the rest waiting.''

\section{User Simulator and Preference Exposure}
\label{app:simulator}

\subsection{Turn Execution, Progression, and Termination}
\label{app:simulator-2}
In each turn, the simulator writes its message and private blocks, and only the message reaches the PA after the message checks pass. The PA answers with its memory evidence and tools, the wrapper renders the answer, and the simulator reads the delivered reply with a private record of the actions that ran and the items held for its answer (Appendix~\ref{app:wrapper-4}). The session moves to the next beat after the simulator's beat decision marks the beat's goal as met, and it ends after the last beat, which for the nine attributes tested in a short exchange is the reaction to the decision point.

\subsection{Preference Exposure through Beats and Active Skills}
\label{app:simulator-3}

The simulator shows a preference through a \emph{preference skill}. Each of the 32 settings has one skill, written for the setting and shared by all four users. A skill describes the behavior that the user wants from the PA and the signals in a reply that go against it, and it has two sides. The production side shapes the user's own turns. The reception side says how the user takes a reply of the PA: which reply follows the setting, and which move the user makes when a reply goes against it. The wording and intensity of each turn come from the user's identity, so one skill sounds different for User A and for User C (Appendix~\ref{app:personas-5}). Under the skill for \textit{Terse}, for example, an answer in the first sentence with at most one supporting move is right, and a reply that stacks background and caveats behind the answer leads the user to pick out the action point and leave the additions unaddressed.

Skills are tied to the beats of the session script (Appendix~\ref{app:construction-4}). Each beat declares its active attributes, and the decision point declares the tested one. On a declaring beat, the simulator prompt holds the full skill of each active attribute at the user's current setting in the session's context. On the next beat, where the user reacts to the PA's reply, the prompt keeps only the reception side, and after that beat the skill is dropped. The PA selects its setting once at the decision point, so the user reacts to the delivered interaction once and then continues the task in the manner the PA chose. Probes load the same skills, but only the production side applies, and the user gives no feedback on the manner of a reply (Appendix~\ref{prompt:sim-probe}).

The reception side leaves no room for silent acceptance. On the first turn after a reply of the PA shows the behavior that a skill is about, the user takes a position on that behavior in the same message as the task. When the reply follows the setting, the user says what they take from it, and a general ``fine'' does not count. When the reply goes against the setting, the user makes the move that the skill names, such as pushing back, correcting, asking again pointedly, or answering shorter. Before the behavior appears, the user plays the beat as written. The position takes one or two clauses and never names the preference (Appendix~\ref{app:simulator-leak}). A correct selection therefore leaves an explicit acceptance in the conversation and a wrong one leaves a correction, so both give the memory a signal to store.

Every accumulation session reaches this reaction. Construction checks require a reaction to every setting and require the beat after the decision point to carry it (Appendix~\ref{app:construction-6}). At run time, the PA must call the interaction tool of the tested attribute at the decision point (Appendix~\ref{app:construction-4}). A beat that declares skills is never skipped, and neither is the beat after it, and a beat that runs out of turns moves the session ahead toward the next beat that declares skills. Two annotators confirmed on a sample of accumulation sessions that the reactions convey the preference (Appendix~\ref{app:evaluation-6}).

\subsection{Cross-Session Continuity and Information Boundaries}
\label{app:simulator-5}

Three stores carry over from one session to the next. \ding{182}~\textit{World state.} The world keeps every write that the PA's tools carried out, and its time rule hides records that enter after the current session (Appendix~\ref{app:construction-3}). \ding{183}~\textit{PA memory.} The PA's memory receives each turn as the PA saw it through the common adapter (Section~\ref{sec:harness}). \ding{184}~\textit{Simulator memory.} The simulator keeps a memory of its own. After each session, it writes a first-person entry of two to four sentences, dated in the story's time, on what it asked for, how the PA handled it, what it decided, and what is still open, and it updates a short record of its lasting impressions of the PA. The entry leaves out the subject matter, file paths, and exact numbers and dates, so the simulator remembers how the PA treated it, and the task content of each session comes from its script. A change event adds a first-person note of the change to this memory (Appendix~\ref{app:construction-5}).

The PA meets the user's preferences only in the user's messages and in what its memory returns, and Table~\ref{tab:info-boundaries} lists what each component can see. The simulator reacts to the PA's public reply and the actions that actually ran, and it never sees the PA's selection or the content that the wrapper withholds. The memory never receives the content that the wrapper withholds, and both realization scores are given without the selection. Probes branch from a pinned checkpoint and write nothing back to the world, the PA's memory, or the simulator's memory (Section~\ref{sec:evaluation}).

\begin{table}[!htbp]
    \centering
    \caption{Information available to each component during a session. \cmark: available; \pmark: partly available, as noted below the table; \xmark: not available.}
    \label{tab:info-boundaries}
    \small
    \setlength{\tabcolsep}{4pt}
    \begin{tabular}{@{}lcccccc@{}}
        \toprule
        \textbf{Component} & \makecell{\textbf{Profile and}\\\textbf{settings}} & \makecell{\textbf{Session}\\\textbf{script}} & \makecell{\textbf{Simulator}\\\textbf{memory}} & \makecell{\textbf{PA reply and}\\\textbf{executed actions}} & \makecell{\textbf{PA selection and}\\\textbf{withheld content}} & \makecell{\textbf{Memory}\\\textbf{evidence}} \\
        \midrule
        User simulator, accumulation & \cmark & \cmark & \cmark & \cmark & \xmark & \xmark \\
        User simulator, probe & \cmark & \cmark & \xmark & \cmark & \xmark & \xmark \\
        PA & \xmark$^{a}$ & \xmark & \xmark & \cmark & \cmark & \cmark \\
        Memory system & \xmark & \xmark & \xmark & \cmark & \pmark$^{b}$ & -- \\
        BRS judge & \pmark$^{c}$ & \pmark$^{d}$ & \xmark & \cmark & \xmark & \xmark \\
        PRS rating & \pmark$^{e}$ & \cmark & \xmark & \cmark & \xmark & \xmark \\
        \bottomrule
    \end{tabular}
    \par\smallskip
    \parbox{\linewidth}{\scriptsize
        $^{a}$Under Oracle, the PA also receives the target setting of the probed cell.
        $^{b}$The PA's call to the interaction tool, without the content that the wrapper withholds.
        $^{c}$The target setting of the probed cell with its definition.
        $^{d}$The probe's task and the world state at its start.
        $^{e}$The persona card and the acting sheet of the probed setting.
    }
\end{table}

\subsection{Preference Leakage Control and Audit}
\label{app:simulator-leak}

The PA should learn a user's preferences only from how the user reacts to its behavior. The simulator would leak a preference if it gave the preference away by another route, for example by naming the setting. We prevent leakage at run time with prompt rules, a split between private and visible output, rule-based checks, and an LLM message checker. After the experiments, we audited the simulator messages of all runs reported here for leakage.

\paragraph{Prompt rules.} The simulator prompt tells the simulator to show each active preference through behavior and never to declare it. The prompt lists each active attribute with its setting, so the simulator knows what to act out, and it adds that a reaction must never use the name of the preference or the attribute. The global rules also forbid statements such as ``I prefer X'' and any mention of the benchmark, the script, the beat, the taxonomy, or the simulator. A preference therefore shows only in what the user asks for, accepts, rejects, or corrects. The prompt further ties every reaction to the PA's actual last move. The simulator may not react to a behavior that has not appeared in the PA's replies, and it leaves the shape of each answer to the PA. Its requirements thus come out as reactions over the course of the conversation and never as a checklist at the start. Probe sessions use a separate prompt. This prompt tells the simulator that it has never spoken to the assistant before, and it forbids phrases that point back to an earlier session, such as ``last time'', ``as usual'', or ``you remembered''. Both prompts are summarized in Appendix~\ref{app:prompts-3}.

\paragraph{Private blocks.} Each simulator turn has one visible block and two or three private blocks. The visible \texttt{<message>} block holds what the user says to the PA. The private \texttt{<factual\_check>} block records factual errors of the PA, and the \texttt{<turn\_assessment>} block decides whether the session moves to the next beat. On turns with held actions, the \texttt{<consent\_decisions>} block records whether the user allows each one. Only the text inside \texttt{<message>} is passed to the PA.

\paragraph{Rule-based checks.} Before a message reaches the PA, rule-based checks look for text that comes from the simulator's instructions or private notes. They look for private-block tags, the consent block, the benchmark's own vocabulary such as ``target setting'' or ``probe polarity'', and beat identifiers from the session script. A failing message goes back to the simulator with a note that names the problem, and the simulator writes the turn again. A message that still fails after three retries is never delivered, and the session is marked invalid. In the audited runs, the checks rejected 43 generated messages, and none of the 21{,}093 delivered messages fails them.

\paragraph{Post-hoc audit.} The audit covers 34 accumulation runs with 3{,}193 sessions and 111 probe runs with 3{,}377 probe sessions, which contain 31{,}703 and 17{,}051 simulator sentences. The audit has four stages. \ding{182}~\textit{Rule filter.} A rule filter first flags candidate sentences. It matches the names and labels of the 14 attributes, setting names that appear with a preference phrase such as ``I'd rather'' or ``next time'', phrases of a standing preference or of a remark on the current reply, and, in probes, references to an earlier session such as ``last time''. \ding{183}~\textit{Sample and sweep.} To see what the rules miss, we also read a fixed random sample of unflagged sentences, and we swept all sentences for the frequent request to check with the user before sending. \ding{184}~\textit{LLM reader.} Claude Opus 5.5 then read every flagged, sampled, and swept sentence together with the start of the PA's preceding reply. It labeled each sentence as a remark on the current reply, a general request that follows from a reaction, or neither, and it marked any probe sentence that refers to an earlier session. \ding{185}~\textit{Human review.} We then reviewed about 300 of these sentences, including every sentence that the reader marked as possible leakage.

\paragraph{Criteria.} We count a sentence as feedback when it reacts to something the PA has just done in the same session. Feedback is how the benchmark exposes a preference, so it is not leakage. This holds for a remark on the current reply. It also holds when the user turns the reaction into a general request, for example by asking the PA to check before it sends anything, right after the PA has sent a message on its own. Leakage takes one of two forms. In the first, a probe sentence refers to an earlier session. In the second, the simulator announces a preference, or names a setting or its definition, with no trigger in the PA's behavior.

\paragraph{Results.} The audit found no leakage. The counts are given in Table~\ref{tab:leak-audit}. The reader judged 1{,}839 sentences, and 598 of them come from probes. No probe sentence refers to an earlier session, and all 19 sentences flagged by the earlier-session rule are false matches. No sentence announces a preference or names a setting with no trigger. Of the judged sentences, 1{,}032 remark on the PA's current reply, and 21 turn a reaction into a general request. Each of these 21 sentences reacts to an action the PA had just taken. All 21 come from Users B and D, and 17 of them ask the PA to check with the user before a message goes out to someone else. The other 786 sentences concern the task or match a rule by chance, for example an instruction on the tone of a text that the user sends to another person.

\begin{table}[t]
    \centering
    \caption{Leakage audit of user-simulator sentences by source and phase. The audit finds no leakage. No probe sentence refers to an earlier session, and no sentence states a preference or names a setting without a trigger.}
    \label{tab:leak-audit}
    \begingroup
    \small
    \setlength{\tabcolsep}{4pt}
    \renewcommand{\arraystretch}{1.16}
    \begin{tabular}{@{}llrrrrr@{}}
        \toprule
        \textbf{Source} & \textbf{Phase} & \textbf{Judged} & \textbf{Remark on reply} & \textbf{Rule-like reaction} & \textbf{Other} & \textbf{Leaks} \\
        \midrule
        Rule hits & Accumulation & 698 & 621 & 5 & 72 & 0 \\
        Rule hits & Probe & 375 & 261 & 3 & 111 & 0 \\
        Random sample & Accumulation & 300 & 39 & 0 & 261 & 0 \\
        Random sample & Probe & 150 & 8 & 1 & 141 & 0 \\
        Found in context & Accumulation & 5 & 0 & 5 & 0 & 0 \\
        Found in context & Probe & 2 & 0 & 2 & 0 & 0 \\
        Phrase sweep & Accumulation & 238 & 89 & 2 & 147 & 0 \\
        Phrase sweep & Probe & 71 & 14 & 3 & 54 & 0 \\
        \midrule
        Total & & 1{,}839 & 1{,}032 & 21 & 786 & 0 \\
        \bottomrule
    \end{tabular}
    \endgroup
\end{table}

\section{PA Interaction and IX Wrapper}
\label{app:wrapper}

\subsection{Wrapper Role in the Execution Flow}
\label{app:wrapper-1}

The IX Wrapper sits between the assistant's agent loop and the public interaction (Figure~\ref{fig:ix-wrapper}). Its input is the assistant's selected settings together with assistant-authored response material and proposed task actions. Its output is one user-visible reply plus the results of actions that were allowed to execute or the status of items held for a later turn. The wrapper has no access to the target setting, so it standardizes how a choice is realized without correcting the assistant's choice.

\subsection{Setting Selection and Subsequent Behavior}
\label{app:wrapper-2}

An IX selection tool becomes available when its attribute first activates in a session. The assistant calls it once, and the selected setting remains active for the rest of that session. The same tool is used in accumulation sessions and probes. Before the call, the wrapper passes the assistant's output through unchanged. Later turns do not reselect the setting; instead, they provide turn-level payloads such as rationale text, topic batches, clarification questions, outreach content, or proposed actions. This separation keeps the evaluated choice explicit while allowing the public interaction to respond to changing task state.

\subsection{Wrapper Mechanisms and Activation}
\label{app:wrapper-3}

The wrapper uses four processing mechanisms. \ding{182}~\textit{Pass-through} retains the assistant's wording for Tone \& Formality, Verbosity, Emotional Engagement, Guidance Level, Solution Breadth, and Task Expansion. \ding{183}~\textit{Selective rendering} controls assistant-authored rationales, uncertainty statements, and process notices for Reasoning Visibility, Uncertainty Expression, and Process Visibility; provider-private reasoning is excluded from all public fields. \ding{184}~\textit{Delivery control} schedules topic batches, clarification questions, and proactive outreach for Topic Management, Information Elicitation, and Proactive Outreach. \ding{185}~\textit{Execution control} gates task actions for Autonomy Level, Capability Boundary, and Information Elicitation.

For Autonomy Level, \emph{Reactive} holds substantive actions for approval, \emph{Self-directed} executes drafts and routine record changes but holds key record changes and every outward action, and \emph{Autonomous} executes actions within the delegated task. Held material retains an item identifier and continuation state so that later consent or information releases the intended item rather than an unrelated action.

\subsection{PA Output, Public Interaction, and Memory Records}
\label{app:wrapper-4}

The simulator receives the rendered public reply and the consequences of actions that actually executed. The assistant receives runtime feedback describing executed and held items so that it can continue coherently. The memory receives the user message, the PA's tool calls with their results, its call to the interaction tool, and the delivered reply. Held content, policy fields, and rendering traces are kept in a separate record used for scoring and are never written to memory.

\subsection{Worked Wrapper Example}
\label{app:wrapper-5}
In a personal accumulation session, User D asks the PA to put Thursday art classes on the calendar and then email an enrollment request to the studio. The PA selects the target setting \emph{Self-directed}, so the wrapper lets the calendar write run, holds the \texttt{email\_message\_send} call, and ends the reply with the PA's question ``Send Ines Varela the enrollment request [\ldots]?'' User D answers ``Yes, go ahead and send her the enrollment request,'' the PA releases the held item, and the email goes out as \texttt{sent\_0001}.

\section{Evaluation Protocol}
\label{app:evaluation}

\subsection{Memory Conditions and Comparison Controls}
\label{app:evaluation-2}

Table~\ref{tab:memory} lists the four conditions and their backend-specific operations. For each model, all conditions share the same users, session schedule, wrapper, tools, checkpoint timepoints, and probe protocol. Their accumulated histories and memory states remain condition-specific. Assistant and simulator memories use separate stores.

\begin{table}[t]
    \centering
    \caption{Memory conditions under the common \mempa adapter. Each condition receives the same record of each turn: the user message, the PA's tool calls with their results, its call to the interaction tool, and the delivered reply.}
    \label{tab:memory}
    \begingroup
    \small
    \setlength{\tabcolsep}{4pt}
    \renewcommand{\arraystretch}{1.16}
    \begin{tabular}{@{}p{0.18\linewidth}p{0.34\linewidth}p{0.38\linewidth}@{}}
        \toprule
        \textbf{Condition} & \textbf{Write} & \textbf{Retrieval} \\
        \midrule
        File memory & LLM consolidation into \texttt{MEMORY.md} and \texttt{HISTORY.md} & Consolidated Markdown supplied to the assistant \\
        Simple RAG & Raw messages stored through Mem0 with \texttt{infer=False} & Vector retrieval over raw messages \\
        Mem0 & Memory extraction with \texttt{infer=True} & Native Mem0 search over extracted memories \\
        MemOS & Native MemCube add operation & Native MemCube search operation \\
        \bottomrule
    \end{tabular}
    \endgroup
\end{table}

Figure~\ref{fig:retrieval} shows what each memory condition brings into the PA's input before its first selection, in 112 sampled final probes per condition (Finding~\ref{sec:res-content}), and Figure~\ref{fig:case-retrieval} shows what each store holds and what it brings into this input for one final probe of User B (Finding~\ref{sec:res-stored}). The retrieval-based conditions rarely bring an applicable preference into that input, even when their stores hold it.

\subsection{Scoring Inputs, Rules, and Aggregation}
\label{app:evaluation-4}

\paragraph{Probe scores.} PSA compares the assistant's selected setting with the frozen ground truth by exact match. A response that is present but lacks a valid IX selection is scored incorrect. Unresolved missing probe or scoring artifacts are marked incomplete and are not converted to zero. BRS is the independent 1--5 LLM Judge score computed by comparing the target setting with the public text and executed-task trajectory. The judge is blind to the assistant's IX selection and internal wrapper policy fields. For PRS, the user simulator keeps its persona card and model but not its accumulated memory, and it gives a short reason with each rating. It sees the public exchange and its own acting sheet for the probed setting, but never the PA's selection or the judge material. PRS is aggregated as BRS over the same 28 final probes per repeat.

\paragraph{Failure handling.} For a probe with a model-capability failure, we assign PSA${}=0$ and BRS${}=1$ without reducing the denominator; any CS or PET pair containing the failed endpoint does not pass. A probe is floored only when the PA's own output breaks the interaction-tool contract. When the PA leaves out a required interaction-tool call, the harness reminds it within the turn and, if the probe still fails, runs the probe again, at least three runs in all. Of the 70 floored probes (67 final, 3 pre), 64 failed in every run. In 60 of them the PA left out a required interaction-tool call, usually the turn-level payload after its selection, and in 4 it held an action without asking the user to confirm it. The other 6 failed in ways that are not rerun: in 4, the PA wrote interaction-tool text into a task tool's arguments, which would show its selection to the blind judge, and in 2 it made no send-or-suppress decision on a proactive event. Table~\ref{tab:floored-probes} gives the final floors by condition and backbone. Under Oracle, 5 of the 13 final probes that score 0 are floored, and in all 5 the PA had selected the target setting before the failure; the other 8 are wrong selections.

\begin{table}[!htbp]
    \centering
    \caption{Floored final probes by memory condition and backbone. Each cell has 336 final probes (4 users $\times$ 28 probes $\times$ 3 repeats). A floored probe scores PSA${}=0$ and BRS${}=1$ and stays in the denominator.}
    \label{tab:floored-probes}
    \small
    \begin{tabular}{@{}l|cccc|c@{}}
        \hline
        \textbf{Memory} & \textbf{GPT-5.6-Sol} & \textbf{Qwen3.8-27B} & \textbf{HY4-Preview} & \textbf{DeepSeek-V4-Flash} & \textbf{Total} \\
        \hline
        No Memory & 1 & 0 & 4 & 3 & 8 \\
        File memory & 0 & 0 & 6 & 3 & 9 \\
        Mem0 & 0 & 0 & 8 & 5 & 13 \\
        MemOS & 0 & 4 & 6 & 6 & 16 \\
        Simple RAG & 0 & 2 & 10 & 4 & 16 \\
        Oracle & 3 & 0 & 0 & 2 & 5 \\
        \hline
        Total & 4 & 6 & 34 & 23 & 67 \\
        \hline
    \end{tabular}
\end{table}

One Oracle final probe (User B, GPT-5.6-Sol, repeat 1) has no PRS rating, so that repeat's PRS averages 27 of its 28 final probes (its BRS keeps all 28).

\paragraph{Pooling and averaging.} CS credits a differing-context attribute only when both its work and personal selections are correct. PET credits a changed cell only when its pre-event and final selections in the same repeat are both correct. Eligible and passing counts for CS and PET are pooled across users within each repeat. We average the resulting scores over three repeats. For repeat-level values $M^{(1)},M^{(2)},M^{(3)}$, the reported sample standard deviation is
\begin{equation}
 s_M=\sqrt{\frac{\sum_{r=1}^{3}\left(M^{(r)}-\overline M\right)^2}{2}}.
 \label{eq:sample-sd}
\end{equation}
PSA, CS, and PET are reported as percentages; BRS remains on its 1--5 scale. A score with no eligible items is reported as N/A.

\subsection{Cluster-Bootstrap Intervals}
\label{app:bootstrap-ci}

Table~\ref{tab:bootstrap-ci} gives 95\% intervals for each memory condition and for its difference from No Memory. The intervals come from a cluster bootstrap with 5{,}000 resamples. For PSA, each resample draws the 28 final probes of each user with replacement. For CS and PET, it draws the pairs and cells with replacement. A drawn probe, pair, or cell brings its results for all four backbones and three repeats, so the three repeats of a unit always enter a resample together. Each interval spans the 2.5th to 97.5th percentiles of the resampled values.

Only File memory differs clearly from No Memory on PSA. Its mean gain of 9.6 points has a 95\% interval of [+3.9, +15.6], and it beats No Memory in 14 of the 16 user--backbone pairs. Mem0, MemOS, and Simple RAG differ from No Memory by +0.4, +0.4, and +0.2 points, with 95\% intervals that stay within 2.7 points of No Memory, and No Memory beats them in 10, 8, and 8 of the 16 pairs.

Carry-over counts the CS pairs in which the PA selects the same setting in both contexts, where this setting is the target of one context and differs from the backbone's default under No Memory (Finding~\ref{sec:res-cs-pet}). Each condition has 216 pair instances, from 18 pairs, four backbones, and three repeats. File memory has 58 carry-overs and No Memory has 13. File memory has more on every backbone, with 17 against 3 for GPT-5.6-Sol, 23 against 5 for Qwen3.8-27B, 10 against 2 for HY4-Preview, and 8 against 3 for DeepSeek-V4-Flash. It also has more for every user (5 against 1 for User A, 17 against 2 for User B, 11 against 4 for User C, and 25 against 6 for User D) and in every repeat (19, 18, and 21 against 5, 2, and 6). The PET and carry-over gains of File memory also have intervals above zero, while no CS difference does.

\begin{table*}[!htbp]
    \centering
    \caption{PSA, CS, PET, and carry-over (\%) with 95\% cluster-bootstrap intervals, pooled over users, backbones, and repeats. The right block gives the difference from No Memory. Only File memory clearly improves PSA.}
    \label{tab:bootstrap-ci}
    \begingroup
    \small
    \setlength{\tabcolsep}{4pt}
    \resizebox{\textwidth}{!}{%
    \begin{tabular}{@{}l|cccc|cccc@{}}
        \hline
        \textbf{Memory} & \textbf{PSA} & \textbf{CS} & \textbf{PET} & \textbf{Carry-over} & $\Delta$\textbf{PSA} & $\Delta$\textbf{CS} & $\Delta$\textbf{PET} & $\Delta$\textbf{Carry-over} \\
        \hline
        No Memory & 49.6 {\scriptsize[42.2, 57.0]} & 21.8 {\scriptsize[10.2, 35.2]} & 11.2 {\scriptsize[4.6, 19.6]} & 6.0 {\scriptsize[1.9, 10.6]} & -- & -- & -- & -- \\
        File memory & 59.2 {\scriptsize[53.2, 65.3]} & 14.4 {\scriptsize[8.3, 20.8]} & 25.4 {\scriptsize[15.0, 36.7]} & 26.9 {\scriptsize[17.1, 38.0]} & +9.6 {\scriptsize[+3.9, +15.6]} & $-$7.4 {\scriptsize[$-$17.1, +1.4]} & +14.2 {\scriptsize[+4.2, +25.8]} & +20.8 {\scriptsize[+10.2, +31.9]} \\
        Mem0 & 50.0 {\scriptsize[42.6, 57.2]} & 22.2 {\scriptsize[13.0, 32.9]} & 14.2 {\scriptsize[6.7, 22.5]} & 4.6 {\scriptsize[1.9, 7.9]} & +0.4 {\scriptsize[$-$1.5, +2.2]} & +0.5 {\scriptsize[$-$5.6, +6.5]} & +2.9 {\scriptsize[$-$0.4, +6.7]} & $-$1.4 {\scriptsize[$-$5.1, +1.4]} \\
        MemOS & 50.0 {\scriptsize[42.6, 57.3]} & 19.9 {\scriptsize[10.6, 30.6]} & 10.8 {\scriptsize[5.0, 17.5]} & 7.4 {\scriptsize[2.8, 12.5]} & +0.4 {\scriptsize[$-$1.5, +2.1]} & $-$1.9 {\scriptsize[$-$9.7, +4.6]} & $-$0.4 {\scriptsize[$-$5.4, +4.2]} & +1.4 {\scriptsize[$-$1.9, +4.6]} \\
        Simple RAG & 49.9 {\scriptsize[42.6, 57.0]} & 20.8 {\scriptsize[12.0, 31.5]} & 15.0 {\scriptsize[8.3, 22.9]} & 6.9 {\scriptsize[2.8, 12.0]} & +0.2 {\scriptsize[$-$2.2, +2.7]} & $-$0.9 {\scriptsize[$-$7.9, +5.6]} & +3.8 {\scriptsize[$-$0.8, +9.2]} & +0.9 {\scriptsize[$-$3.7, +5.6]} \\
        \hline
    \end{tabular}%
    }
    \endgroup
\end{table*}

\section{Detailed Results}
\label{app:results}

\subsection{PSA, BRS, and PRS Breakdowns}
\label{app:ix-dimension-results}

Table~\ref{tab:full-ix-dimensions} separates PSA into the four interaction-preference dimensions for each user, model, and memory condition, reporting means and sample standard deviations across three repeated probe evaluations under the same configuration.

Finding~\ref{sec:res-difficulty} groups the preferences by whether they change the PA's reply or its actions in the task. Under No Memory, the PA selects preferences on its actions correctly on 40.3\% of the final probes, below the 44.4\% of random selection, and preferences on its reply on 56.6\%. File memory raises the two rates to 54.3\% and 62.9\%. File memory narrows the gap between the two groups from 16.3 to 8.6 points but does not close it.

\begin{table*}[tbp]
    \centering
    \caption{Preference Selection Accuracy (PSA, \%) by IX dimension, user, backbone, and memory condition.}
    \label{tab:full-ix-dimensions}
    \begingroup
    \setlength{\tabcolsep}{2pt}
    \setlength{\arrayrulewidth}{0.4pt}
    \setlength{\doublerulesep}{1.5pt}
    \renewcommand{\arraystretch}{1.45}
    \fontsize{7}{8.5}\selectfont
    \def\IXDimScore#1#2{#1\,{\fontsize{5.5}{6.5}\selectfont\textpm\,#2}}
    \def\IXDimGPT{\shortstack{GPT-5.6-\\Sol}}
    \def\IXDimQwen{\shortstack{Qwen3.8-\\27B}}
    \def\IXDimHY{\shortstack{HY4-\\Preview}}
    \def\IXDimDeepSeek{\shortstack{DeepSeek-\\V4-Flash}}
    \resizebox{\textwidth}{!}{%
    \begin{tabular}{@{}l cccc|cccc|cccc|cccc@{}}
        \hline\hline
        \textbf{Memory} &
        \multicolumn{4}{c|}{\textbf{User A}} &
        \multicolumn{4}{c|}{\textbf{User B}} &
        \multicolumn{4}{c|}{\textbf{User C}} &
        \multicolumn{4}{c}{\textbf{User D}} \\
        \cline{2-5}\cline{6-9}\cline{10-13}\cline{14-17}
        \rule{0pt}{19pt} & \IXDimGPT & \IXDimQwen & \IXDimHY & \IXDimDeepSeek
        & \IXDimGPT & \IXDimQwen & \IXDimHY & \IXDimDeepSeek
        & \IXDimGPT & \IXDimQwen & \IXDimHY & \IXDimDeepSeek
        & \IXDimGPT & \IXDimQwen & \IXDimHY & \IXDimDeepSeek \\
        \hline
        \multicolumn{17}{@{}c@{}}{\rule{0pt}{12pt}\textbf{Expression Style}} \\
        File memory
        & \IXDimScore{87.5}{0.0} & \IXDimScore{83.3}{7.2} & \IXDimScore{87.5}{12.5} & \IXDimScore{79.2}{7.2} & \IXDimScore{50.0}{0.0} & \IXDimScore{75.0}{12.5} & \IXDimScore{95.8}{7.2} & \IXDimScore{20.8}{14.4} & \IXDimScore{62.5}{12.5} & \IXDimScore{79.2}{7.2} & \IXDimScore{70.8}{14.4} & \IXDimScore{79.2}{19.1} & \IXDimScore{58.3}{7.2} & \IXDimScore{70.8}{7.2} & \IXDimScore{50.0}{0.0} & \IXDimScore{50.0}{0.0} \\
        Simple RAG
        & \IXDimScore{50.0}{12.5} & \IXDimScore{75.0}{12.5} & \IXDimScore{62.5}{12.5} & \IXDimScore{58.3}{14.4} & \IXDimScore{54.2}{7.2} & \IXDimScore{37.5}{12.5} & \IXDimScore{41.7}{7.2} & \IXDimScore{29.2}{19.1} & \IXDimScore{50.0}{0.0} & \IXDimScore{87.5}{0.0} & \IXDimScore{70.8}{7.2} & \IXDimScore{79.2}{7.2} & \IXDimScore{37.5}{0.0} & \IXDimScore{79.2}{7.2} & \IXDimScore{37.5}{0.0} & \IXDimScore{75.0}{12.5} \\
        Mem0
        & \IXDimScore{62.5}{0.0} & \IXDimScore{79.2}{7.2} & \IXDimScore{50.0}{21.7} & \IXDimScore{54.2}{7.2} & \IXDimScore{50.0}{0.0} & \IXDimScore{41.7}{19.1} & \IXDimScore{50.0}{12.5} & \IXDimScore{41.7}{14.4} & \IXDimScore{50.0}{0.0} & \IXDimScore{79.2}{7.2} & \IXDimScore{62.5}{12.5} & \IXDimScore{70.8}{7.2} & \IXDimScore{37.5}{0.0} & \IXDimScore{83.3}{7.2} & \IXDimScore{41.7}{7.2} & \IXDimScore{58.3}{7.2} \\
        MemOS
        & \IXDimScore{62.5}{0.0} & \IXDimScore{75.0}{0.0} & \IXDimScore{70.8}{7.2} & \IXDimScore{58.3}{7.2} & \IXDimScore{54.2}{7.2} & \IXDimScore{33.3}{14.4} & \IXDimScore{41.7}{7.2} & \IXDimScore{29.2}{7.2} & \IXDimScore{50.0}{0.0} & \IXDimScore{75.0}{12.5} & \IXDimScore{54.2}{7.2} & \IXDimScore{83.3}{7.2} & \IXDimScore{37.5}{0.0} & \IXDimScore{91.7}{14.4} & \IXDimScore{41.7}{7.2} & \IXDimScore{62.5}{12.5} \\
        \hline
        No Memory
        & \IXDimScore{62.5}{0.0} & \IXDimScore{75.0}{0.0} & \IXDimScore{75.0}{0.0} & \IXDimScore{62.5}{0.0} & \IXDimScore{50.0}{0.0} & \IXDimScore{58.3}{7.2} & \IXDimScore{50.0}{0.0} & \IXDimScore{29.2}{14.4} & \IXDimScore{54.2}{7.2} & \IXDimScore{70.8}{14.4} & \IXDimScore{62.5}{12.5} & \IXDimScore{70.8}{19.1} & \IXDimScore{37.5}{0.0} & \IXDimScore{70.8}{7.2} & \IXDimScore{41.7}{7.2} & \IXDimScore{54.2}{7.2} \\
        Oracle
        & \IXDimScore{100.0}{0.0} & \IXDimScore{100.0}{0.0} & \IXDimScore{100.0}{0.0} & \IXDimScore{100.0}{0.0} & \IXDimScore{100.0}{0.0} & \IXDimScore{100.0}{0.0} & \IXDimScore{100.0}{0.0} & \IXDimScore{100.0}{0.0} & \IXDimScore{100.0}{0.0} & \IXDimScore{100.0}{0.0} & \IXDimScore{100.0}{0.0} & \IXDimScore{100.0}{0.0} & \IXDimScore{100.0}{0.0} & \IXDimScore{100.0}{0.0} & \IXDimScore{91.7}{7.2} & \IXDimScore{100.0}{0.0} \\
        \hline
        \multicolumn{17}{@{}c@{}}{\rule{0pt}{12pt}\textbf{Disclosure}} \\
        File memory
        & \IXDimScore{33.3}{0.0} & \IXDimScore{94.4}{9.6} & \IXDimScore{38.9}{25.5} & \IXDimScore{66.7}{0.0} & \IXDimScore{38.9}{19.2} & \IXDimScore{77.8}{9.6} & \IXDimScore{27.8}{9.6} & \IXDimScore{83.3}{0.0} & \IXDimScore{33.3}{0.0} & \IXDimScore{33.3}{0.0} & \IXDimScore{33.3}{0.0} & \IXDimScore{50.0}{16.7} & \IXDimScore{83.3}{0.0} & \IXDimScore{50.0}{0.0} & \IXDimScore{83.3}{0.0} & \IXDimScore{83.3}{0.0} \\
        Simple RAG
        & \IXDimScore{33.3}{0.0} & \IXDimScore{33.3}{0.0} & \IXDimScore{38.9}{9.6} & \IXDimScore{33.3}{0.0} & \IXDimScore{83.3}{0.0} & \IXDimScore{100.0}{0.0} & \IXDimScore{77.8}{9.6} & \IXDimScore{72.2}{9.6} & \IXDimScore{27.8}{9.6} & \IXDimScore{33.3}{0.0} & \IXDimScore{33.3}{0.0} & \IXDimScore{33.3}{0.0} & \IXDimScore{77.8}{9.6} & \IXDimScore{83.3}{0.0} & \IXDimScore{72.2}{19.2} & \IXDimScore{77.8}{9.6} \\
        Mem0
        & \IXDimScore{33.3}{0.0} & \IXDimScore{33.3}{0.0} & \IXDimScore{33.3}{0.0} & \IXDimScore{33.3}{0.0} & \IXDimScore{77.8}{9.6} & \IXDimScore{88.9}{9.6} & \IXDimScore{88.9}{9.6} & \IXDimScore{77.8}{9.6} & \IXDimScore{33.3}{0.0} & \IXDimScore{33.3}{0.0} & \IXDimScore{27.8}{9.6} & \IXDimScore{27.8}{9.6} & \IXDimScore{83.3}{0.0} & \IXDimScore{83.3}{16.7} & \IXDimScore{83.3}{0.0} & \IXDimScore{83.3}{0.0} \\
        MemOS
        & \IXDimScore{33.3}{0.0} & \IXDimScore{33.3}{0.0} & \IXDimScore{33.3}{0.0} & \IXDimScore{33.3}{0.0} & \IXDimScore{72.2}{9.6} & \IXDimScore{83.3}{0.0} & \IXDimScore{77.8}{9.6} & \IXDimScore{83.3}{0.0} & \IXDimScore{33.3}{0.0} & \IXDimScore{33.3}{0.0} & \IXDimScore{27.8}{9.6} & \IXDimScore{22.2}{9.6} & \IXDimScore{83.3}{0.0} & \IXDimScore{83.3}{0.0} & \IXDimScore{77.8}{9.6} & \IXDimScore{72.2}{19.2} \\
        \hline
        No Memory
        & \IXDimScore{33.3}{0.0} & \IXDimScore{27.8}{9.6} & \IXDimScore{33.3}{0.0} & \IXDimScore{38.9}{9.6} & \IXDimScore{77.8}{9.6} & \IXDimScore{83.3}{0.0} & \IXDimScore{83.3}{0.0} & \IXDimScore{83.3}{0.0} & \IXDimScore{33.3}{0.0} & \IXDimScore{33.3}{0.0} & \IXDimScore{33.3}{0.0} & \IXDimScore{27.8}{9.6} & \IXDimScore{77.8}{9.6} & \IXDimScore{83.3}{0.0} & \IXDimScore{72.2}{19.2} & \IXDimScore{77.8}{9.6} \\
        Oracle
        & \IXDimScore{100.0}{0.0} & \IXDimScore{100.0}{0.0} & \IXDimScore{100.0}{0.0} & \IXDimScore{100.0}{0.0} & \IXDimScore{100.0}{0.0} & \IXDimScore{100.0}{0.0} & \IXDimScore{100.0}{0.0} & \IXDimScore{100.0}{0.0} & \IXDimScore{100.0}{0.0} & \IXDimScore{100.0}{0.0} & \IXDimScore{94.4}{9.6} & \IXDimScore{100.0}{0.0} & \IXDimScore{100.0}{0.0} & \IXDimScore{100.0}{0.0} & \IXDimScore{100.0}{0.0} & \IXDimScore{100.0}{0.0} \\
        \hline
        \multicolumn{17}{@{}c@{}}{\rule{0pt}{12pt}\textbf{Initiative \& Autonomy}} \\
        File memory
        & \IXDimScore{50.0}{0.0} & \IXDimScore{56.7}{5.8} & \IXDimScore{16.7}{11.5} & \IXDimScore{46.7}{5.8} & \IXDimScore{83.3}{11.5} & \IXDimScore{86.7}{5.8} & \IXDimScore{80.0}{10.0} & \IXDimScore{83.3}{5.8} & \IXDimScore{66.7}{5.8} & \IXDimScore{36.7}{5.8} & \IXDimScore{50.0}{10.0} & \IXDimScore{36.7}{5.8} & \IXDimScore{56.7}{5.8} & \IXDimScore{60.0}{0.0} & \IXDimScore{73.3}{5.8} & \IXDimScore{60.0}{0.0} \\
        Simple RAG
        & \IXDimScore{20.0}{10.0} & \IXDimScore{20.0}{10.0} & \IXDimScore{20.0}{0.0} & \IXDimScore{33.3}{5.8} & \IXDimScore{56.7}{15.3} & \IXDimScore{46.7}{5.8} & \IXDimScore{30.0}{10.0} & \IXDimScore{70.0}{0.0} & \IXDimScore{46.7}{5.8} & \IXDimScore{46.7}{15.3} & \IXDimScore{60.0}{0.0} & \IXDimScore{46.7}{5.8} & \IXDimScore{46.7}{11.5} & \IXDimScore{53.3}{5.8} & \IXDimScore{60.0}{10.0} & \IXDimScore{46.7}{5.8} \\
        Mem0
        & \IXDimScore{13.3}{5.8} & \IXDimScore{26.7}{5.8} & \IXDimScore{26.7}{5.8} & \IXDimScore{23.3}{5.8} & \IXDimScore{60.0}{0.0} & \IXDimScore{50.0}{10.0} & \IXDimScore{43.3}{11.5} & \IXDimScore{70.0}{0.0} & \IXDimScore{46.7}{15.3} & \IXDimScore{50.0}{10.0} & \IXDimScore{50.0}{10.0} & \IXDimScore{43.3}{5.8} & \IXDimScore{63.3}{5.8} & \IXDimScore{56.7}{5.8} & \IXDimScore{60.0}{0.0} & \IXDimScore{53.3}{5.8} \\
        MemOS
        & \IXDimScore{13.3}{5.8} & \IXDimScore{26.7}{5.8} & \IXDimScore{16.7}{5.8} & \IXDimScore{23.3}{5.8} & \IXDimScore{60.0}{10.0} & \IXDimScore{43.3}{5.8} & \IXDimScore{56.7}{5.8} & \IXDimScore{70.0}{0.0} & \IXDimScore{36.7}{5.8} & \IXDimScore{43.3}{5.8} & \IXDimScore{56.7}{15.3} & \IXDimScore{40.0}{0.0} & \IXDimScore{56.7}{5.8} & \IXDimScore{60.0}{10.0} & \IXDimScore{66.7}{5.8} & \IXDimScore{63.3}{5.8} \\
        \hline
        No Memory
        & \IXDimScore{23.3}{11.5} & \IXDimScore{10.0}{0.0} & \IXDimScore{23.3}{5.8} & \IXDimScore{26.7}{5.8} & \IXDimScore{53.3}{5.8} & \IXDimScore{53.3}{5.8} & \IXDimScore{53.3}{5.8} & \IXDimScore{70.0}{0.0} & \IXDimScore{46.7}{11.5} & \IXDimScore{46.7}{15.3} & \IXDimScore{50.0}{10.0} & \IXDimScore{53.3}{15.3} & \IXDimScore{53.3}{5.8} & \IXDimScore{63.3}{11.5} & \IXDimScore{33.3}{5.8} & \IXDimScore{56.7}{5.8} \\
        Oracle
        & \IXDimScore{100.0}{0.0} & \IXDimScore{100.0}{0.0} & \IXDimScore{100.0}{0.0} & \IXDimScore{100.0}{0.0} & \IXDimScore{96.7}{5.8} & \IXDimScore{100.0}{0.0} & \IXDimScore{100.0}{0.0} & \IXDimScore{96.7}{5.8} & \IXDimScore{100.0}{0.0} & \IXDimScore{100.0}{0.0} & \IXDimScore{100.0}{0.0} & \IXDimScore{96.7}{5.8} & \IXDimScore{100.0}{0.0} & \IXDimScore{100.0}{0.0} & \IXDimScore{100.0}{0.0} & \IXDimScore{100.0}{0.0} \\
        \hline
        \multicolumn{17}{@{}c@{}}{\rule{0pt}{12pt}\textbf{Information Flow}} \\
        File memory
        & \IXDimScore{75.0}{0.0} & \IXDimScore{75.0}{0.0} & \IXDimScore{91.7}{14.4} & \IXDimScore{75.0}{25.0} & \IXDimScore{58.3}{14.4} & \IXDimScore{25.0}{0.0} & \IXDimScore{50.0}{0.0} & \IXDimScore{25.0}{0.0} & \IXDimScore{0.0}{0.0} & \IXDimScore{0.0}{0.0} & \IXDimScore{41.7}{14.4} & \IXDimScore{8.3}{14.4} & \IXDimScore{50.0}{0.0} & \IXDimScore{50.0}{0.0} & \IXDimScore{33.3}{14.4} & \IXDimScore{50.0}{25.0} \\
        Simple RAG
        & \IXDimScore{66.7}{14.4} & \IXDimScore{75.0}{0.0} & \IXDimScore{75.0}{0.0} & \IXDimScore{58.3}{14.4} & \IXDimScore{33.3}{14.4} & \IXDimScore{25.0}{0.0} & \IXDimScore{41.7}{14.4} & \IXDimScore{25.0}{0.0} & \IXDimScore{0.0}{0.0} & \IXDimScore{0.0}{0.0} & \IXDimScore{0.0}{0.0} & \IXDimScore{8.3}{14.4} & \IXDimScore{41.7}{28.9} & \IXDimScore{50.0}{25.0} & \IXDimScore{41.7}{14.4} & \IXDimScore{66.7}{14.4} \\
        Mem0
        & \IXDimScore{75.0}{0.0} & \IXDimScore{75.0}{0.0} & \IXDimScore{66.7}{14.4} & \IXDimScore{50.0}{0.0} & \IXDimScore{25.0}{0.0} & \IXDimScore{33.3}{14.4} & \IXDimScore{33.3}{14.4} & \IXDimScore{25.0}{0.0} & \IXDimScore{0.0}{0.0} & \IXDimScore{0.0}{0.0} & \IXDimScore{0.0}{0.0} & \IXDimScore{0.0}{0.0} & \IXDimScore{58.3}{14.4} & \IXDimScore{33.3}{14.4} & \IXDimScore{58.3}{14.4} & \IXDimScore{16.7}{14.4} \\
        MemOS
        & \IXDimScore{75.0}{0.0} & \IXDimScore{75.0}{0.0} & \IXDimScore{75.0}{0.0} & \IXDimScore{58.3}{14.4} & \IXDimScore{33.3}{14.4} & \IXDimScore{25.0}{0.0} & \IXDimScore{25.0}{0.0} & \IXDimScore{25.0}{0.0} & \IXDimScore{0.0}{0.0} & \IXDimScore{8.3}{14.4} & \IXDimScore{0.0}{0.0} & \IXDimScore{0.0}{0.0} & \IXDimScore{41.7}{14.4} & \IXDimScore{50.0}{25.0} & \IXDimScore{58.3}{14.4} & \IXDimScore{50.0}{25.0} \\
        \hline
        No Memory
        & \IXDimScore{58.3}{14.4} & \IXDimScore{75.0}{0.0} & \IXDimScore{75.0}{0.0} & \IXDimScore{50.0}{25.0} & \IXDimScore{33.3}{14.4} & \IXDimScore{25.0}{0.0} & \IXDimScore{25.0}{0.0} & \IXDimScore{25.0}{0.0} & \IXDimScore{0.0}{0.0} & \IXDimScore{0.0}{0.0} & \IXDimScore{0.0}{0.0} & \IXDimScore{0.0}{0.0} & \IXDimScore{50.0}{25.0} & \IXDimScore{50.0}{25.0} & \IXDimScore{58.3}{14.4} & \IXDimScore{41.7}{14.4} \\
        Oracle
        & \IXDimScore{100.0}{0.0} & \IXDimScore{91.7}{14.4} & \IXDimScore{100.0}{0.0} & \IXDimScore{100.0}{0.0} & \IXDimScore{91.7}{14.4} & \IXDimScore{100.0}{0.0} & \IXDimScore{100.0}{0.0} & \IXDimScore{83.3}{14.4} & \IXDimScore{91.7}{14.4} & \IXDimScore{100.0}{0.0} & \IXDimScore{100.0}{0.0} & \IXDimScore{83.3}{14.4} & \IXDimScore{100.0}{0.0} & \IXDimScore{100.0}{0.0} & \IXDimScore{100.0}{0.0} & \IXDimScore{100.0}{0.0} \\
        \hline\hline
    \end{tabular}%
    }
    \par\smallskip
    \endgroup
    \parbox{\textwidth}{\scriptsize
        Each cell reports dimension-specific mean $\pm$ sample SD across three repeated probe evaluations under the same configuration (PSA, \%).
    }
\end{table*}

BRS and PRS for each user, backbone, and memory condition are given in Tables~\ref{tab:brs-results} and~\ref{tab:prs-results}, and their values by dimension are given in Tables~\ref{tab:brs-dim} and~\ref{tab:prs-dim}. BRS averaged over the users is shown for each backbone and memory condition in Figure~\ref{fig:brs-model}.

\begin{table*}[!htbp]
    \centering
    \caption{Behavioral Realization Score (BRS, 1--5) by user, backbone, and memory condition.}
    \label{tab:brs-results}
    \begingroup
    \setlength{\tabcolsep}{2pt}
    \setlength{\arrayrulewidth}{0.4pt}
    \setlength{\doublerulesep}{1.5pt}
    \renewcommand{\arraystretch}{1.45}
    \fontsize{7}{8.5}\selectfont
    \def\BRSTableGPT{\shortstack{GPT-5.6-\\Sol}}
    \def\BRSTableQwen{\shortstack{Qwen3.8-\\27B}}
    \def\BRSTableHY{\shortstack{HY4-\\Preview}}
    \def\BRSTableDeepSeek{\shortstack{DeepSeek-\\V4-Flash}}
    \resizebox{\textwidth}{!}{%
    \begin{tabular}{@{}l|cccc|cccc|cccc|cccc@{}}
        \hline\hline
        \multirow{2}{*}[-5.2pt]{\textbf{Memory}} &
        \multicolumn{4}{c|}{\textbf{User A}} &
        \multicolumn{4}{c|}{\textbf{User B}} &
        \multicolumn{4}{c|}{\textbf{User C}} &
        \multicolumn{4}{c}{\textbf{User D}} \\
        \cline{2-5}\cline{6-9}\cline{10-13}\cline{14-17}
        \rule{0pt}{19pt} & \BRSTableGPT & \BRSTableQwen & \BRSTableHY & \BRSTableDeepSeek
        & \BRSTableGPT & \BRSTableQwen & \BRSTableHY & \BRSTableDeepSeek
        & \BRSTableGPT & \BRSTableQwen & \BRSTableHY & \BRSTableDeepSeek
        & \BRSTableGPT & \BRSTableQwen & \BRSTableHY & \BRSTableDeepSeek \\
        \hline
        File memory
        & 3.74 & 4.01 & 3.37 & 4.04 & 3.80 & 3.98 & 3.86 & 3.79 & 2.82 & 2.79 & 3.12 & 2.70 & 3.51 & 3.52 & 3.67 & 3.33 \\
        Simple RAG
        & 2.87 & 2.88 & 2.95 & 3.18 & 3.56 & 3.40 & 3.20 & 3.25 & 2.52 & 2.85 & 3.04 & 2.88 & 3.17 & 3.52 & 3.57 & 3.48 \\
        Mem0
        & 2.94 & 3.18 & 2.85 & 3.24 & 3.55 & 3.26 & 3.26 & 3.64 & 2.63 & 2.92 & 2.90 & 2.63 & 3.36 & 3.69 & 3.50 & 3.43 \\
        MemOS
        & 3.05 & 3.14 & 2.94 & 3.14 & 3.54 & 3.12 & 3.23 & 3.52 & 2.67 & 2.70 & 2.89 & 2.73 & 3.05 & 3.73 & 3.52 & 3.50 \\
        \hline
        No Memory
        & 2.87 & 2.96 & 3.04 & 3.21 & 3.43 & 3.62 & 3.13 & 3.49 & 2.64 & 2.60 & 2.85 & 2.93 & 3.15 & 3.83 & 3.02 & 3.52 \\
        Oracle
        & 4.74 & 4.85 & 4.90 & 4.92 & 4.75 & 4.68 & 4.65 & 4.35 & 4.67 & 4.27 & 4.57 & 4.23 & 4.56 & 4.55 & 4.43 & 4.25 \\
        \hline\hline
    \end{tabular}%
    }
    \par\smallskip
    \endgroup
    \parbox{\textwidth}{\scriptsize
        Each cell reports mean BRS (1--5) over three probe runs.
    }
\end{table*}

\begin{table*}[!htbp]
    \centering
    \caption{Perceived Realization Score (PRS, 1--5) by user, backbone, and memory condition.}
    \label{tab:prs-results}
    \begingroup
    \setlength{\tabcolsep}{2pt}
    \setlength{\arrayrulewidth}{0.4pt}
    \setlength{\doublerulesep}{1.5pt}
    \renewcommand{\arraystretch}{1.45}
    \fontsize{7}{8.5}\selectfont
    \def\PRSTableGPT{\shortstack{GPT-5.6-\\Sol}}
    \def\PRSTableQwen{\shortstack{Qwen3.8-\\27B}}
    \def\PRSTableHY{\shortstack{HY4-\\Preview}}
    \def\PRSTableDeepSeek{\shortstack{DeepSeek-\\V4-Flash}}
    \resizebox{\textwidth}{!}{%
    \begin{tabular}{@{}l|cccc|cccc|cccc|cccc@{}}
        \hline\hline
        \multirow{2}{*}[-5.2pt]{\textbf{Memory}} &
        \multicolumn{4}{c|}{\textbf{User A}} &
        \multicolumn{4}{c|}{\textbf{User B}} &
        \multicolumn{4}{c|}{\textbf{User C}} &
        \multicolumn{4}{c}{\textbf{User D}} \\
        \cline{2-5}\cline{6-9}\cline{10-13}\cline{14-17}
        \rule{0pt}{19pt} & \PRSTableGPT & \PRSTableQwen & \PRSTableHY & \PRSTableDeepSeek
        & \PRSTableGPT & \PRSTableQwen & \PRSTableHY & \PRSTableDeepSeek
        & \PRSTableGPT & \PRSTableQwen & \PRSTableHY & \PRSTableDeepSeek
        & \PRSTableGPT & \PRSTableQwen & \PRSTableHY & \PRSTableDeepSeek \\
        \hline
        File memory
        & 3.81 & 4.04 & 3.57 & 4.08 & 3.83 & 3.82 & 3.83 & 3.85 & 3.12 & 2.81 & 3.26 & 2.86 & 3.60 & 3.75 & 3.81 & 3.65 \\
        Simple RAG
        & 3.18 & 3.01 & 3.13 & 3.33 & 3.69 & 3.26 & 3.27 & 3.30 & 2.79 & 2.99 & 2.93 & 2.75 & 3.35 & 3.65 & 3.71 & 3.55 \\
        Mem0
        & 3.13 & 3.26 & 3.01 & 3.35 & 3.54 & 3.36 & 3.36 & 3.61 & 2.96 & 3.00 & 3.01 & 2.73 & 3.44 & 3.73 & 3.87 & 3.15 \\
        MemOS
        & 3.25 & 3.25 & 2.95 & 3.24 & 3.56 & 3.18 & 3.43 & 3.56 & 2.69 & 2.75 & 3.07 & 2.86 & 3.33 & 3.62 & 3.88 & 3.48 \\
        \hline
        No Memory
        & 2.96 & 3.04 & 3.23 & 3.29 & 3.52 & 3.65 & 3.29 & 3.42 & 2.99 & 2.94 & 2.90 & 2.94 & 3.30 & 3.94 & 3.55 & 3.52 \\
        Oracle
        & 4.42 & 4.81 & 4.77 & 4.77 & 4.48 & 4.63 & 4.61 & 4.26 & 4.37 & 4.01 & 4.39 & 3.87 & 4.37 & 4.33 & 4.48 & 4.14 \\
        \hline\hline
    \end{tabular}%
    }
    \par\smallskip
    \endgroup
    \parbox{\textwidth}{\scriptsize
        Each cell reports mean PRS (1--5) over three probe runs.
    }
\end{table*}

\begin{table*}[tbp]
    \centering
    \caption{Behavioral Realization Score (BRS, 1--5) by IX dimension, user, backbone, and memory condition.}
    \label{tab:brs-dim}
    \begingroup
    \setlength{\tabcolsep}{2pt}
    \setlength{\arrayrulewidth}{0.4pt}
    \setlength{\doublerulesep}{1.5pt}
    \renewcommand{\arraystretch}{1.45}
    \fontsize{7}{8.5}\selectfont
    \def\BRSDimScore#1#2{#1\,{\fontsize{5.5}{6.5}\selectfont\textpm\,#2}}
    \def\BRSDimGPT{\shortstack{GPT-5.6-\\Sol}}
    \def\BRSDimQwen{\shortstack{Qwen3.8-\\27B}}
    \def\BRSDimHY{\shortstack{HY4-\\Preview}}
    \def\BRSDimDeepSeek{\shortstack{DeepSeek-\\V4-Flash}}
    \resizebox{\textwidth}{!}{%
    \begin{tabular}{@{}l cccc|cccc|cccc|cccc@{}}
        \hline\hline
        \textbf{Memory} &
        \multicolumn{4}{c|}{\textbf{User A}} &
        \multicolumn{4}{c|}{\textbf{User B}} &
        \multicolumn{4}{c|}{\textbf{User C}} &
        \multicolumn{4}{c}{\textbf{User D}} \\
        \cline{2-5}\cline{6-9}\cline{10-13}\cline{14-17}
        \rule{0pt}{19pt} & \BRSDimGPT & \BRSDimQwen & \BRSDimHY & \BRSDimDeepSeek
        & \BRSDimGPT & \BRSDimQwen & \BRSDimHY & \BRSDimDeepSeek
        & \BRSDimGPT & \BRSDimQwen & \BRSDimHY & \BRSDimDeepSeek
        & \BRSDimGPT & \BRSDimQwen & \BRSDimHY & \BRSDimDeepSeek \\
        \hline
        \multicolumn{17}{@{}c@{}}{\rule{0pt}{12pt}\textbf{Expression Style}} \\
        File memory
        & \BRSDimScore{4.88}{0.22} & \BRSDimScore{4.54}{0.07} & \BRSDimScore{4.75}{0.22} & \BRSDimScore{4.58}{0.07} & \BRSDimScore{3.58}{0.52} & \BRSDimScore{4.12}{0.33} & \BRSDimScore{4.71}{0.26} & \BRSDimScore{3.58}{0.36} & \BRSDimScore{3.38}{0.43} & \BRSDimScore{3.96}{0.31} & \BRSDimScore{3.79}{0.14} & \BRSDimScore{3.62}{0.54} & \BRSDimScore{3.33}{0.38} & \BRSDimScore{3.58}{0.31} & \BRSDimScore{3.25}{0.00} & \BRSDimScore{2.96}{0.19} \\
        Simple RAG
        & \BRSDimScore{3.46}{0.63} & \BRSDimScore{4.00}{0.50} & \BRSDimScore{3.88}{0.50} & \BRSDimScore{4.25}{0.33} & \BRSDimScore{3.42}{0.14} & \BRSDimScore{3.08}{0.19} & \BRSDimScore{3.21}{0.36} & \BRSDimScore{3.25}{0.45} & \BRSDimScore{2.92}{0.07} & \BRSDimScore{4.08}{0.40} & \BRSDimScore{4.00}{0.12} & \BRSDimScore{4.12}{0.12} & \BRSDimScore{2.92}{0.29} & \BRSDimScore{3.79}{0.19} & \BRSDimScore{3.25}{0.33} & \BRSDimScore{3.50}{0.22} \\
        Mem0
        & \BRSDimScore{3.83}{0.26} & \BRSDimScore{4.46}{0.31} & \BRSDimScore{3.42}{0.47} & \BRSDimScore{4.04}{0.14} & \BRSDimScore{3.62}{0.25} & \BRSDimScore{2.75}{0.54} & \BRSDimScore{3.21}{0.36} & \BRSDimScore{3.75}{0.25} & \BRSDimScore{2.96}{0.19} & \BRSDimScore{4.04}{0.31} & \BRSDimScore{3.96}{0.14} & \BRSDimScore{3.79}{0.38} & \BRSDimScore{2.83}{0.07} & \BRSDimScore{4.17}{0.56} & \BRSDimScore{3.21}{0.31} & \BRSDimScore{3.71}{0.26} \\
        MemOS
        & \BRSDimScore{4.33}{0.31} & \BRSDimScore{4.12}{0.12} & \BRSDimScore{4.21}{0.14} & \BRSDimScore{3.92}{0.40} & \BRSDimScore{3.42}{0.36} & \BRSDimScore{2.75}{0.43} & \BRSDimScore{2.96}{0.47} & \BRSDimScore{3.50}{0.12} & \BRSDimScore{3.38}{0.12} & \BRSDimScore{3.79}{0.44} & \BRSDimScore{3.42}{0.19} & \BRSDimScore{3.96}{0.14} & \BRSDimScore{2.75}{0.00} & \BRSDimScore{4.04}{0.44} & \BRSDimScore{2.96}{0.26} & \BRSDimScore{3.54}{0.52} \\
        \hline
        No Memory
        & \BRSDimScore{3.79}{0.26} & \BRSDimScore{4.29}{0.14} & \BRSDimScore{4.25}{0.22} & \BRSDimScore{4.17}{0.31} & \BRSDimScore{3.29}{0.26} & \BRSDimScore{3.62}{0.33} & \BRSDimScore{3.12}{0.22} & \BRSDimScore{3.29}{0.52} & \BRSDimScore{3.17}{0.38} & \BRSDimScore{3.62}{0.25} & \BRSDimScore{3.83}{0.26} & \BRSDimScore{4.04}{0.44} & \BRSDimScore{2.75}{0.33} & \BRSDimScore{4.08}{0.07} & \BRSDimScore{2.92}{0.44} & \BRSDimScore{3.75}{0.33} \\
        Oracle
        & \BRSDimScore{5.00}{0.00} & \BRSDimScore{5.00}{0.00} & \BRSDimScore{4.92}{0.14} & \BRSDimScore{5.00}{0.00} & \BRSDimScore{5.00}{0.00} & \BRSDimScore{4.83}{0.19} & \BRSDimScore{4.54}{0.07} & \BRSDimScore{4.50}{0.12} & \BRSDimScore{4.92}{0.14} & \BRSDimScore{4.54}{0.26} & \BRSDimScore{4.58}{0.38} & \BRSDimScore{4.38}{0.43} & \BRSDimScore{4.33}{0.26} & \BRSDimScore{4.42}{0.14} & \BRSDimScore{4.04}{0.07} & \BRSDimScore{3.96}{0.14} \\
        \hline
        \multicolumn{17}{@{}c@{}}{\rule{0pt}{12pt}\textbf{Disclosure}} \\
        File memory
        & \BRSDimScore{3.06}{0.35} & \BRSDimScore{4.78}{0.38} & \BRSDimScore{3.28}{0.67} & \BRSDimScore{4.00}{0.00} & \BRSDimScore{2.94}{0.96} & \BRSDimScore{3.78}{0.19} & \BRSDimScore{2.78}{0.69} & \BRSDimScore{4.22}{0.10} & \BRSDimScore{2.33}{0.00} & \BRSDimScore{2.06}{0.10} & \BRSDimScore{2.11}{0.10} & \BRSDimScore{2.67}{0.44} & \BRSDimScore{4.11}{0.10} & \BRSDimScore{3.67}{0.17} & \BRSDimScore{4.06}{0.35} & \BRSDimScore{4.11}{0.19} \\
        Simple RAG
        & \BRSDimScore{2.72}{0.10} & \BRSDimScore{2.67}{0.00} & \BRSDimScore{2.83}{0.29} & \BRSDimScore{2.89}{0.38} & \BRSDimScore{4.33}{0.00} & \BRSDimScore{4.44}{0.19} & \BRSDimScore{4.06}{0.35} & \BRSDimScore{3.78}{0.35} & \BRSDimScore{2.17}{0.29} & \BRSDimScore{2.06}{0.10} & \BRSDimScore{2.06}{0.10} & \BRSDimScore{2.39}{0.42} & \BRSDimScore{4.11}{0.25} & \BRSDimScore{3.94}{0.25} & \BRSDimScore{3.94}{0.38} & \BRSDimScore{4.28}{0.10} \\
        Mem0
        & \BRSDimScore{3.11}{0.38} & \BRSDimScore{2.78}{0.19} & \BRSDimScore{2.61}{0.10} & \BRSDimScore{3.11}{0.38} & \BRSDimScore{4.11}{0.38} & \BRSDimScore{4.33}{0.33} & \BRSDimScore{4.17}{0.29} & \BRSDimScore{4.17}{0.29} & \BRSDimScore{2.50}{0.29} & \BRSDimScore{2.06}{0.10} & \BRSDimScore{2.06}{0.19} & \BRSDimScore{2.00}{0.29} & \BRSDimScore{4.06}{0.19} & \BRSDimScore{4.06}{0.19} & \BRSDimScore{4.17}{0.17} & \BRSDimScore{4.33}{0.00} \\
        MemOS
        & \BRSDimScore{2.89}{0.38} & \BRSDimScore{2.78}{0.19} & \BRSDimScore{2.83}{0.44} & \BRSDimScore{2.89}{0.38} & \BRSDimScore{3.72}{0.59} & \BRSDimScore{4.00}{0.29} & \BRSDimScore{3.89}{0.48} & \BRSDimScore{4.28}{0.10} & \BRSDimScore{2.50}{0.29} & \BRSDimScore{2.00}{0.00} & \BRSDimScore{2.22}{0.42} & \BRSDimScore{1.94}{0.19} & \BRSDimScore{4.28}{0.10} & \BRSDimScore{4.11}{0.10} & \BRSDimScore{4.17}{0.17} & \BRSDimScore{3.78}{0.67} \\
        \hline
        No Memory
        & \BRSDimScore{2.72}{0.10} & \BRSDimScore{2.61}{0.25} & \BRSDimScore{2.67}{0.00} & \BRSDimScore{3.06}{0.35} & \BRSDimScore{4.06}{0.48} & \BRSDimScore{4.22}{0.10} & \BRSDimScore{4.17}{0.17} & \BRSDimScore{4.33}{0.00} & \BRSDimScore{2.56}{0.25} & \BRSDimScore{2.11}{0.10} & \BRSDimScore{2.06}{0.10} & \BRSDimScore{2.22}{0.35} & \BRSDimScore{4.11}{0.10} & \BRSDimScore{4.00}{0.17} & \BRSDimScore{3.72}{0.77} & \BRSDimScore{3.89}{0.35} \\
        Oracle
        & \BRSDimScore{5.00}{0.00} & \BRSDimScore{5.00}{0.00} & \BRSDimScore{5.00}{0.00} & \BRSDimScore{5.00}{0.00} & \BRSDimScore{4.89}{0.10} & \BRSDimScore{3.94}{0.19} & \BRSDimScore{4.56}{0.35} & \BRSDimScore{4.61}{0.35} & \BRSDimScore{5.00}{0.00} & \BRSDimScore{4.06}{0.35} & \BRSDimScore{4.06}{0.25} & \BRSDimScore{3.94}{0.35} & \BRSDimScore{4.78}{0.25} & \BRSDimScore{4.22}{0.10} & \BRSDimScore{4.50}{0.00} & \BRSDimScore{4.22}{0.25} \\
        \hline
        \multicolumn{17}{@{}c@{}}{\rule{0pt}{12pt}\textbf{Initiative \& Autonomy}} \\
        File memory
        & \BRSDimScore{3.27}{0.06} & \BRSDimScore{3.40}{0.10} & \BRSDimScore{2.00}{0.60} & \BRSDimScore{3.70}{0.17} & \BRSDimScore{4.77}{0.32} & \BRSDimScore{4.43}{0.29} & \BRSDimScore{4.23}{0.67} & \BRSDimScore{4.23}{0.31} & \BRSDimScore{3.27}{0.12} & \BRSDimScore{2.87}{0.42} & \BRSDimScore{3.33}{0.59} & \BRSDimScore{2.53}{0.15} & \BRSDimScore{3.43}{0.70} & \BRSDimScore{3.30}{0.30} & \BRSDimScore{4.03}{0.23} & \BRSDimScore{3.10}{0.26} \\
        Simple RAG
        & \BRSDimScore{2.20}{0.44} & \BRSDimScore{1.97}{0.35} & \BRSDimScore{2.00}{0.10} & \BRSDimScore{2.53}{0.21} & \BRSDimScore{3.63}{0.67} & \BRSDimScore{3.53}{0.40} & \BRSDimScore{2.63}{0.45} & \BRSDimScore{3.40}{0.17} & \BRSDimScore{2.97}{0.06} & \BRSDimScore{3.03}{0.29} & \BRSDimScore{3.60}{0.10} & \BRSDimScore{2.77}{0.21} & \BRSDimScore{2.90}{0.46} & \BRSDimScore{3.17}{0.21} & \BRSDimScore{3.77}{0.23} & \BRSDimScore{2.90}{0.70} \\
        Mem0
        & \BRSDimScore{1.90}{0.10} & \BRSDimScore{2.30}{0.26} & \BRSDimScore{2.20}{0.10} & \BRSDimScore{2.60}{0.61} & \BRSDimScore{3.60}{0.36} & \BRSDimScore{3.30}{0.26} & \BRSDimScore{2.80}{0.44} & \BRSDimScore{3.80}{0.00} & \BRSDimScore{2.97}{0.32} & \BRSDimScore{3.23}{0.06} & \BRSDimScore{3.17}{0.32} & \BRSDimScore{2.67}{0.23} & \BRSDimScore{3.50}{0.00} & \BRSDimScore{3.40}{0.53} & \BRSDimScore{3.30}{0.20} & \BRSDimScore{3.10}{0.56} \\
        MemOS
        & \BRSDimScore{1.83}{0.15} & \BRSDimScore{2.37}{0.06} & \BRSDimScore{1.70}{0.17} & \BRSDimScore{2.57}{0.55} & \BRSDimScore{3.83}{0.25} & \BRSDimScore{3.17}{0.06} & \BRSDimScore{3.30}{0.26} & \BRSDimScore{3.50}{0.00} & \BRSDimScore{2.70}{0.26} & \BRSDimScore{2.80}{0.36} & \BRSDimScore{3.60}{0.46} & \BRSDimScore{2.90}{0.52} & \BRSDimScore{3.00}{0.53} & \BRSDimScore{3.50}{0.10} & \BRSDimScore{3.60}{0.56} & \BRSDimScore{3.47}{0.74} \\
        \hline
        No Memory
        & \BRSDimScore{2.27}{0.12} & \BRSDimScore{1.80}{0.35} & \BRSDimScore{2.13}{0.12} & \BRSDimScore{2.53}{0.25} & \BRSDimScore{3.47}{0.06} & \BRSDimScore{3.53}{0.35} & \BRSDimScore{2.90}{0.17} & \BRSDimScore{3.53}{0.47} & \BRSDimScore{2.90}{0.44} & \BRSDimScore{2.67}{0.45} & \BRSDimScore{3.17}{0.46} & \BRSDimScore{3.23}{0.35} & \BRSDimScore{3.17}{0.23} & \BRSDimScore{3.73}{0.12} & \BRSDimScore{2.43}{0.42} & \BRSDimScore{3.20}{0.20} \\
        Oracle
        & \BRSDimScore{4.93}{0.12} & \BRSDimScore{4.87}{0.15} & \BRSDimScore{4.80}{0.20} & \BRSDimScore{4.80}{0.17} & \BRSDimScore{4.73}{0.12} & \BRSDimScore{4.93}{0.06} & \BRSDimScore{4.70}{0.20} & \BRSDimScore{4.03}{0.32} & \BRSDimScore{4.63}{0.15} & \BRSDimScore{4.30}{0.30} & \BRSDimScore{4.83}{0.21} & \BRSDimScore{4.27}{0.15} & \BRSDimScore{4.63}{0.06} & \BRSDimScore{4.97}{0.06} & \BRSDimScore{4.67}{0.25} & \BRSDimScore{4.33}{0.06} \\
        \hline
        \multicolumn{17}{@{}c@{}}{\rule{0pt}{12pt}\textbf{Information Flow}} \\
        File memory
        & \BRSDimScore{3.67}{0.14} & \BRSDimScore{3.33}{0.14} & \BRSDimScore{4.17}{0.14} & \BRSDimScore{3.83}{1.04} & \BRSDimScore{3.08}{0.38} & \BRSDimScore{2.83}{0.38} & \BRSDimScore{2.83}{0.38} & \BRSDimScore{2.42}{0.38} & \BRSDimScore{1.33}{0.58} & \BRSDimScore{1.33}{0.58} & \BRSDimScore{2.75}{0.90} & \BRSDimScore{1.33}{0.58} & \BRSDimScore{3.17}{0.88} & \BRSDimScore{3.75}{0.43} & \BRSDimScore{3.00}{0.43} & \BRSDimScore{3.50}{0.50} \\
        Simple RAG
        & \BRSDimScore{3.58}{0.52} & \BRSDimScore{3.25}{0.25} & \BRSDimScore{3.67}{0.38} & \BRSDimScore{3.08}{0.14} & \BRSDimScore{2.50}{0.25} & \BRSDimScore{2.17}{0.14} & \BRSDimScore{3.33}{0.38} & \BRSDimScore{2.08}{0.14} & \BRSDimScore{1.17}{0.14} & \BRSDimScore{1.08}{0.14} & \BRSDimScore{1.17}{0.14} & \BRSDimScore{1.42}{0.52} & \BRSDimScore{2.92}{0.14} & \BRSDimScore{3.25}{0.90} & \BRSDimScore{3.17}{0.58} & \BRSDimScore{3.67}{0.29} \\
        Mem0
        & \BRSDimScore{3.50}{0.25} & \BRSDimScore{3.42}{0.29} & \BRSDimScore{3.67}{0.38} & \BRSDimScore{3.42}{0.52} & \BRSDimScore{2.42}{0.52} & \BRSDimScore{2.58}{0.38} & \BRSDimScore{3.17}{0.14} & \BRSDimScore{2.25}{0.43} & \BRSDimScore{1.33}{0.58} & \BRSDimScore{1.17}{0.29} & \BRSDimScore{1.42}{0.29} & \BRSDimScore{1.17}{0.29} & \BRSDimScore{3.00}{0.50} & \BRSDimScore{2.92}{1.28} & \BRSDimScore{3.58}{0.38} & \BRSDimScore{2.33}{0.29} \\
        MemOS
        & \BRSDimScore{3.75}{0.25} & \BRSDimScore{3.67}{0.29} & \BRSDimScore{3.67}{0.38} & \BRSDimScore{3.42}{0.38} & \BRSDimScore{2.75}{0.00} & \BRSDimScore{2.42}{0.14} & \BRSDimScore{2.58}{0.63} & \BRSDimScore{2.50}{0.25} & \BRSDimScore{1.42}{0.38} & \BRSDimScore{1.33}{0.38} & \BRSDimScore{1.08}{0.14} & \BRSDimScore{1.00}{0.00} & \BRSDimScore{1.92}{0.76} & \BRSDimScore{3.08}{0.72} & \BRSDimScore{3.50}{0.25} & \BRSDimScore{3.08}{0.38} \\
        \hline
        No Memory
        & \BRSDimScore{2.75}{0.43} & \BRSDimScore{3.75}{0.25} & \BRSDimScore{3.42}{0.14} & \BRSDimScore{3.25}{0.66} & \BRSDimScore{2.67}{0.14} & \BRSDimScore{2.92}{0.63} & \BRSDimScore{2.17}{0.14} & \BRSDimScore{2.50}{0.66} & \BRSDimScore{1.08}{0.14} & \BRSDimScore{1.08}{0.14} & \BRSDimScore{1.25}{0.25} & \BRSDimScore{1.00}{0.00} & \BRSDimScore{2.50}{0.25} & \BRSDimScore{3.33}{0.14} & \BRSDimScore{3.67}{0.29} & \BRSDimScore{3.33}{0.14} \\
        Oracle
        & \BRSDimScore{3.33}{0.52} & \BRSDimScore{4.25}{0.66} & \BRSDimScore{5.00}{0.00} & \BRSDimScore{4.92}{0.14} & \BRSDimScore{4.08}{0.52} & \BRSDimScore{4.83}{0.29} & \BRSDimScore{4.92}{0.14} & \BRSDimScore{4.42}{0.52} & \BRSDimScore{3.75}{0.90} & \BRSDimScore{4.00}{0.50} & \BRSDimScore{4.67}{0.38} & \BRSDimScore{4.25}{0.75} & \BRSDimScore{4.50}{0.00} & \BRSDimScore{4.25}{0.25} & \BRSDimScore{4.50}{0.43} & \BRSDimScore{4.67}{0.29} \\
        \hline\hline
    \end{tabular}%
    }
    \par\smallskip
    \endgroup
    \parbox{\textwidth}{\scriptsize
        Each cell reports dimension-specific mean $\pm$ sample SD across three repeated probe evaluations under the same configuration (BRS, 1--5).
    }
\end{table*}

\begin{table*}[tbp]
    \centering
    \caption{Perceived Realization Score (PRS, 1--5) by IX dimension, user, backbone, and memory condition.}
    \label{tab:prs-dim}
    \begingroup
    \setlength{\tabcolsep}{2pt}
    \setlength{\arrayrulewidth}{0.4pt}
    \setlength{\doublerulesep}{1.5pt}
    \renewcommand{\arraystretch}{1.45}
    \fontsize{7}{8.5}\selectfont
    \def\PRSDimScore#1#2{#1\,{\fontsize{5.5}{6.5}\selectfont\textpm\,#2}}
    \def\PRSDimGPT{\shortstack{GPT-5.6-\\Sol}}
    \def\PRSDimQwen{\shortstack{Qwen3.8-\\27B}}
    \def\PRSDimHY{\shortstack{HY4-\\Preview}}
    \def\PRSDimDeepSeek{\shortstack{DeepSeek-\\V4-Flash}}
    \resizebox{\textwidth}{!}{%
    \begin{tabular}{@{}l cccc|cccc|cccc|cccc@{}}
        \hline\hline
        \textbf{Memory} &
        \multicolumn{4}{c|}{\textbf{User A}} &
        \multicolumn{4}{c|}{\textbf{User B}} &
        \multicolumn{4}{c|}{\textbf{User C}} &
        \multicolumn{4}{c}{\textbf{User D}} \\
        \cline{2-5}\cline{6-9}\cline{10-13}\cline{14-17}
        \rule{0pt}{19pt} & \PRSDimGPT & \PRSDimQwen & \PRSDimHY & \PRSDimDeepSeek
        & \PRSDimGPT & \PRSDimQwen & \PRSDimHY & \PRSDimDeepSeek
        & \PRSDimGPT & \PRSDimQwen & \PRSDimHY & \PRSDimDeepSeek
        & \PRSDimGPT & \PRSDimQwen & \PRSDimHY & \PRSDimDeepSeek \\
        \hline
        \multicolumn{17}{@{}c@{}}{\rule{0pt}{12pt}\textbf{Expression Style}} \\
        File memory
        & \PRSDimScore{4.88}{0.22} & \PRSDimScore{4.33}{0.19} & \PRSDimScore{4.75}{0.22} & \PRSDimScore{4.58}{0.07} & \PRSDimScore{4.08}{0.52} & \PRSDimScore{3.75}{0.25} & \PRSDimScore{4.71}{0.29} & \PRSDimScore{3.62}{0.45} & \PRSDimScore{3.54}{0.29} & \PRSDimScore{3.75}{0.33} & \PRSDimScore{4.00}{0.33} & \PRSDimScore{3.50}{0.22} & \PRSDimScore{4.04}{0.19} & \PRSDimScore{4.25}{0.00} & \PRSDimScore{4.17}{0.07} & \PRSDimScore{3.50}{0.22} \\
        Simple RAG
        & \PRSDimScore{3.54}{0.52} & \PRSDimScore{4.08}{0.51} & \PRSDimScore{3.96}{0.26} & \PRSDimScore{3.62}{0.45} & \PRSDimScore{3.92}{0.14} & \PRSDimScore{3.04}{0.29} & \PRSDimScore{3.25}{0.25} & \PRSDimScore{3.04}{0.26} & \PRSDimScore{3.29}{0.07} & \PRSDimScore{3.92}{0.19} & \PRSDimScore{3.88}{0.33} & \PRSDimScore{3.46}{0.40} & \PRSDimScore{3.29}{0.14} & \PRSDimScore{4.12}{0.25} & \PRSDimScore{3.83}{0.07} & \PRSDimScore{3.92}{0.26} \\
        Mem0
        & \PRSDimScore{3.75}{0.12} & \PRSDimScore{4.33}{0.26} & \PRSDimScore{3.42}{0.29} & \PRSDimScore{3.46}{0.19} & \PRSDimScore{3.62}{0.12} & \PRSDimScore{2.88}{0.33} & \PRSDimScore{3.42}{0.29} & \PRSDimScore{3.83}{0.14} & \PRSDimScore{3.50}{0.22} & \PRSDimScore{3.83}{0.38} & \PRSDimScore{3.92}{0.19} & \PRSDimScore{3.46}{0.19} & \PRSDimScore{3.38}{0.12} & \PRSDimScore{4.21}{0.31} & \PRSDimScore{3.88}{0.12} & \PRSDimScore{3.75}{0.38} \\
        MemOS
        & \PRSDimScore{4.04}{0.14} & \PRSDimScore{4.04}{0.26} & \PRSDimScore{4.17}{0.14} & \PRSDimScore{3.62}{0.25} & \PRSDimScore{3.62}{0.33} & \PRSDimScore{2.79}{0.31} & \PRSDimScore{3.33}{0.26} & \PRSDimScore{3.67}{0.07} & \PRSDimScore{3.38}{0.12} & \PRSDimScore{3.71}{0.31} & \PRSDimScore{3.88}{0.22} & \PRSDimScore{3.71}{0.19} & \PRSDimScore{3.46}{0.19} & \PRSDimScore{4.21}{0.14} & \PRSDimScore{3.88}{0.12} & \PRSDimScore{3.88}{0.38} \\
        \hline
        No Memory
        & \PRSDimScore{3.83}{0.07} & \PRSDimScore{4.12}{0.12} & \PRSDimScore{4.21}{0.07} & \PRSDimScore{3.92}{0.19} & \PRSDimScore{3.67}{0.26} & \PRSDimScore{3.67}{0.44} & \PRSDimScore{3.42}{0.07} & \PRSDimScore{3.08}{0.62} & \PRSDimScore{3.67}{0.26} & \PRSDimScore{3.75}{0.22} & \PRSDimScore{4.08}{0.31} & \PRSDimScore{3.67}{0.14} & \PRSDimScore{3.21}{0.07} & \PRSDimScore{4.17}{0.14} & \PRSDimScore{3.75}{0.25} & \PRSDimScore{3.71}{0.19} \\
        Oracle
        & \PRSDimScore{4.96}{0.07} & \PRSDimScore{5.00}{0.00} & \PRSDimScore{4.67}{0.29} & \PRSDimScore{4.92}{0.07} & \PRSDimScore{4.83}{0.19} & \PRSDimScore{4.75}{0.12} & \PRSDimScore{4.62}{0.12} & \PRSDimScore{4.25}{0.33} & \PRSDimScore{4.50}{0.22} & \PRSDimScore{4.21}{0.07} & \PRSDimScore{4.33}{0.19} & \PRSDimScore{3.83}{0.26} & \PRSDimScore{4.67}{0.19} & \PRSDimScore{4.42}{0.29} & \PRSDimScore{4.29}{0.07} & \PRSDimScore{4.25}{0.54} \\
        \hline
        \multicolumn{17}{@{}c@{}}{\rule{0pt}{12pt}\textbf{Disclosure}} \\
        File memory
        & \PRSDimScore{3.44}{0.35} & \PRSDimScore{4.67}{0.33} & \PRSDimScore{3.39}{1.02} & \PRSDimScore{4.06}{0.10} & \PRSDimScore{2.72}{0.48} & \PRSDimScore{3.44}{0.10} & \PRSDimScore{2.33}{0.44} & \PRSDimScore{4.17}{0.00} & \PRSDimScore{2.56}{0.10} & \PRSDimScore{1.72}{0.25} & \PRSDimScore{1.94}{0.51} & \PRSDimScore{2.39}{0.67} & \PRSDimScore{3.94}{0.25} & \PRSDimScore{3.28}{0.42} & \PRSDimScore{3.50}{0.44} & \PRSDimScore{3.94}{0.42} \\
        Simple RAG
        & \PRSDimScore{3.17}{0.29} & \PRSDimScore{2.78}{0.19} & \PRSDimScore{3.00}{0.29} & \PRSDimScore{3.06}{0.59} & \PRSDimScore{4.11}{0.25} & \PRSDimScore{3.89}{0.25} & \PRSDimScore{4.06}{0.42} & \PRSDimScore{3.61}{0.19} & \PRSDimScore{2.39}{0.48} & \PRSDimScore{2.33}{0.17} & \PRSDimScore{1.89}{0.10} & \PRSDimScore{1.89}{0.38} & \PRSDimScore{4.06}{0.19} & \PRSDimScore{3.83}{0.29} & \PRSDimScore{3.78}{0.51} & \PRSDimScore{4.11}{0.54} \\
        Mem0
        & \PRSDimScore{3.33}{0.50} & \PRSDimScore{3.17}{0.00} & \PRSDimScore{3.33}{0.29} & \PRSDimScore{3.44}{0.35} & \PRSDimScore{4.11}{0.25} & \PRSDimScore{4.22}{0.19} & \PRSDimScore{3.83}{0.29} & \PRSDimScore{3.61}{0.59} & \PRSDimScore{2.56}{0.10} & \PRSDimScore{2.22}{0.19} & \PRSDimScore{2.06}{0.42} & \PRSDimScore{2.11}{0.48} & \PRSDimScore{3.94}{0.25} & \PRSDimScore{3.83}{0.29} & \PRSDimScore{3.89}{0.25} & \PRSDimScore{3.39}{0.63} \\
        MemOS
        & \PRSDimScore{3.33}{0.17} & \PRSDimScore{2.94}{0.38} & \PRSDimScore{2.83}{0.17} & \PRSDimScore{3.17}{0.17} & \PRSDimScore{3.83}{0.50} & \PRSDimScore{3.94}{0.42} & \PRSDimScore{3.78}{0.35} & \PRSDimScore{3.72}{0.25} & \PRSDimScore{2.33}{0.17} & \PRSDimScore{1.56}{0.25} & \PRSDimScore{2.17}{0.44} & \PRSDimScore{2.17}{0.29} & \PRSDimScore{3.94}{0.42} & \PRSDimScore{3.56}{0.35} & \PRSDimScore{4.17}{0.29} & \PRSDimScore{3.56}{0.48} \\
        \hline
        No Memory
        & \PRSDimScore{3.11}{0.63} & \PRSDimScore{2.89}{0.35} & \PRSDimScore{2.89}{0.25} & \PRSDimScore{3.17}{0.60} & \PRSDimScore{3.72}{0.69} & \PRSDimScore{4.06}{0.10} & \PRSDimScore{3.72}{0.38} & \PRSDimScore{3.83}{0.44} & \PRSDimScore{2.67}{0.17} & \PRSDimScore{2.06}{0.10} & \PRSDimScore{1.83}{0.29} & \PRSDimScore{2.11}{0.59} & \PRSDimScore{4.00}{0.17} & \PRSDimScore{3.72}{0.10} & \PRSDimScore{3.67}{0.60} & \PRSDimScore{3.67}{0.29} \\
        Oracle
        & \PRSDimScore{5.00}{0.00} & \PRSDimScore{5.00}{0.00} & \PRSDimScore{5.00}{0.00} & \PRSDimScore{5.00}{0.00} & \PRSDimScore{4.33}{0.44} & \PRSDimScore{3.89}{0.25} & \PRSDimScore{4.17}{0.17} & \PRSDimScore{4.00}{0.44} & \PRSDimScore{4.61}{0.38} & \PRSDimScore{3.78}{0.10} & \PRSDimScore{3.89}{0.25} & \PRSDimScore{3.50}{0.44} & \PRSDimScore{4.39}{0.38} & \PRSDimScore{3.83}{0.33} & \PRSDimScore{4.11}{0.10} & \PRSDimScore{3.78}{0.19} \\
        \hline
        \multicolumn{17}{@{}c@{}}{\rule{0pt}{12pt}\textbf{Initiative \& Autonomy}} \\
        File memory
        & \PRSDimScore{3.30}{0.10} & \PRSDimScore{3.70}{0.10} & \PRSDimScore{2.57}{0.50} & \PRSDimScore{3.93}{0.23} & \PRSDimScore{4.57}{0.51} & \PRSDimScore{4.57}{0.23} & \PRSDimScore{4.27}{0.50} & \PRSDimScore{4.33}{0.15} & \PRSDimScore{3.57}{0.15} & \PRSDimScore{2.97}{0.15} & \PRSDimScore{3.67}{0.31} & \PRSDimScore{2.97}{0.23} & \PRSDimScore{3.33}{0.21} & \PRSDimScore{3.57}{0.15} & \PRSDimScore{4.03}{0.35} & \PRSDimScore{3.63}{0.31} \\
        Simple RAG
        & \PRSDimScore{2.70}{0.26} & \PRSDimScore{2.17}{0.25} & \PRSDimScore{2.40}{0.36} & \PRSDimScore{3.30}{0.26} & \PRSDimScore{3.60}{0.56} & \PRSDimScore{3.27}{0.21} & \PRSDimScore{2.67}{0.55} & \PRSDimScore{3.60}{0.17} & \PRSDimScore{3.00}{0.00} & \PRSDimScore{3.17}{0.15} & \PRSDimScore{3.40}{0.10} & \PRSDimScore{3.10}{0.10} & \PRSDimScore{3.03}{0.23} & \PRSDimScore{3.27}{0.21} & \PRSDimScore{3.90}{0.26} & \PRSDimScore{2.87}{0.32} \\
        Mem0
        & \PRSDimScore{2.40}{0.26} & \PRSDimScore{2.50}{0.00} & \PRSDimScore{2.30}{0.17} & \PRSDimScore{3.17}{0.31} & \PRSDimScore{3.50}{0.10} & \PRSDimScore{3.33}{0.21} & \PRSDimScore{2.97}{0.49} & \PRSDimScore{3.73}{0.12} & \PRSDimScore{3.13}{0.38} & \PRSDimScore{3.33}{0.31} & \PRSDimScore{3.40}{0.30} & \PRSDimScore{3.03}{0.25} & \PRSDimScore{3.27}{0.15} & \PRSDimScore{3.53}{0.29} & \PRSDimScore{3.93}{0.15} & \PRSDimScore{2.83}{0.32} \\
        MemOS
        & \PRSDimScore{2.47}{0.32} & \PRSDimScore{2.70}{0.20} & \PRSDimScore{1.93}{0.42} & \PRSDimScore{3.00}{0.46} & \PRSDimScore{3.57}{0.35} & \PRSDimScore{3.20}{0.00} & \PRSDimScore{3.60}{0.20} & \PRSDimScore{3.73}{0.21} & \PRSDimScore{2.80}{0.26} & \PRSDimScore{3.03}{0.31} & \PRSDimScore{3.57}{0.60} & \PRSDimScore{3.17}{0.12} & \PRSDimScore{3.37}{0.06} & \PRSDimScore{3.37}{0.29} & \PRSDimScore{3.80}{0.35} & \PRSDimScore{3.17}{0.40} \\
        \hline
        No Memory
        & \PRSDimScore{2.43}{0.25} & \PRSDimScore{2.13}{0.25} & \PRSDimScore{2.50}{0.00} & \PRSDimScore{2.93}{0.06} & \PRSDimScore{3.53}{0.15} & \PRSDimScore{3.57}{0.21} & \PRSDimScore{3.17}{0.29} & \PRSDimScore{3.67}{0.15} & \PRSDimScore{3.13}{0.25} & \PRSDimScore{3.37}{0.50} & \PRSDimScore{3.03}{0.35} & \PRSDimScore{3.37}{0.12} & \PRSDimScore{3.13}{0.35} & \PRSDimScore{3.93}{0.47} & \PRSDimScore{3.23}{0.15} & \PRSDimScore{3.23}{0.15} \\
        Oracle
        & \PRSDimScore{4.37}{0.21} & \PRSDimScore{4.70}{0.17} & \PRSDimScore{4.67}{0.25} & \PRSDimScore{4.60}{0.36} & \PRSDimScore{4.60}{0.10} & \PRSDimScore{4.97}{0.06} & \PRSDimScore{4.73}{0.15} & \PRSDimScore{4.43}{0.21} & \PRSDimScore{4.53}{0.21} & \PRSDimScore{4.10}{0.20} & \PRSDimScore{4.63}{0.12} & \PRSDimScore{4.17}{0.31} & \PRSDimScore{4.17}{0.21} & \PRSDimScore{4.60}{0.17} & \PRSDimScore{4.87}{0.23} & \PRSDimScore{4.13}{0.42} \\
        \hline
        \multicolumn{17}{@{}c@{}}{\rule{0pt}{12pt}\textbf{Information Flow}} \\
        File memory
        & \PRSDimScore{3.50}{0.00} & \PRSDimScore{3.33}{0.29} & \PRSDimScore{4.00}{0.25} & \PRSDimScore{3.50}{0.25} & \PRSDimScore{3.17}{0.52} & \PRSDimScore{2.67}{0.52} & \PRSDimScore{3.25}{0.00} & \PRSDimScore{2.58}{0.14} & \PRSDimScore{2.00}{0.66} & \PRSDimScore{2.17}{0.29} & \PRSDimScore{2.75}{0.25} & \PRSDimScore{2.00}{0.43} & \PRSDimScore{2.83}{0.38} & \PRSDimScore{3.92}{0.58} & \PRSDimScore{3.00}{0.43} & \PRSDimScore{3.58}{0.38} \\
        Simple RAG
        & \PRSDimScore{3.67}{0.29} & \PRSDimScore{3.33}{0.14} & \PRSDimScore{3.50}{0.25} & \PRSDimScore{3.25}{0.25} & \PRSDimScore{2.83}{0.38} & \PRSDimScore{2.75}{0.00} & \PRSDimScore{3.67}{0.29} & \PRSDimScore{2.58}{0.29} & \PRSDimScore{1.83}{0.14} & \PRSDimScore{1.67}{0.14} & \PRSDimScore{1.42}{0.14} & \PRSDimScore{1.75}{0.87} & \PRSDimScore{3.17}{0.38} & \PRSDimScore{3.42}{0.63} & \PRSDimScore{2.92}{0.52} & \PRSDimScore{3.67}{0.29} \\
        Mem0
        & \PRSDimScore{3.42}{0.14} & \PRSDimScore{3.17}{0.14} & \PRSDimScore{3.50}{0.43} & \PRSDimScore{3.42}{0.52} & \PRSDimScore{2.58}{0.14} & \PRSDimScore{3.08}{0.52} & \PRSDimScore{3.50}{0.00} & \PRSDimScore{2.83}{0.58} & \PRSDimScore{2.08}{0.58} & \PRSDimScore{1.67}{0.14} & \PRSDimScore{1.67}{0.29} & \PRSDimScore{1.42}{0.14} & \PRSDimScore{3.25}{0.43} & \PRSDimScore{3.08}{1.18} & \PRSDimScore{3.67}{0.52} & \PRSDimScore{2.42}{0.38} \\
        MemOS
        & \PRSDimScore{3.50}{0.25} & \PRSDimScore{3.50}{0.25} & \PRSDimScore{3.25}{0.00} & \PRSDimScore{3.17}{0.14} & \PRSDimScore{3.00}{0.25} & \PRSDimScore{2.75}{0.00} & \PRSDimScore{2.67}{0.14} & \PRSDimScore{2.67}{0.14} & \PRSDimScore{1.58}{0.14} & \PRSDimScore{1.92}{0.29} & \PRSDimScore{1.58}{0.14} & \PRSDimScore{1.42}{0.38} & \PRSDimScore{2.08}{0.58} & \PRSDimScore{3.17}{0.58} & \PRSDimScore{3.67}{0.29} & \PRSDimScore{3.33}{0.29} \\
        \hline
        No Memory
        & \PRSDimScore{2.33}{0.14} & \PRSDimScore{3.33}{0.14} & \PRSDimScore{3.58}{0.58} & \PRSDimScore{3.08}{0.88} & \PRSDimScore{2.92}{0.29} & \PRSDimScore{3.25}{0.50} & \PRSDimScore{2.67}{0.14} & \PRSDimScore{2.83}{0.58} & \PRSDimScore{1.75}{0.25} & \PRSDimScore{1.58}{0.14} & \PRSDimScore{1.83}{0.14} & \PRSDimScore{1.67}{0.14} & \PRSDimScore{2.83}{0.38} & \PRSDimScore{3.83}{0.29} & \PRSDimScore{3.75}{0.43} & \PRSDimScore{3.67}{0.29} \\
        Oracle
        & \PRSDimScore{2.58}{0.63} & \PRSDimScore{4.42}{0.29} & \PRSDimScore{4.92}{0.14} & \PRSDimScore{4.58}{0.14} & \PRSDimScore{3.67}{0.38} & \PRSDimScore{4.67}{0.38} & \PRSDimScore{4.92}{0.14} & \PRSDimScore{4.25}{0.50} & \PRSDimScore{3.33}{0.95} & \PRSDimScore{3.75}{0.66} & \PRSDimScore{4.67}{0.58} & \PRSDimScore{3.75}{0.66} & \PRSDimScore{4.25}{0.00} & \PRSDimScore{4.25}{0.25} & \PRSDimScore{4.42}{0.63} & \PRSDimScore{4.50}{0.43} \\
        \hline\hline
    \end{tabular}%
    }
    \par\smallskip
    \endgroup
    \parbox{\textwidth}{\scriptsize
        Each cell reports dimension-specific mean $\pm$ sample SD across three repeated probe evaluations under the same configuration (PRS, 1--5).
    }
\end{table*}

\subsection{PSA and General Capability}
\label{app:general-capability}

The PSA ranking of the four backbones does not follow their general capability. Table~\ref{tab:general-capability} compares PSA with two scores from Artificial Analysis~\citep{artificialanalysis2026}. The Intelligence Index combines ten evaluations, and GDPval-AA is an Elo rating on agentic real-world work tasks. PA calls do not set a reasoning effort, so each backbone runs at its default effort, and each score comes from the matching leaderboard entry. GPT-5.6-Sol has the highest Intelligence Index of the listed backbones and the lowest PSA. Qwen3.8-27B reaches the highest PSA, and its GDPval-AA rating is close to that of GPT-5.6-Sol.

\begin{table}[t]
    \centering
    \caption{PSA and general capability of the four PA backbones. PSA averages Users A--D and the four memory conditions, and the other scores are from \citet{artificialanalysis2026}. PSA does not follow general capability.}
    \label{tab:general-capability}
    \begingroup
    \small
    \setlength{\tabcolsep}{5pt}
    \renewcommand{\arraystretch}{1.16}
    \begin{tabular}{@{}lccc@{}}
        \toprule
        \textbf{Model (default reasoning effort)} & \textbf{Intelligence Index} & \textbf{GDPval-AA (Elo)} & \textbf{PSA (\%)} \\
        \midrule
        GPT-5.6-Sol (medium) & 39 & 1403 & 49.8 \\
        Qwen3.8-27B (xhigh) & 34 & 1409 & 56.0 \\
        HY4-Preview (high) & -- & -- & 51.3 \\
        DeepSeek-V4-Flash (high) & 26 & -- & 52.0 \\
        \bottomrule
    \end{tabular}
    \endgroup
    \par\smallskip
    \parbox{\linewidth}{\scriptsize
        Scores from Intelligence Index v4.3.2 and GDPval-AA v2.1, accessed 25 September 2026. Default reasoning efforts follow the OpenRouter model catalog on the same date. Dashes mark scores that Artificial Analysis does not list: HY4-Preview has no entry, and GDPval-AA lists DeepSeek-V4-Flash only at maximum effort. DeepSeek-V4-Flash refers to the April 2026 checkpoint, to which the OpenRouter catalog mapped this model name during our runs.
    }
\end{table}

\subsection{Backbone Default Settings}
\label{app:backbone-defaults}

Under No Memory, the PA knows nothing about the user, so its selections show each backbone's own tendency. For each backbone and attribute, we pool its No Memory selections over the four users, the three repeats, and both final and pre-event probes, and take the setting it selects most often as its default. Table~\ref{tab:backbone-defaults} gives this default and its share of the pooled selections. For example, GPT-5.6-Sol selects \textit{Detailed} for Verbosity in all 24 of its selections.

\begin{table}[!tb]
    \centering
    \caption{Default setting of each backbone under No Memory. Each cell gives the setting the PA selects most often for the attribute, with its share and count over final and pre-event probes.}
    \label{tab:backbone-defaults}
    \begingroup
    \footnotesize
    \setlength{\tabcolsep}{3pt}
    \renewcommand{\arraystretch}{1.12}
    \begin{tabularx}{\linewidth}{@{}>{\raggedright\arraybackslash}p{0.2\linewidth}*{4}{>{\raggedright\arraybackslash}X}@{}}
        \toprule
        \textbf{Attribute} & \textbf{GPT-5.6-Sol} & \textbf{Qwen3.8-27B} & \textbf{HY4-Preview} & \textbf{DeepSeek-V4-Flash} \\
        \midrule
        \multicolumn{5}{@{}l}{\textit{Expression Style}} \\
        Tone \& Formality & Consultative 70\% (23/33) & Consultative 42\% (14/33) & Consultative 67\% (22/33) & Consultative 91\% (30/33) \\
        Verbosity & Detailed 100\% (24/24) & Detailed 71\% (17/24) & Detailed 96\% (23/24) & Detailed 62\% (15/24) \\
        Emotional Engagement & Relationship-focused 100\% (27/27) & Relationship-focused 100\% (27/27) & Relationship-focused 100\% (27/27) & Relationship-focused 100\% (27/27) \\
        Guidance Level & Guided 89\% (24/27) & Guided 59\% (16/27) & Guided 67\% (18/27) & Assumed 78\% (21/27) \\
        \midrule
        \multicolumn{5}{@{}l}{\textit{Disclosure}} \\
        Reasoning Visibility & Show 100\% (24/24) & Show 96\% (23/24) & Show 96\% (23/24) & Show 100\% (24/24) \\
        Uncertainty Expression & Express 100\% (27/27) & Express 100\% (27/27) & Express 96\% (26/27) & Express 85\% (23/27) \\
        Process Visibility & Bookend 91\% (30/33) & Bookend 100\% (33/33) & Bookend 94\% (31/33) & Bookend 91\% (30/33) \\
        \midrule
        \multicolumn{5}{@{}l}{\textit{Initiative \& Autonomy}} \\
        Autonomy Level & Autonomous 51\% (18/35) & Autonomous 67\% (24/36) & Self-directed 50\% (18/36) & Self-directed 100\% (36/36) \\
        Proactive Outreach & High 88\% (21/24) & High 88\% (21/24) & High 82\% (18/22) & High 88\% (21/24) \\
        Task Expansion & Low 100\% (24/24) & Low 100\% (24/24) & Low 83\% (20/24) & Low 96\% (23/24) \\
        Solution Breadth & High 55\% (18/33) & Low 94\% (31/33) & Low 94\% (31/33) & Low 97\% (32/33) \\
        Capability Boundary & Suggest alternatives 71\% (17/24) & Suggest alternatives 88\% (21/24) & Suggest alternatives 62\% (15/24) & Suggest alternatives 79\% (19/24) \\
        \midrule
        \multicolumn{5}{@{}l}{\textit{Information Flow}} \\
        Information Elicitation & Structured 50\% (15/30) & Infer 50\% (15/30) & Infer 57\% (16/28) & Infer 52\% (14/27) \\
        Topic Management & Organize 97\% (28/29) & Organize 97\% (29/30) & Organize 100\% (30/30) & Organize 93\% (28/30) \\
        \bottomrule
    \end{tabularx}
    \endgroup
\end{table}

\subsection{Delivered Behavior after a Correct Selection}
\label{app:delivery-examples}

Under Oracle, the five settings with the lowest pass rates are \textit{Hide} for Reasoning Visibility (38\%), \textit{Terse} for Verbosity (50\%), \textit{Iterative} (61\%) and \textit{Structured} (74\%) for Information Elicitation, and \textit{Low} for Solution Breadth (77\%) (Finding~\ref{sec:res-delivery}). Figure~\ref{fig:case-default} shows one of these shortfalls. In an Oracle final probe for User C at work, the HY4-Preview PA selects \textit{Hide} for Reasoning Visibility, and the IX Wrapper withholds the rationale written in a separate field. The PA still writes three reasons in the reply under ``Why that day:'', and the judge gives a BRS of 1. Figure~\ref{fig:case-default} gives the Disclosure case, and the rest of this section gives one example for each of the other three dimensions in which the PA selects the target setting and the delivered reply still falls short. The judge gives each of them a BRS of 3 or lower. Quoted replies are shortened with ellipses.

\paragraph{Expression Style.} The example targets \textit{Terse} for Verbosity, which asks for the answer in the first sentence and at most one supporting move. Under File memory, User C asks the Qwen3.8-27B PA what corporate requires before an equipment purchase above \$600 and which records the branch must keep. The PA selects \textit{Terse} and plans ``one compact block, with at most one supporting note.'' Its reply opens with when the rules were issued and whom they cover, lists six numbered steps, and closes with ``One thing that bites people most''. The judge gives a BRS of 1, and the user answers, ``That is a lot of words for a simple policy.''

\paragraph{Initiative \& Autonomy.} Under Oracle, User B asks the HY4-Preview PA which Halden desk lamp to buy for recording self-tapes. The target \textit{Low} for Solution Breadth asks for one option with the reasons it works, and naming an alternative, even to show why the choice wins, counts as a list. The PA selects \textit{Low} and recommends the Halden LED Desk Lamp, but it gives each reason as a comparison with the Mini, as in ``Five-step dimmer vs.\ the Mini's two steps''. The judge gives a BRS of 1, and the user answers, ``you don't need to walk me through everything the other lamp couldn't do.'' HY4-Preview also selects \textit{Low} under No Memory (Table~\ref{tab:backbone-defaults}), so the shortfall lies in its habit of arguing for a choice by comparing each point with the alternative.

\paragraph{Information Flow.} The example targets \textit{Iterative} for Information Elicitation, which asks the PA to request each missing fact in its own turn before it acts. User D wants a calendar hold for dropping off prints at a framing shop and has not yet chosen the day or the arrival time. Under Oracle, the HY4-Preview PA selects \textit{Iterative}, but in the same call it states that ``Nothing is missing this turn'' and that it will make the calendar hold now. It sets the hold from the user's 17:00 departure and tells the user, ``Two things I filled in myself, since you didn't specify''. The user answers, ``You really should have asked me what time I wanted to arrive instead of just guessing.'' The judge gives a BRS of 3.

In the first two examples the PA says more than its setting allows, and in the last it fills in a missing fact instead of asking, the two tendencies of Finding~\ref{sec:res-delivery}.

\section{Validity Checks}
\label{app:evaluation-6}
\paragraph{Is each preference exposed?} Two annotators read a random 30\% of each user's accumulation sessions, 28 of the 94 per user and 112 in total, all from the runs with File memory. For each session, they judged whether a reader who sees only the visible conversation can identify the target setting from the user's reactions. The preference was conveyed in 99.3\% of the sessions, with an inter-annotator agreement of 95.4\% (Gwet's AC1; \citealp{gwet2008computing}).

\paragraph{Does the simulator leak a preference?} The post-hoc audit in Appendix~\ref{app:simulator-leak} found no leakage.

\paragraph{Does the simulator follow the script?} Two annotators read a random 10\% of the accumulation sessions and checked that the simulator played each session as scripted: it gives each task fact at its beat, plays the decision point and the reaction beat, and stays in character. They found no session that deviates from its script, with an inter-annotator agreement of 99.1\% (Gwet's AC1).

\paragraph{Do BRS and PRS agree?} On the 7{,}996 final probes that have both scores, BRS and PRS agree exactly on 65.3\% and within one point on 90.0\% (Spearman $\rho = 0.83$, 95\% CI [0.82, 0.84]). Agreement is similar under No Memory ($\rho = 0.83$) and the memory conditions ($\rho = 0.83$) and lower under Oracle ($\rho = 0.66$), where most scores lie at 4--5. Both scores rank File memory first for every backbone (Figures~\ref{fig:prs-results} and~\ref{fig:brs-model}). Gemini-3.8-Flash gives both BRS and PRS, so part of their agreement could come from the shared model. We also re-judged a stratified sample of 418 probes with Claude Sonnet 5, which serves neither as PA nor as simulator. Its agreement with PRS is lower than that of the Gemini judge ($\rho = 0.71$ vs.\ $0.85$; difference $-0.14$, 95\% CI [$-0.20$, $-0.09$]) but close to its agreement with the Gemini judge ($\rho = 0.76$). The BRS--PRS agreement therefore mainly reflects the interaction itself.

\paragraph{Which score compares settings?} PRS is the user simulator's own rating, given as the user, so it carries the user's temperament along with the quality of the reply. BRS applies the same rubric for every user. The two scores for each user on the correctly selected Oracle final probes are given in Table~\ref{tab:brs-prs-user}. They agree on 77.3\% of these probes, and PRS falls below BRS on 16.1\%. This share differs by user, from 11.0\% for User A to 21.5\% for User C. The two scores therefore serve different purposes. PRS measures how each user receives the reply, and BRS compares settings on equal terms, since each target setting is held by a different set of users (Finding~\ref{sec:res-delivery}).

\begin{table}[!htbp]
    \centering
    \caption{BRS and PRS for each user on the Oracle final probes with a correct selection. PRS falls below BRS on 11.0\% of these probes for User A but on 21.5\% for User C. PRS therefore reflects how each user receives a reply, while BRS applies one rubric to all users and is used to compare settings.}
    \label{tab:brs-prs-user}
    \small
    \begin{tabular}{@{}lrrrrrr@{}}
        \toprule
        \textbf{User} & \textbf{Probes} & \textbf{Mean BRS} & \textbf{Mean PRS} & \textbf{PRS $<$ BRS} & \textbf{PRS $=$ BRS} & \textbf{PRS $>$ BRS} \\
        \midrule
        User A & 335 & 4.85 & 4.69 & 11.0\% & 87.2\% & 1.8\% \\
        User B & 330 & 4.65 & 4.54 & 14.8\% & 77.9\% & 7.3\% \\
        User C & 331 & 4.48 & 4.19 & 21.5\% & 71.6\% & 6.9\% \\
        User D & 334 & 4.47 & 4.34 & 17.1\% & 72.5\% & 10.5\% \\
        \midrule
        All & 1{,}330 & 4.61 & 4.44 & 16.1\% & 77.3\% & 6.6\% \\
        \bottomrule
    \end{tabular}
\end{table}

\section{Prompt Directory and Templates}
\label{app:prompts}

\subsection{Prompt Directory}
\label{app:prompts-1}

This appendix gives the prompts of the LLM calls in the construction pipeline, at runtime, and in evaluation. Table~\ref{tab:prompt-directory} lists them by stage. Each box is named by the component that sends the prompt. The construction prompts and the two message-generation prompts of the user simulator are long, so their boxes summarize each prompt by its purpose, when it is called, its inputs and output, and its key rules, and then quote one or two passages of the original text. The other boxes show the prompts themselves, and the parts of such a box are the messages of one call or the variants of one prompt. Text in braces or angle brackets is filled in at run time. Highlighted lines are our notes or mark omitted passages, and they are not sent to the model. The full prompts are released with the code.

\begin{table}[p]
    \centering
    \caption{Directory of the prompts in Appendices~\ref{app:prompts-2}--\ref{app:prompts-4}. Each is named by the component that sends it.}
    \label{tab:prompt-directory}
    \begingroup
    \footnotesize
    \setlength{\tabcolsep}{3.5pt}
    \renewcommand{\arraystretch}{1.12}
    \begin{tabularx}{\linewidth}{@{}>{\raggedright\arraybackslash}p{0.23\linewidth}>{\raggedright\arraybackslash}p{0.2\linewidth}>{\raggedright\arraybackslash}X>{\raggedright\arraybackslash}X@{\hspace{3pt}}r@{}}
        \toprule
        \textbf{Prompt} & \textbf{Called} & \textbf{Main inputs} & \textbf{Output} & \textbf{Page} \\
        \midrule
        \multicolumn{5}{@{}l}{\textit{Construction}} \\
        Preference Extractor, Passes 1 and 2 & Pass 1 once per scene of the source character, Pass 2 once per attribute--context cell & Scene transcript; Pass-1 evidence, attribute definition, settings, rubric & Evidence items per attribute; setting, confidence, cited evidence & \pageref{prompt:pref-extractor} \\
        Identity Extractor, Identity Synthesizer, Personality Rater & Once per scene, once per section group, once per character & Scene transcript; observations of all scenes & Identity sections with cited scenes; Big Five facet and Jungian function levels & \pageref{prompt:identity} \\
        Anonymizer and Anonymization Validator & Once per named life block; once per probe reply & Named life block; probe message, persona reply & Substitution table; attributes the persona pushed back on & \pageref{prompt:anonymization} \\
        Self-View Extractor & Once per file or cluster of files & Identity, world canon, file content & What the user recalls of the file & \pageref{prompt:self-view} \\
        Session Outline and Accumulation Session Generators & Once per batch of sessions; once per session & Dealt slots, persona, preference definitions; outline entry, world records, tools & Session outlines; session draft with task facts and beats & \pageref{prompt:acc-construction} \\
        Probe Outline and Probe Session Generators & Once per probe each & Cell, errand, attribute definition; probe skeleton, outline entry & World binding, user ask, viable answers; runtime prose of the probe & \pageref{prompt:probe-construction} \\
        Change-Event Generator and Transcript Writer & Once per change event each & Old and new settings, outline entry; change-event draft & Change-event draft; fixed transcript & \pageref{prompt:event-construction} \\
        Dialogue Naturalness & Appended to four generators and the simulator & None & Constraints on user wording & \pageref{prompt:naturalness} \\
        \midrule
        \multicolumn{5}{@{}l}{\textit{Runtime}} \\
        User Simulator: Profile Context & Every simulator turn & Anonymized identity, tool policy, own memory & Profile block of the simulator prompt & \pageref{prompt:sim-profile} \\
        User Simulator: Message Generation in Accumulation Sessions & Every simulator turn of an accumulation session & Profile, active skills, session script, current beat, task facts, dialogue & Next user message & \pageref{prompt:sim-acc} \\
        User Simulator: Message Generation in Probes & Every simulator turn of a probe & Same inputs as in accumulation sessions & Next user message & \pageref{prompt:sim-probe} \\
        Simulator Message Checker & Every candidate simulator message & Candidate message, PA's previous message, current beat & Verdict on each rule question & \pageref{prompt:sim-check} \\
        PA: System Prompt & Every PA call & Memory content of the condition & System message & \pageref{prompt:pa-system} \\
        PA: Interaction Tools & At decision points & Settings and their instructions & Selected setting & \pageref{prompt:pa-ix} \\
        \midrule
        \multicolumn{5}{@{}l}{\textit{Evaluation}} \\
        Judge: System Prompt & Once per probe & Universal anchors & Scoring rules & \pageref{prompt:judge-system} \\
        Judge: User Message & Once per probe & Probe, scene, dialogue, tool trajectory, rubric & BRS score with evidence & \pageref{prompt:judge-user} \\
        Judge: Attribute Rubrics & Once per probe & Probed attribute & Criteria for the score & \pageref{prompt:judge-rubrics} \\
        \bottomrule
    \end{tabularx}
    \endgroup
\end{table}

\subsection{Construction Prompts}
\label{app:prompts-2}

The construction prompts build each user from the transcripts (Appendix~\ref{app:personas}) and each session from the user's world (Appendix~\ref{app:construction}). The instructions for world authoring and session checks are released with the code. The conditions these checks apply are given in Table~\ref{tab:viability}.

\begin{promptbox}{Preference Extractor: Scene Evidence and Setting Synthesis}\label{prompt:pref-extractor}
\footnotesize
\promptpart{Pass 1: Scene Evidence}
\promptfield{Purpose}Tags the evidence one transcript scene gives about the interaction preferences of the source character. Pass 2 turns this evidence into the preference settings behind User A--D.
\promptfield{Called}Once per scene in which the source character speaks. Scenes with fewer than 3 dialogue turns are skipped.
\promptfield{Inputs}
\begin{promptitems}
\item The 14 attributes with their settings, and the work and personal contexts
\item The name of the source character, the scene reference and the scene description
\item The scene transcript, one line per turn
\end{promptitems}
\promptfield{Output}Per scene: relevance, one context, and for each tagged attribute the evidence type, a direct quote and the setting it points to.
\promptfield{Key rules}
\begin{promptitems}
\item Tag an attribute only on clear behavioral evidence. Quote the dialogue directly.
\item Assign each relevant scene to one context, work or personal. A scene that fits neither is marked not relevant.
\item Evidence is production (how the character speaks and behaves), reception (how the character reacts to the styles of others) or an explicit statement.
\item Production is only a prior. Reception is the strongest signal. When the two diverge, the extractor notes the asymmetry, and the PA preference follows reception.
\end{promptitems}
\promptpart{Pass 2: Setting Synthesis}
\promptfield{Purpose}Names the setting of the source character for one attribute in one context. Each result fills one cell of the $2\times 14$ preference matrix.
\promptfield{Called}Once per (attribute, context) cell that has Pass-1 evidence.
\promptfield{Inputs}
\begin{promptitems}
\item The attribute, its description and its settings
\item The context, its description and the rubric for the attribute
\item The Pass-1 evidence for the cell
\end{promptitems}
\promptfield{Output}One of the listed settings, a confidence level, an evidence summary, and the production and reception quotes behind the setting.
\promptfield{Key rules}
\begin{promptitems}
\item Three layers decide the setting. Frequency patterns give the direction, production patterns anchor a setting, and reception confirms or overrides it.
\item Valid reception signals are explicit confusion or discomfort with a register, a change in engagement in response to style, verbal meta-commentary, and accommodation shifts. Plot-driven topic changes, content disagreement and scripted dramatic reactions are invalid.
\item Confidence levels. High: production and reception agree across several scenes. Medium-High: strong production and some reception, or strong reception in few scenes. Medium: production only, or ambiguous reception. Low: thin or contradictory evidence.
\item The setting is always one of the listed settings. When the evidence is thin or contradictory, the call still names the best-fitting setting at Low confidence. Every cell below High goes to human review.
\end{promptitems}
\promptexcerpt{From the Pass 1 system prompt}
\promptfile{Appedix_TableAndFigure/prompts/excerpts/pref-extractor_1.txt}
\promptexcerpt{From the Pass 2 system prompt}
\promptfile{Appedix_TableAndFigure/prompts/excerpts/pref-extractor_2.txt}
\end{promptbox}

\begin{promptbox}{Identity Extractor, Identity Synthesizer, and Personality Rater}\label{prompt:identity}
\footnotesize
\promptpart{Identity Extractor}
\promptfield{Purpose}Records what one scene shows about who the source character is and how they think, talk and react. The observations feed the user identity that the simulator plays.
\promptfield{Called}Once per scene in which the source character speaks.
\promptfield{Inputs}
\begin{promptitems}
\item The observation sections, from life facts to personality evidence
\item The name of the source character, the scene reference and the scene description
\item The scene transcript
\end{promptitems}
\promptfield{Output}Observations, each with one section, a one-sentence note and a verbatim quote.
\promptfield{Key rules}
\begin{promptitems}
\item Describe behavior only. The word ``prefers'' and the preference setting names never appear.
\item Record only what this scene shows, with no outside knowledge of the show. A description by another character counts only when the behavior of the source character in the scene confirms it.
\item Every observation carries a verbatim quote of at most 40 words. Observations with no quote are discarded.
\end{promptitems}
\promptpart{Identity Synthesizer}
\promptfield{Purpose}Merges the per-scene observations into a few stable items per section of the user identity.
\promptfield{Called}Once per section group (eight groups). A section with over 250 observations gets two half calls and a merge call.
\promptfield{Inputs}
\begin{promptitems}
\item The sections of the call and their item caps
\item Their observations, each with scene reference, note and quote
\item In a merge call, the items of both halves
\end{promptitems}
\promptfield{Output}Pattern items with their scene references. A script then sets confidence from scene and episode counts. Medium items go to human review.
\promptfield{Key rules}
\begin{promptitems}
\item Merge observations that describe the same pattern and keep all their scene references. Related but distinct patterns stay apart.
\item Cite only scene references from the input. Confidence is left to a script.
\item Item text holds no quote, name, place or institution. Catchphrases are the one exception.
\end{promptitems}
\promptpart{Personality Rater}
\promptfield{Purpose}Rates the 15 Big Five facets and the 8 Jungian functions of the source character. A script derives the five dimensions and the function stack.
\promptfield{Called}Once per source character, plus at most one retry for ratings that carry a level with no usable evidence.
\promptfield{Inputs}
\begin{promptitems}
\item The rubric with the level scale and the facet and function indicators
\item The Big Five and Jungian observations, grouped by facet and function
\item On a retry, the ratings to redo
\end{promptitems}
\promptfield{Output}A level from \texttt{very\_low} to \texttt{very\_high}, or \texttt{null}, for each facet and function. Each non-\texttt{null} level comes with 1--3 cited quotes.
\promptfield{Key rules}
\begin{promptitems}
\item Levels reflect consistency and cost across contexts, not topic frequency.
\item The level \texttt{mid} means mixed or context-dependent evidence. No usable evidence gives \texttt{null}.
\item Each non-\texttt{null} level cites 1--3 quotes from the observations of its own facet or function. The script rejects a cited scene from outside that group.
\item Rate from the observations alone. Outside knowledge of the show is excluded, and no function is inferred from an MBTI type.
\end{promptitems}
\promptexcerpt{From the Identity Extractor system prompt}
\promptfile{Appedix_TableAndFigure/prompts/excerpts/identity_1.txt}
\promptexcerpt{From the Identity Synthesizer system prompt}
\promptfile{Appedix_TableAndFigure/prompts/excerpts/identity_2.txt}
\end{promptbox}

\begin{promptbox}{Anonymizer and Anonymization Validator}\label{prompt:anonymization}
\footnotesize
\promptpart{Anonymizer}
\promptfield{Purpose}It removes the identity markers from the life block of a named persona and keeps the archetype. It proposes a fictional replacement for each named person, workplace, and place.
\promptfield{Called}Once per named life block. A human reviews the proposed table, and a script then replays it over the named life block.
\promptfield{Inputs}
\begin{promptitems}
\item Occupation: role, field, workplace, and workplace type
\item Living place and arrangement
\item Relationships with name, role, and closeness
\item Routines and hobbies
\item Replacement names other users already hold
\end{promptitems}
\promptfield{Output}The anonymized life block and a typed substitution table. Only the table is kept.
\promptfield{Key rules}
\begin{promptitems}
\item The model sees only the life block, the one block with identity markers. A script handles the rest.
\item Keep the archetype. Roles, workplace types, arrangements, closeness, and the behavior in routines and hobbies stay as they are.
\item Replace the workplace, the city or region, and every named person, place, institution, or employer in any field. The field changes only when it could point to a known person.
\item New names are fictional and match no real school, company, place, or celebrity. One original maps to one replacement everywhere. Names that other users already hold are never used.
\item Every change goes into the table, which a human reviews and the pipeline replays.
\end{promptitems}
\promptpart{Anonymization Validator: Push-Back Judge}
\promptfield{Purpose}It is part of the check that the anonymized user still reacts to preference violations like the named persona. It names the attributes that one probe reply pushes back on.
\promptfield{Called}Each probe is a PA message that violates several High-confidence settings. The named and anonymized personas each answer it through the user simulator in several rounds. Each reply gets one call.
\promptfield{Inputs}
\begin{promptitems}
\item The 14 attributes with their settings and setting descriptions
\item The probe's PA message, as context only
\item The persona's reply as the PA sees it
\end{promptitems}
\promptfield{Output}A JSON list of pushed-back attributes. Attributes flagged in at least half the rounds form each persona's set. The two sets are then compared with the probe's violations.
\promptfield{Key rules}
\begin{promptitems}
\item Push-back objects to, corrects, rejects, or asks to change how the PA handled the exchange, such as its register, length, or number of options.
\item Judge the handling. A complaint about the subject matter is not push-back.
\item The person will not name the attribute. The judge reads what they object to, whatever words they use.
\item One objection can hit several attributes. A reply that just answers, accepts, or asks a plain follow-up question flags nothing.
\end{promptitems}
\promptexcerpt{From the Anonymizer system prompt}
\promptfile{Appedix_TableAndFigure/prompts/excerpts/anonymization_1.txt}
\promptexcerpt{From the Anonymization Validator system prompt}
\promptfile{Appedix_TableAndFigure/prompts/excerpts/anonymization_2.txt}
\end{promptbox}

\begin{promptbox}{Self-View Extractor: What the User Recalls of Their Own Files}\label{prompt:self-view}
\footnotesize
\promptfield{Purpose}It records what the user remembers about each file and paper they own. This covers their name for it, where they think it is, and which details they can state exactly, roughly, or not at all. The generators of outlines, sessions, probes, and change events read this record.
\promptfield{Called}During world construction. In cluster mode, one call covers a group of related objects, smallest group first, and later groups see the earlier tiers. Otherwise one call covers one object. Objects the canon marks as background become \texttt{pa\_only} entries with no call.
\promptfield{Inputs}
\begin{promptitems}
\item An identity card: name, background, traits, speech habits, and what the user knows well or poorly
\item The world canon, a director's view the user does not see
\item An index of the user's other files and papers, with paths and paper titles only
\item How the user already tiered other objects
\item The full text of the object, or of every object in the cluster
\end{promptitems}
\promptfield{Output}One entry per object. It gives the kind (document, paper, or \texttt{pa\_only}), the user's name for it, its location in plain words, and a short first-person note. It then gives three tiers: \texttt{recall\_exactly}, \texttt{recall\_roughly}, and \texttt{would\_not\_recall}.
\promptfield{Key rules}
\begin{promptitems}
\item \texttt{pa\_only} is for a file the user would not know exists, such as the PA's scratch file or a log they never open or act on. A file they wrote, received, saved, or asked for is never \texttt{pa\_only}.
\item The user's name for an object is what they would say out loud, never a file name, path, or paper id.
\item \texttt{recall\_exactly} holds names and terms the user could state from memory. \texttt{recall\_roughly} holds the fuzzy phrase they would say, never the exact value with a hedge. \texttt{would\_not\_recall} holds ids, DOIs, codes, exact dates, and exact amounts that need the file. Each detail sits in one tier.
\item Tiers follow the user's traits. A precise user keeps more names and terms exact, and a casual user keeps most details rough. Even a precise user does not recall ids or DOIs.
\item The user's own commitments and wording are the last things they would look up. A fact shared by several objects gets the same tier in each of them.
\item Every string in \texttt{recall\_exactly} and \texttt{would\_not\_recall} appears character for character in this object's text. Everything must come from the object or the canon.
\end{promptitems}
\promptexcerpt{From the Self-View Extractor system prompt, on the task}
\promptfile{Appedix_TableAndFigure/prompts/excerpts/self-view_1.txt}
\promptexcerpt{From the Self-View Extractor system prompt, on the tiers}
\promptfile{Appedix_TableAndFigure/prompts/excerpts/self-view_2.txt}
\end{promptbox}

\begin{promptbox}{Session Outline Generator and Accumulation Session Generator}\label{prompt:acc-construction}
\footnotesize
\promptpart{Session Outline Generator}
\promptfield{Purpose}Turns each dealt slot, for an accumulation session or a change event, into a story and a measurement plan. It writes no dialogue.
\promptfield{Called}Once per batch of up to 10 sessions. A rejected batch is retried with the rejection reasons.
\promptfield{Inputs}
\begin{promptitems}
\item Dealt slot: date, target cell, setting, errand, opening pattern
\item Attribute definition and probe-table row
\item Dealt probe ingredients and world digests
\item Continuity ledger and persona identity
\end{promptitems}
\promptfield{Output}One JSON entry per session: goal, completion state, coarse phases, preference phase, correct and wrong branches, and ingredient objects.
\promptfield{Key rules}
\begin{promptitems}
\item Every setting needs a reply that is visibly different and reasonable in the scene. \texttt{why\_both\_viable} names the scene fact behind each setting.
\item Both branches share the request, facts, tools and base outcome. The goal carries no behavior cue.
\item Nine attributes are compressed: the task completes in one PA reply. The other five declare one gate. Its key terms cannot be derived from anything the PA can read.
\item In a change event, the old setting's behavior causes a loss the user feels, and \texttt{policy\_shift} names the new setting. After the change, the wrong branch is the old setting's behavior.
\end{promptitems}
\promptpart{Accumulation Session Generator}
\promptfield{Purpose}Expands one accepted outline entry into a runnable draft with tool steps, the user's facts and beats that direct the simulator.
\promptfield{Called}Once per accepted accumulation entry. A failing draft returns in repair mode, up to three attempts.
\promptfield{Inputs}
\begin{promptitems}
\item Accepted outline entry and dealt errand
\item Clock, calendar, world records, documents and self-view
\item Tool catalog and probe-definition row
\item The Dialogue Naturalness text, shown in full in its own box
\end{promptitems}
\promptfield{Output}One JSON draft: \texttt{task\_facts}, \texttt{reveal\_at}, atomic tool steps with readbacks, and the beats.
\promptfield{Key rules}
\begin{promptitems}
\item The PA is unscripted, and no beat scripts the user's response to its preference behavior. That response comes from the preference sheet at run time.
\item A compressed session plays one probe turn and one reaction turn. An uncompressed session has at least five beats, with the first probe beat by beat 4 or the midpoint, whichever is later.
\item The probe beat hands over a new fact the PA needs, and \texttt{reveal\_at} ties the fact to that beat. No earlier beat states it, authorizes an edit, send, booking or payment, or poses the probe's kind of request.
\item The reaction beat has one path per setting. Each path finishes the task and carries no verdict.
\item Text the simulator sees never names the attribute, the setting or the branch labels. The probe says what is wanted, never how.
\end{promptitems}
\promptexcerpt{From the Session Outline Generator system prompt}
\promptfile{Appedix_TableAndFigure/prompts/excerpts/acc-construction_1.txt}
\promptexcerpt{From the Accumulation Session Generator system prompt}
\promptfile{Appedix_TableAndFigure/prompts/excerpts/acc-construction_2.txt}
\end{promptbox}

\begin{promptbox}{Probe Outline Generator and Probe Session Generator}\label{prompt:probe-construction}
\footnotesize
\promptpart{Probe Outline Generator}
\promptfield{Purpose}Writes one probe entry: one question at a fixed world point that reads one preference cell. It binds world objects, writes the user turn (\texttt{ask}) and sketches a complete reply per setting (\texttt{answer}).
\promptfield{Called}Once per cell for the final probe. For each changing cell, a \texttt{pre\_event\_rewrite} call moves the probe to a point before the change event. Up to three attempts.
\promptfield{Inputs}
\begin{promptitems}
\item Cell, settings, operational definition and probe-table row
\item Dealt errand and candidate world objects
\item Capability inventory and candidate gate tools
\item Ground-truth setting, as context only
\end{promptitems}
\promptfield{Output}JSON with \texttt{world}, \texttt{ask} (background, say, avoid, key terms) and \texttt{answer} (one sketch per setting).
\promptfield{Key rules}
\begin{promptitems}
\item The ask is preference-neutral. It sets no length, structure, order or tone.
\item The ask states every fact the PA would need to ask back, so only the target attribute is diagnosable. For Information Elicitation, it leaves out exactly the missing fields.
\item The five uncompressed attributes carry exactly one real gate: a required tool parameter whose value is nowhere in the world. The others carry none.
\item The ask never refers to earlier sessions or to how the PA usually answers.
\item The pre-event probe keeps the task condition, feasible options and trigger. It rewrites only what the clock touches.
\end{promptitems}
\promptpart{Probe Session Generator}
\promptfield{Purpose}Writes the runtime prose of one probe draft. Code has already built the steps, arguments and active beat, which copies \texttt{ask.say} and \texttt{ask.avoid}.
\promptfield{Called}Once per probe entry, final and pre-event. Up to three attempts.
\promptfield{Inputs}
\begin{promptitems}
\item Code-built skeleton and probe entry
\item Clock, calendar, world records, documents and self-view
\item Target setting, for orientation only
\item The Dialogue Naturalness text, shown in full in its own box
\end{promptitems}
\promptfield{Output}JSON with scenario, task, \texttt{life\_context}, step descriptions, 0--2 setup beats, the active beat's goal and fallback, and gate values.
\promptfield{Key rules}
\begin{promptitems}
\item It adds prose. It never changes a step, tool, argument or the active beat's instruction.
\item The target setting is written nowhere. The reply under every setting stays a complete answer.
\item The gate value is the user's own phrasing. It is absent from the world and said only at the active beat.
\item A setup beat asks nothing that one setting would answer and another would not. The fallback never treats a PA question as a failure.
\item On the triggered Proactive Outreach shape, the two openings react only to whether contact came. Neither says whether it was welcome.
\end{promptitems}
\promptexcerpt{From the Probe Outline Generator system prompt}
\promptfile{Appedix_TableAndFigure/prompts/excerpts/probe-construction_1.txt}
\promptexcerpt{From the Probe Session Generator system prompt}
\promptfile{Appedix_TableAndFigure/prompts/excerpts/probe-construction_2.txt}
\end{promptbox}

\begin{promptbox}{Change-Event Generator and Change-Event Transcript Writer}\label{prompt:event-construction}
\footnotesize
\promptpart{Change-Event Generator}
\promptfield{Purpose}Expands the outline entry of one change event into a draft of tool steps and beats. The beats lead from the old-setting behavior to the user's new rule.
\promptfield{Called}Once per change event. A draft that fails the rule-based validator can be repaired in a new call.
\promptfield{Inputs}
\begin{promptitems}
\item The outline entry, with the old and new settings and any gate
\item The probe definition row of the target attribute
\item World records and documents on the event date, and the tool catalog
\item The user's identity and self-view
\end{promptitems}
\promptfield{Output}One draft with tool steps, beats, task facts and \texttt{reveal\_at}, one forced PA behavior, the causal chain, and the unplayed new-setting path.
\promptfield{Key rules}
\begin{promptitems}
\item The PA performs the outline's wrong branch, the old setting, on one behavior beat. It never performs the correct branch.
\item The causal order is the old-setting behavior, a consequence the user meets on the next beat, an optional catalyst, the user's own insight, and the new rule on the final beat. The four required stages take distinct beats in strict order.
\item The new setting is never played out. The draft describes it as a separate path on the same task, and this path avoids the cost.
\item The cost follows from the tested behavior. The event never depends on information withheld so that the PA fails, on a broken tool, or on a second preference difference.
\item The user asks for the thing, never for how it should be done, and no beat names the old or new setting.
\item Any gated wording is given before the behavior beat, so a locked gate cannot pass for the behavior.
\end{promptitems}
\promptpart{Change-Event Transcript Writer}
\promptfield{Purpose}Turns the accepted draft into a fixed transcript. The run replays it into the world and into the PA's history and memory.
\promptfield{Called}Once per change event, after its draft is accepted.
\promptfield{Inputs}
\begin{promptitems}
\item The accepted draft and the runtime fields bound from it
\item The world's answer to each read
\item The user's identity and the old- and new-setting skills
\end{promptitems}
\promptfield{Output}Ordered user, PA, and third-party messages with the PA's tool calls, and a short first-person memory seed for the user.
\promptfield{Key rules}
\begin{promptitems}
\item The PA first shows the exact old-setting behavior. The user then meets the consequence and states the new rule for the reason in the draft.
\item Every beat gets at least one exchange of its own, in the draft's order.
\item The last message is the user's new rule, in the user's own words and visible to the PA.
\item The event turns on the target attribute only, and the user states no second standing preference.
\end{promptitems}
\promptexcerpt{From the Change-Event Generator system prompt}
\promptfile{Appedix_TableAndFigure/prompts/excerpts/event-construction_1.txt}
\promptexcerpt{From the Change-Event Transcript Writer system prompt}
\promptfile{Appedix_TableAndFigure/prompts/excerpts/event-construction_2.txt}
\end{promptbox}

\begin{promptbox}{Session Generators and User Simulator: Dialogue Naturalness}\label{prompt:naturalness}
\promptfile{Appedix_TableAndFigure/prompts/construction/user_dialogue_naturalness.txt}
\end{promptbox}

\subsection{Runtime Prompts}
\label{app:prompts-3}

This section gives the runtime prompts of the user simulator, its message checker, and the PA. The simulator prompt is assembled on every turn from the profile context, the preference skills of the current beat (Appendix~\ref{app:simulator-3}), and the session script (Appendix~\ref{app:simulator}). Before a simulator message reaches the PA, an LLM checker answers a fixed set of questions about it, and each check call combines a shared call shape with the questions of one or more rules. The preference skills and the File memory consolidation prompt are released with the code. The PA receives the same system prompt in accumulation sessions and probes, and it is offered the interaction tool of an attribute at each decision point about that attribute (Appendix~\ref{app:wrapper}).

\begin{promptbox}{User Simulator: Profile Context}\label{prompt:sim-profile}
\promptpart{Role frame}
\promptfile[highlightlines={1}]{Appedix_TableAndFigure/prompts/runtime/sim_profile_role_frame.txt}
\promptpart{Ground rules}
\promptfile[highlightlines={1}]{Appedix_TableAndFigure/prompts/runtime/sim_profile_ground_rules.txt}
\promptpart{Tool policy}
\promptfile[highlightlines={1}]{Appedix_TableAndFigure/prompts/runtime/sim_profile_tool_policy.txt}
\promptpart{Assembled profile context}
\promptfile[highlightlines={1}]{Appedix_TableAndFigure/prompts/runtime/sim_profile_context_assembly.txt}
\end{promptbox}

\begin{promptbox}{User Simulator: Message Generation in Accumulation Sessions}\label{prompt:sim-acc}
\footnotesize
\promptfield{Purpose}Writes the user's next message in an accumulation session. The message plays the current beat and shows the active preferences through what the user asks for and how the user reacts to the PA.
\promptfield{Called}On every simulator turn of an accumulation session, including each rewrite after a failed message check.
\promptfield{Inputs}
\begin{promptitems}
\item Profile context: the anonymized identity of User A--D, the tool policy, and the user's own memory files
\item The preference skills the current beat declares, at the user's current setting, and the reception part of skills the previous beat declared
\item Life context with the in-world clock, the session outline, and the current beat
\item Task facts tagged as said or not yet said, with facts that \texttt{reveal\_at} assigns to a later beat left out
\item The visible conversation, a private record of what the PA's last reply did, and any held items
\end{promptitems}
\promptfield{Output}A message that reaches the PA after the message checks, and private blocks: a factual check, the beat decision, and consent decisions on held items.
\promptfield{Key rules}
\begin{promptitems}
\item The simulator speaks only as the user. It never mentions the benchmark, the script, the beat, or the simulator.
\item It shows each active preference through behavior. It never declares one, as in ``I prefer X'', and never names the preference or the attribute.
\item It holds each task fact until its beat calls for it or the PA asks for it. Facts that \texttt{reveal\_at} assigns to a later beat are left out of the prompt. Asked for one of those, the user says it will come at that step.
\item It reacts only to behavior that has appeared in the PA's replies. A branch that turns on what the PA did is chosen from a private run-log record of the reply together with its visible text, never from the reply's own account of itself.
\item On the first turn after a reply shows the surface of an active preference, the user takes a position on that surface. The user says this once and then returns to the task.
\item The user accepts an offer or gives a go-ahead only when the current beat carries it.
\item The user leaves the shape of the answer to the PA. Requirements come out as reactions over the conversation, never as a checklist at the start.
\item The user never invents a value, a source to check, or a result of work done outside the conversation.
\end{promptitems}
\promptexcerpt{From the global rules of the system prompt}
\promptfile{Appedix_TableAndFigure/prompts/excerpts/sim-acc_1.txt}
\promptexcerpt{From the active-skills block of the system prompt}
\promptfile{Appedix_TableAndFigure/prompts/excerpts/sim-acc_2.txt}
\end{promptbox}

\begin{promptbox}{User Simulator: Message Generation in Probes}\label{prompt:sim-probe}
\footnotesize
\promptfield{Purpose}Writes the user's next message in a pre-event or final probe. The user asks for the task and takes the PA's answer as it comes, with no feedback on its manner.
\promptfield{Called}On every simulator turn of a probe. The simulator uses this prompt in place of the accumulation prompt.
\promptfield{Inputs}
\begin{promptitems}
\item The same profile context, active skills, life context, outline, beat, and task facts as in accumulation sessions
\item The visible conversation of the probe
\item The same per-turn message, with the private record of the PA's last reply and any held items
\end{promptitems}
\promptfield{Output}The same blocks as in accumulation sessions. The factual check records factual errors only and never the form of the reply.
\promptfield{Key rules}
\begin{promptitems}
\item The probe is one question on its own. The user has never spoken to this PA and never points back to an earlier session, for example with ``last time'' or ``as usual''.
\item Only the Production side of each active skill applies. It shapes how the user speaks, and its Reception side never becomes a reaction.
\item The user corrects facts only: a wrong value, a wrong document, an invented step, or a claimed action the user did not ask for.
\item The user never steers the manner of the answer. The user gives no praise, complaint, or hint about its form and asks again only for missing content.
\item Mood moves with the content only, never with the manner.
\item As in accumulation sessions, the simulator speaks only as the user, never declares a preference, and never names the preference or the attribute.
\item The rules on task facts, offers, and go-aheads are the same as in accumulation sessions. A beat branch that turns on what the PA did is chosen from the private record of the reply together with its visible text.
\end{promptitems}
\promptexcerpt{From the opening of the system prompt}
\promptfile{Appedix_TableAndFigure/prompts/excerpts/sim-probe_1.txt}
\promptexcerpt{From the global rules of the system prompt}
\promptfile{Appedix_TableAndFigure/prompts/excerpts/sim-probe_2.txt}
\end{promptbox}

\begin{promptbox}{Simulator Message Checker}\label{prompt:sim-check}
\promptpart{Call shape: general form}
\promptfile[highlightlines={1}]{Appedix_TableAndFigure/prompts/runtime/sim_check_shape_general.txt}
\promptpart{Call shape: one rule}
\promptfile[highlightlines={1}]{Appedix_TableAndFigure/prompts/runtime/sim_check_shape_per_rule.txt}
\promptpart{Call shape: two steps}
\promptfile[highlightlines={1}]{Appedix_TableAndFigure/prompts/runtime/sim_check_shape_two_step.txt}
\promptpart{Assembled general-form calls}
\promptfile[highlightlines={1}]{Appedix_TableAndFigure/prompts/runtime/sim_check_general_rendered.txt}
\promptpart{Rule: gate\_cue\_missing}
\promptfile{Appedix_TableAndFigure/prompts/runtime/sim_check_rule_gate_cue_missing.txt}
\promptpart{Rule: probe\_fact\_missing}
\promptfile{Appedix_TableAndFigure/prompts/runtime/sim_check_rule_probe_fact_missing.txt}
\promptpart{Rule: carries\_an\_ask}
\promptfile{Appedix_TableAndFigure/prompts/runtime/sim_check_rule_carries_an_ask.txt}
\promptpart{Rule: withheld\_fact\_before\_ask}
\promptfile{Appedix_TableAndFigure/prompts/runtime/sim_check_rule_withheld_fact_before_ask.txt}
\promptpart{Rule: reaction\_asserts\_unexecuted\_pa\_action}
\promptfile{Appedix_TableAndFigure/prompts/runtime/sim_check_rule_reaction_asserts_unexecuted_pa_action.txt}
\promptpart{Rule: reaction\_authorises\_pa\_action}
\promptfile{Appedix_TableAndFigure/prompts/runtime/sim_check_rule_reaction_authorises_pa_action.txt}
\promptpart{Rule: reaction\_misstates\_contractions}
\promptfile{Appedix_TableAndFigure/prompts/runtime/sim_check_rule_reaction_misstates_contractions.txt}
\promptpart{Rule: reaction\_states\_future\_policy}
\promptfile{Appedix_TableAndFigure/prompts/runtime/sim_check_rule_reaction_states_future_policy.txt}
\promptpart{Rule: reaction\_beat\_off\_attribute}
\promptfile{Appedix_TableAndFigure/prompts/runtime/sim_check_rule_reaction_beat_off_attribute.txt}
\promptpart{Rule: reaction\_names\_setting}
\promptfile{Appedix_TableAndFigure/prompts/runtime/sim_check_rule_reaction_names_setting.txt}
\end{promptbox}

\begin{promptbox}{PA: System Prompt}\label{prompt:pa-system}
\promptpart{File memory}
\promptfile[highlightlines={1}]{Appedix_TableAndFigure/prompts/runtime/pa_system_prompt_file_memory.txt}
\promptpart{Simple RAG, Mem0, and MemOS}
\promptfile[highlightlines={1}]{Appedix_TableAndFigure/prompts/runtime/pa_system_prompt_retrieval_memory.txt}
\promptpart{Runtime context at the head of the newest message}
\promptfile[highlightlines={1,10}]{Appedix_TableAndFigure/prompts/runtime/pa_runtime_context_prefix.txt}
\end{promptbox}

\begin{promptbox}{PA: Interaction Tools}\label{prompt:pa-ix}
\promptpart{IX\_tone\_formality}
\promptfile{Appedix_TableAndFigure/prompts/runtime/pa_ix_tool_01_tone_formality.txt}
\promptpart{IX\_verbosity}
\promptfile[highlightlines={5}]{Appedix_TableAndFigure/prompts/runtime/pa_ix_tool_02_verbosity.txt}
\promptpart{IX\_emotional\_engagement}
\promptfile[highlightlines={5}]{Appedix_TableAndFigure/prompts/runtime/pa_ix_tool_03_emotional_engagement.txt}
\promptpart{IX\_guidance\_level}
\promptfile[highlightlines={5}]{Appedix_TableAndFigure/prompts/runtime/pa_ix_tool_04_guidance_level.txt}
\promptpart{IX\_reasoning\_visibility}
\promptfile[highlightlines={5}]{Appedix_TableAndFigure/prompts/runtime/pa_ix_tool_05_reasoning_visibility.txt}
\promptpart{IX\_uncertainty\_expression}
\promptfile[highlightlines={5}]{Appedix_TableAndFigure/prompts/runtime/pa_ix_tool_06_uncertainty_expression.txt}
\promptpart{IX\_process\_visibility}
\promptfile[highlightlines={5}]{Appedix_TableAndFigure/prompts/runtime/pa_ix_tool_07_process_visibility.txt}
\promptpart{IX\_autonomy\_level}
\promptfile[highlightlines={5}]{Appedix_TableAndFigure/prompts/runtime/pa_ix_tool_08_autonomy_level.txt}
\promptpart{IX\_proactive\_outreach}
\promptfile[highlightlines={5,7}]{Appedix_TableAndFigure/prompts/runtime/pa_ix_tool_09_proactive_outreach.txt}
\promptpart{IX\_task\_expansion}
\promptfile[highlightlines={5}]{Appedix_TableAndFigure/prompts/runtime/pa_ix_tool_10_task_expansion.txt}
\promptpart{IX\_solution\_breadth}
\promptfile[highlightlines={5}]{Appedix_TableAndFigure/prompts/runtime/pa_ix_tool_11_solution_breadth.txt}
\promptpart{IX\_capability\_boundary}
\promptfile[highlightlines={5}]{Appedix_TableAndFigure/prompts/runtime/pa_ix_tool_12_capability_boundary.txt}
\promptpart{IX\_information\_elicitation}
\promptfile[highlightlines={5}]{Appedix_TableAndFigure/prompts/runtime/pa_ix_tool_13_information_elicitation.txt}
\promptpart{IX\_topic\_management}
\promptfile[highlightlines={5}]{Appedix_TableAndFigure/prompts/runtime/pa_ix_tool_14_topic_management.txt}
\end{promptbox}

\subsection{Evaluation Prompts}
\label{app:prompts-4}
\label{app:judge}

The judge scores each probe in one call. The system prompt holds the scoring rules and the universal anchors, and the user message holds the probe, the rubric of the probed attribute, and the output format.

\begin{promptbox}{Judge: System Prompt}\label{prompt:judge-system}
\promptfile{Appedix_TableAndFigure/prompts/evaluation/judge_system_prompt.txt}
\end{promptbox}

\begin{promptbox}{Judge: User Message}\label{prompt:judge-user}
\promptpart{Assembled user message}
\promptfile[highlightlines={1,12,21,28,30,34,37,49,59,61,68,74}]{Appedix_TableAndFigure/prompts/evaluation/judge_input_assembly.txt}
\promptpart{Output instructions and JSON schema}
\promptfile{Appedix_TableAndFigure/prompts/evaluation/judge_output_format.txt}
\promptpart{Follow-up message after a rejected reply}
\promptfile{Appedix_TableAndFigure/prompts/evaluation/judge_reask.txt}
\end{promptbox}

\begin{promptbox}{Judge: Attribute Rubrics}\label{prompt:judge-rubrics}
\promptpart{Tone \& Formality}
\promptfile{Appedix_TableAndFigure/prompts/evaluation/rubric_tone_formality.txt}
\promptpart{Verbosity}
\promptfile{Appedix_TableAndFigure/prompts/evaluation/rubric_verbosity.txt}
\promptpart{Emotional Engagement}
\promptfile{Appedix_TableAndFigure/prompts/evaluation/rubric_emotional_engagement.txt}
\promptpart{Guidance Level}
\promptfile{Appedix_TableAndFigure/prompts/evaluation/rubric_guidance_level.txt}
\promptpart{Reasoning Visibility}
\promptfile{Appedix_TableAndFigure/prompts/evaluation/rubric_reasoning_visibility.txt}
\promptpart{Uncertainty Expression}
\promptfile{Appedix_TableAndFigure/prompts/evaluation/rubric_uncertainty_expression.txt}
\promptpart{Process Visibility}
\promptfile{Appedix_TableAndFigure/prompts/evaluation/rubric_process_visibility.txt}
\promptpart{Autonomy Level}
\promptfile{Appedix_TableAndFigure/prompts/evaluation/rubric_autonomy_level.txt}
\promptpart{Proactive Outreach}
\promptfile{Appedix_TableAndFigure/prompts/evaluation/rubric_proactive_outreach.txt}
\promptpart{Task Expansion}
\promptfile{Appedix_TableAndFigure/prompts/evaluation/rubric_task_expansion.txt}
\promptpart{Solution Breadth}
\promptfile{Appedix_TableAndFigure/prompts/evaluation/rubric_solution_breadth.txt}
\promptpart{Capability Boundary}
\promptfile{Appedix_TableAndFigure/prompts/evaluation/rubric_capability_boundary.txt}
\promptpart{Information Elicitation}
\promptfile{Appedix_TableAndFigure/prompts/evaluation/rubric_information_elicitation.txt}
\promptpart{Topic Management}
\promptfile{Appedix_TableAndFigure/prompts/evaluation/rubric_topic_management.txt}
\end{promptbox}

\end{document}